 \documentclass[final,5p,times]{elsarticle}

\usepackage{amssymb}
\usepackage{color, tikz}
\usepackage{color, soul}
\usepackage{xcolor}
\usepackage{graphicx} 
\usepackage{multirow}
\usepackage{booktabs}
\usepackage{lipsum}
\usepackage{array}
\usepackage{siunitx}
\usepackage[breaklinks,colorlinks=true, citecolor=blue, linkcolor=black, urlcolor=blue]{hyperref}
\usepackage{subcaption} 
\usepackage{float}
\usepackage{parskip}

\journal{Environmental Modelling \& Software}

\begin{document}

\begin{frontmatter}



\title{A Machine Learning Framework for Handling Unreliable Absence Label and Class Imbalance for Marine Stinger Beaching Prediction}


\author[inst1]{Amuche Ibenegbu}
\author[inst2,inst3]{Amandine Schaeffer}
\author[inst2]{Pierre Lafaye de Micheaux}
\author[inst1]{Rohitash Chandra}

\affiliation[inst1]{organization={Transitional Artificial Intelligence Research Group, School of Mathematics and Statistics},
            addressline={UNSW}, 
            city={Sydney}, 
            country={Australia}}

\affiliation[inst2]{organization={School of Mathematics and Statistics},
            addressline={UNSW}, 
            city={Sydney}, 
            country={Australia}}

\affiliation[inst3]{organization={Center for Marine Science and Innovation},
            addressline={UNSW}, 
            city={Sydney}, 
            country={Australia}}
\begin{abstract}
Bluebottles (\textit{Physalia} spp.) are marine stingers resembling jellyfish, whose presence on Australian beaches poses a significant public risk due to their venomous nature. Understanding the environmental factors driving bluebottles ashore is crucial for mitigating their impact, and machine learning tools are to date relatively unexplored.
We use bluebottle marine stinger presence/absence data from beaches in Eastern Sydney, Australia, and compare machine learning models (Multilayer Perceptron, Random Forest, and XGBoost) to identify factors influencing their presence. We address challenges such as class imbalance, class overlap, and unreliable absence data by employing data augmentation techniques, including the Synthetic Minority Oversampling Technique (SMOTE), Random Undersampling, and Synthetic Negative Approach pproach that excludes the negative class. 
Our results show that SMOTE failed to resolve class overlap, but the presence-focused approach  effectively handled imbalance, class overlap, and ambiguous absence data. 
The data attributes such as  wind direction, which is a circular variable,   emerged as a key factor influencing bluebottle presence, confirming previous inference studies. However, in the absence of population dynamics, biological behaviours, and life cycles, the best predictive model appear to be Random Forests combined with Synthetic Negative Approach. This research contributes to mitigating the risks posed by bluebottles to beachgoers and provides insights into handling class overlap and unreliable negative class in environmental modelling.


\end{abstract}


\end{frontmatter}


\section{Introduction}
\label{sec:sample1}
 

 \textit{Physalia} spp. are marine stingers that inhabit the ocean surface in tropical and subtropical regions, with occasional occurrences in temperate zones \citep{munro2019morphology}. Known as Portuguese Man-of-War in the Atlantic Ocean \citep{ferrer2017portuguese} and bluebottles in the Indian and Pacific Oceans \citep{bourg2022driving}, their presence along coastlines is frequently reported. \\
There is limited understanding of the factors influencing the distribution and movements of \textit{Physalia} \citep{bourg2022driving}. Similar to jellyfish, these organisms are known for rapid and widespread blooms that can disrupt food webs, biochemical processes, and socioeconomic activities, including tourism, fishing, aquaculture, and industrial operations \citep{purcell2005climate, martins2022modelling}. Their abundance, long tentacles, and painful stings pose threats to humans and marine industries, with stings causing severe pain and sometimes serious health issues, impacting gastrointestinal, muscular, neurological, and cardio-respiratory systems \citep{cegolon2013jellyfish, stein1989fatal}.\\
Limited attention has been given to the occurrence of marine stingers  due to these risks, which extend to beachgoers and marine-related sectors \citep{canepa2014pelagia, purcell2005climate,munro2019morphology}. However, ultimately, the risks are higher when these organisms reach the coast. Hence, the term ``beaching'', commonly associated with stranded whales and dolphins \citep{coombs2019can}, is also used to describe the occurrence of bluebottles and jellyfish on shorelines.\\
To mitigate the negative impacts of marine stingers, it is crucial to understand the environmental factors driving their arrival on beaches \citep{bellido2020atmospheric}. Furthermore, a deeper understanding of the relationship between their migration, beaching events, and environmental conditions is essential for managing their hazards effectively \citep{castro2024using}. 

While research on marine stinger behavior and movements is ongoing, there is limited work on modeling jellyfish beaching events using machine learning tools, and even less for the bluebottle (or \textit{Physalia}). Several studies attempted the task, highlighting critical challenges rising from the type of data. 
\cite{castro2024using} applied a MultiLayer Perceptron (MLP) model to predict jellyfish presence along the Spanish coast. While the model performed well in identifying absence, its precision for detecting presence was low, likely due to class imbalance in the dataset.
Similarly, 
\cite{pontin2009factors} used an MLP to predict \textit{Physalia}'s presence in New Zealand, emphasizing challenges with class imbalance and noisy data. To improve predictions, they incorporated time-lagged information into the input features.  
\cite{albajes2011jellyfish} used remote sensing data to predict jellyfish occurrences but reported highly unbalanced data, with only 9\% of observations in the positive (i.e. presence) class. The model achieved 91\% accuracy, matching the majority class ratio, indicating that it predominantly classified all observations as the negative class (i.e. absence).

\cite{martins2022modelling} investigated the distribution of \textit{Physalia physalis} (bluebottle) in the North Atlantic Ocean (the Azores) at various spatial and temporal scales. The model achieved high accuracy, identifying primary productivity, temperature, and current direction as key macro-ecological factors, while wind patterns and primary productivity were critical at regional scales. However, metrics such as AUC, sensitivity, and specificity were also considered due to the class imbalance.
In a related study, 
\cite{martins2024unravelling} explored environmental drivers of \textit{Physalia} more broadly, including four year data from Eastern Australia \citep{bourg2022driving}. Unfortunately, the seasonal cycle was not considered separately from other factors (such as the temperature or productivity). Moreover, all presence and environmental data were investigated at monthly time-scales. Hence, the key issues for a daily prediction, such as class imbalance, class overlap, or unreliable absence data were not addressed. \\

Class imbalance is a significant challenge in machine learning, affecting both supervised and semi-supervised models \citep{haixiang2017learning, rezvani2023broad, de2023systematic}. Most machine learning classification algorithms assume balanced data, but in real-world scenarios, such as presence / absence data, some classes are more frequent than others. This imbalance biases learning algorithms towards the majority class, resulting in poor classification performance \citep{ghosh2024class, krawczyk2016learning, niaz2022class, khan2023review}.
The issue arises because, during model training, the gradient of the loss function (used to measure the error between predicted and actual values) is dominated by the majority class. This occurs because the majority class contributes more data points, resulting in larger cumulative updates to the model's weights, which are the adjustable parameters used to make predictions. As a result, the learning process becomes biased towards the majority class, making it harder for the model to correctly classify the minority class \citep{anand1993improved, bach2019proposal}.
The performance of classification models on imbalanced data depends on their design and their ability to distinguish between classes. While the models’ ability to correctly classify the majority class typically remains unaffected, their performance on the minority class suffers significantly. As the class distribution becomes more skewed, the overall performance of the models deteriorates \citep{murphey2004neural, masko2015impact}. In the example of a majority class of "absence" (such as for marine stingers), the prediction of the "presence" class, which is the one of interest, would be affected.\\
Class imbalance problems can be addressed using three main approaches: data-level methods (e.g., random oversampling and undersampling), algorithm-level methods (e.g., cost-sensitive learning), and hybrid methods (e.g., cluster-based undersampling, SMOTEBoost, RUSBoost), each with their strengths and limitations \citep{niaz2022class}. \\
Data augmentation is a common data-level technique that generates synthetic training samples and augments them with real data to mitigate imbalance \citep{khan2023review, bayer2022survey}. 
Data augmentation addresses imbalance through re-sampling (oversampling the minority class or undersampling the majority class) or assigning different weights to training examples. Random oversampling duplicates or creates new samples from the minority class, while random undersampling removes samples from the majority class to match the minority one  \citep{bach2019proposal}. \\
SMOTE generates synthetic samples using $k$-Nearest Neighbors instead of duplication, improving performance but facing challenges with sparse or noisy data \citep{chawla2002smote, khan2023review}.\\
Generative Adversarial Networks (GANs) \citep{goodfellow2014generative}, originally developed for image generation, have also been applied to class imbalance problems by creating synthetic data for re-sampling. Combining SMOTE with GANs has shown promise in improving accuracy for tabular datasets \citep{sharma2022smotified}. \\
Beyond augmentation, ensemble learning methods like Random Forests and Gradient Boosting have proven effective in handling class imbalance \citep{galar2011review}. Recent studies, such as in  \cite{khan2023review}, have reviewed and evaluated combinations of ensemble learning models with data augmentation techniques for tackling these challenges.\\
 In some cases, negative data may be missing or imprecise, leaving only positive or unlabeled data, a scenario common in ecology and geoscience. For example, the "absence" of marine stinger is hard to observe with 100\% accuracy, and is highly susceptible to human error.  
\cite{rider2013classifier} evaluated classifiers under such conditions, where the negative class may be missing or mislabeled, and found that data bias significantly affects classifier ranking and performance. 
\cite{fung2005text} addressed this issue with the Positive and Negative Examples Labelling Heuristic (PNLH), which extracts high-quality labeled samples from unlabeled data. PNLH was particularly effective in datasets with very few positive instances.\\
Several methods have been developed for handling datasets without negative samples. Positive Example-Based Learning (PEBL) \citep{yu2004pebl} and SHRINK \textcolor{red}{\citep{kubat1997learning}} label mixed regions as positive regardless of the majority class, focusing on maximizing margins or managing mixed regions. 
\cite{haque2022negative} used a one-class SVM trained on positive data, generating synthetic negative samples for model evaluation.
\cite{zhang2020generative} utilized GANs combined with an LSTM recurrent neural network \citep{hochreiter1997long} to generate negative samples. The GAN produced data closely resembling the original distribution, while the LSTM accounted for time-series characteristics, improving precision and recall compared to baseline models.
In summary, machine learning models face significant challenges with class imbalance and unreliable absence data. Extreme class imbalance often results in overlapping class boundaries, complicating clear separations in feature space. For marine stingers such as bluebottles, absence data in beaching is unreliable as it merely reflects environmental and ecological conditions that are not documented. For example, the population and ecological dynamics of the organisms, or the environmental factors during their whole life-time. Most methods addressing class imbalance prioritize the minority class, often overlooking valuable information from the majority class and struggling with extreme imbalances.\\

In this study, we use machine learning to investigate the environmental factors which drive bluebottle marine organisms to Australian beaches. To address class imbalance, overlap, and unreliable absence data, we apply data augmentation with SMOTE, Random Sampling and a ``no-negative-class" approaches, and use supervised models such as MLP and ensemble methods like Random Forest and XGBoost \citep{ferreira2012boosting, breiman2001random}.
To account for unreliable and absent negative data, we also employ unsupervised learning methods, including one-class SVM \citep{muller2018introduction} and CT-GAN. Environmental factors influencing bluebottle migration to beaches are evaluated, and a dimensionality reduction technique is used to visualize model performance and assess the effectiveness of the data augmentation process.


The remainder of this manuscript is organized as follows. Section~2 reviews related work on class imbalance in environmental problems, bluebottle conservation, and data augmentation. Section~3 outlines the methodology, including model and implementation details. Sections~4 and 5 present the results and discussion, respectively, and Section~6 concludes the study.
\section{Literature review }
\label{sec:sample:appendix}

\subsection {Class Imbalance in environmental problems}
Imbalanced classification datasets are common in scenarios involving rare event (or anomaly) detection, such as identifying uncommon diseases, detecting fraud, or addressing environmental issues \citep{khan2023review}. Class imbalance frequently occurs  in machine learning classification tasks, and here we list several examples related to environmental challenges.
\cite{kim2020predicting}, while predicting cyanobacteria occurrence using climatological and environmental controls in South Korea, encountered class imbalance in the occurrence dataset. To prevent model bias, they proposed a parameter estimation method that assigns equal importance to cases with very few occurrences compared to the more frequent non-occurrence cases.
\cite{mellios2020machine} highlighted the inherent imbalance in their dataset while applying various machine learning classification algorithms to predict the health risks of cyanobacteria blooms in Northern European lakes. Despite achieving high accuracy, they did not consider accuracy as the primary evaluation metric, focusing instead on the confusion matrix.
For early warning systems to predict cyanobacteria bloom outbreaks in freshwater reservoirs using machine learning models, 
\cite{park2021machine} faced an imbalance in the number of observations for each class level. They evaluated model performance similarly to 
\cite{mellios2020machine}, placing emphasis on metrics derived from the confusion matrix.
\cite{kim2022classification} addressed the class imbalance in their dataset by applying oversampling techniques to predict cyanobacteria blooms in Chilgok, South Korea, using a classification-based machine learning model. Similarly, 
\cite{clements2024classification} dealt with a binary class imbalance problem, where normal cases outnumbered algae or wastewater cases in detecting de facto reuse and cyanobacteria in drinking water intake. They employed Cohen’s Kappa, boosting, and bagging methods to mitigate the imbalance and used accuracy as a key performance metric.
\cite{duarte2020addressing} tackled the class imbalance issue in the automatic image classification of coastal marine litter items from orthophotos. They tested various techniques such as class weighting, oversampling, and classifier thresholding, with oversampling yielding the best results based on the F1-score.
Finally, to address the challenge of imbalanced datasets,\cite{hong2020oversampling} applied oversampling techniques to environmental complaints related to construction projects. Their model outperformed existing models in predicting imbalanced environmental complaints, demonstrating improved accuracy and overall performance.

 \subsection{Data augmentation with SMOTE and GANs}

Data augmentation involves techniques for generating synthetic data from existing datasets, particularly for minority classes, to address class imbalance problems \citep{van2001art, sharma2022smotified}. This approach has proven effective in mitigating the issue of insufficient training data in real-world scenarios. The primary objective of data augmentation is to increase the size, quality, and diversity of the training dataset \citep{mumuni2022data}. It has been widely applied across various fields facing class imbalance issues, including both tabular and image data domains. Examples include its use in education\citep{rattan2021analyzing}, data mining \citep{li2011application}, healthcare \citep{joloudari2023effective}, finance \citep{zhang2020generative}, and environmental studies \citep{junsomboon2017combining}. 
The application of SMOTE for addressing class imbalance problems has expanded across various fields, resulting in the development of several SMOTE variants over the years \citep{fernandez2018smote}. Notable variants include SMOTE-ENN, K-Means SMOTE, SMOTE-SVM, Borderline SMOTE, Geometric SMOTE, and Weighted SMOTE \citep{khan2023review}. 
\cite{rattan2021analyzing} applied SMOTE to tackle class imbalance in predicting student performance. 
\cite{li2011application} employed Random-SMOTE (R-S) on five imbalanced datasets for data mining tasks, reporting significant improvements in performance. 
\cite{skryjomski2017influence} introduced a selective oversampling technique by adapting SMOTE to focus on the most challenging minority class samples rather than uniformly sampling all minority instances.

\cite{joloudari2023effective} integrated SMOTE with Convolutional Neural Networks (CNNs) to address binary classification problems, observing that SMOTE-CNN outperformed other comparable methods. 
\cite{hussein2019smote} highlighted two major challenges associated with SMOTE: noise and borderline data samples. To address this, they proposed a method that first generated synthetic instances for the minority class, followed by removing minority class samples near the majority class. This approach consistently outperformed traditional methods across 44 datasets.
 
In a Generative Adversarial Network (GAN), the generator network is responsible for creating realistic samples, while the discriminator network distinguishes these generated samples from the original data \citep{goodfellow2014generative}. 
\cite{lee2021gan} employed GAN to address class imbalance in an intrusion detection system, reporting improved classifier performance. Similarly, 
\cite{kim2020gan} utilized a GAN-based anomaly detection model to tackle class imbalance in defect detection, achieving enhanced predictive performance.
\cite{sharma2022smotified} introduced SMOTified-GAN, which combines synthetic data generated by SMOTE with GAN to create more realistic samples, significantly improving performance in addressing class imbalance problems. 
\cite{farahbakhsh2023prospectivity} applied SMOTified-GAN to model critical mineral deposits using a Random Forest classifier, achieving better prediction accuracy.
\cite{cheah2023enhancing} proposed SMOTE+GAN and GANified-SMOTE for financial fraud detection. They demonstrated improved performance across minority samples generated by these methods, effectively enhancing model outcomes for imbalanced datasets.

\subsection{\textit{Physalia}, the bluebottle}

Jellyfishes, box jellies, and the \textit{Physalia} genus which includes the bluebottle, are of the phylum Cnidaria but through different classes \citep{cegolon2013jellyfish}, and can thrive in both low- and high-productivity environments \citep{mills1995medusae}.
\textit{Physalia} has been reported with differences in size, colors, and habitat \citep{pontin2012molecular,munro2019morphology}, and recently classified in four different species \cite{church2024global}. They are colonies of floating organisms that inhabit the water-air interface (pleustonic organisms), composed of highly specialized and interdependent zooids that rely on one-another for survival. To travel long distances, \textit{Physalia} uses an inflated gas-filled float for buoyancy, while its long tentacles, equipped with venomous cells, can deliver stings as it moves \citep{munro2019morphology}. These aquatic predators are highly effective hunters, using their venomous cells to immobilize and consume fish and fish larvae.\\
\textit{Physalia} colonies exhibit left- or right-handed asymmetry, determined by the position of the underwater zooids and the attachment point of the tentacle relative to the float and sail.\citep{munro2019morphology}. 
\cite{helm2021mysterious} highlighted that \textit{Physalia} and other epleustonic organisms are not uniformly distributed in the ocean and face environmental challenges such as climate change, oil spills, pollution, and fishing. Increased understanding of their habitats and distributions is crucial for their conservation and prediction.\\

A few studies have already attempted to understand the arrival of \textit{Physalia} to the shore.
\cite{pontin2009factors}, in predicting factors influencing the presence of \textit{Physalia} on New Zealand beaches, found that wave and wind directions from the northwest significantly increase their presence. Conversely, wave directions to the north one day prior lead to their absence. 
\cite{martins2022modelling}, modeling \textit{Physalia} occurrence of \textit{Physalia} in the North Atlantic, identified macro-ecological drivers like primary productivity, temperature, and current direction. At regional scales, primary productivity and wind patterns were the key determinants. The study warned that future climate change and reduced primary productivity could cause significant shift in the distribution of this species.
\cite{bourg2022driving} studied \textit{Physalia} beaching events and stings in relation to environmental factors. Their findings revealed a seasonal pattern, with greater occurrences during the Austral summer on the East coast of Australia. The temperature seasonal cycle, with a phase lag of 3–4 months, also influences their distribution. Wind direction emerged as the primary driver of their presence on beaches. Similarly, 
\cite{castro2024using} modeled all jellyfish presence together and included \textit{Physalia}, emphasizing the substantial impact of environmental factors on their occurrence. They recommended regular monitoring of sea surface temperature (SST) and wind conditions to improve prediction accuracy.
\cite{martins2024unravelling} investigated the occurrence of \textit{Physalia} in the North Atlantic (Azores, Portugal) and the southeast pacific (Australian East Coast) using a machine learning approach based on boosted regression trees. Despite the class imbalance and class overlap in the dataset, their model demonstrated excellent predictive performance, highlighting that \textit{Physalia} occurrences are influenced by region-specific wind patterns and enhanced productivity. They also modelled the Australian East Coast data on a monthly time-scale rather than the daily scale reported by the Randwick council.

 \section{Methodology}

 \subsection{Data}

\begin{figure}[htbp]
    \centering
\includegraphics[width=0.45\textwidth]{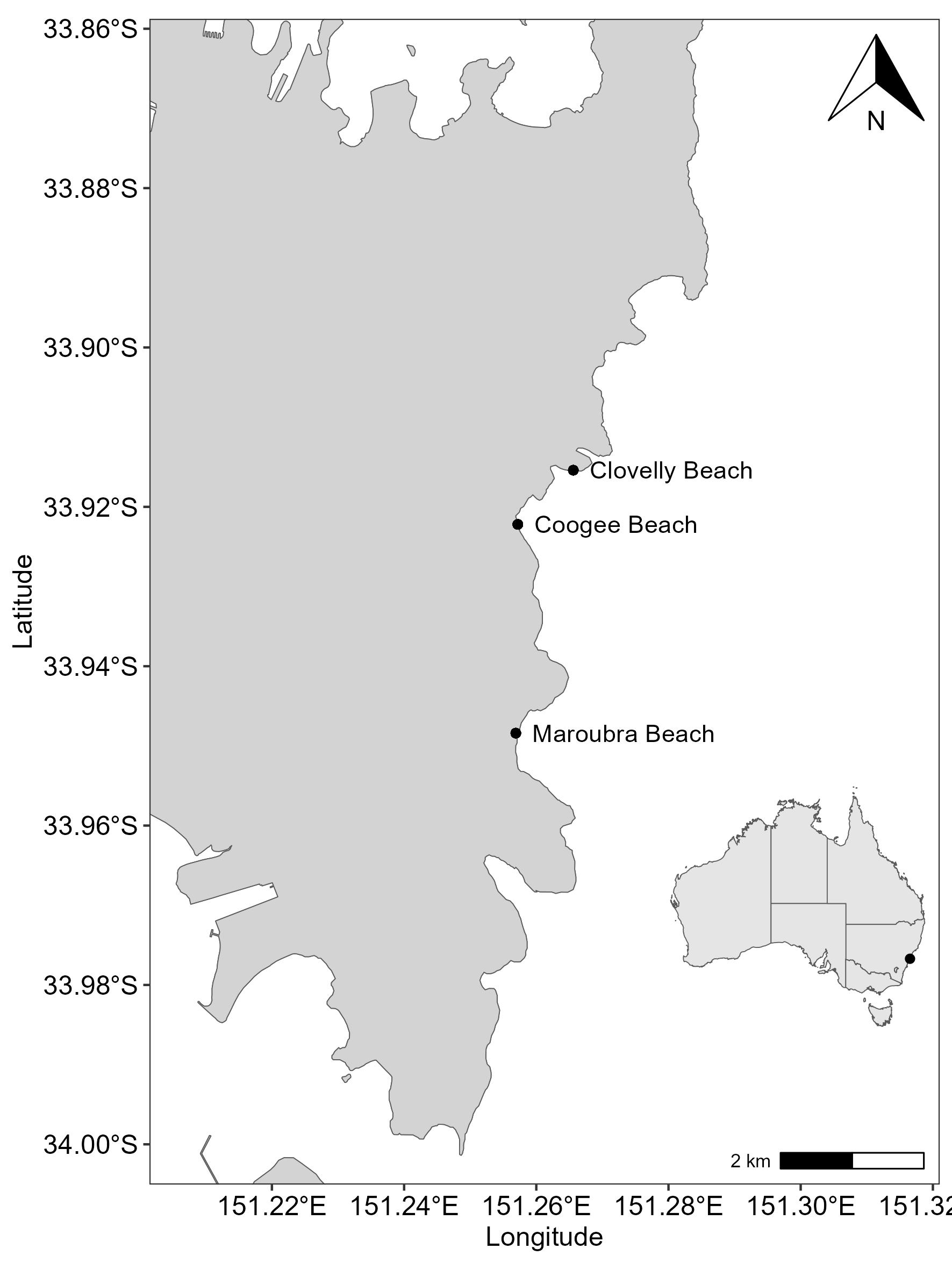}
    \caption{Map showing the spatial distribution of data points within the Randwick council area in the eastern Sydney, Australia. Each blue dot represents the beaches based on latitude and longitude coordinates, plotted using geospatial data. 
    }
    \label{fig:beachmap}
\end{figure}

This study utilizes data collected from the southeastern region of Sydney, Australia, focusing on three beaches: Maroubra, Clovelly, and Coogee used in \cite{bourg2022driving} and \citep{martins2024unravelling}. Maroubra, the longest and most exposed of the three, is located to the South. Further north, Coogee, is somehow protected by Wedding Cake Island, and Clovelly is nestled in a narrow bay (Figure \ref{fig:beachmap}).

The dataset is based on Bluebottle beaching events, which serve as indicators of Bluebottle occurrences (daily presence and absence). Beaching data were extracted from lifeguard reports provided by the respective local councils. The dataset spans four years, from March 2016 to June 2020, and includes daily observations. Maroubra is monitored year-round, with data available for all seasons from January to December. Similarly, Coogee has year-round data coverage, whereas Clovelly only includes observations from the warmer months (September to April) due to limited activity at the beach during winter \citep{bourg2022driving}.

Figure~\ref{fig:beachmap} illustrates the locations of the three beaches in Sydney. The daily observational bluebottle presence and absence dataset used in this study is available in the supplementary material from \citep{bourg2022driving}. 

In addition, environmental variables were extracted from publicly available gridded reanalysis products, corresponding to the model grid cell nearest to the beach where an observation was made. This includes wind speed (m s$^{-1}$) and direction ($^{\circ}$, where 0 = north and 180 = south) at 10 m height from the Australian Bureau of Meteorology Atmospheric high-resolution Regional Reanalysis for Australia (BARRA2; \cite{su2019barra}; 12 km resolution, averaged daily) and surface ocean current speed (m s$^{-1}$) and direction ($^{\circ}$, where 0 = north and 180 = south) and sea surface temperature (SST, $^{\circ}$C) from the Bluelink ReANalysis (BRAN2020; \cite{chamberlain2021next}; $\approx$ 10 km, daily).

\begin{figure*}
    
\includegraphics[width=0.9\linewidth]{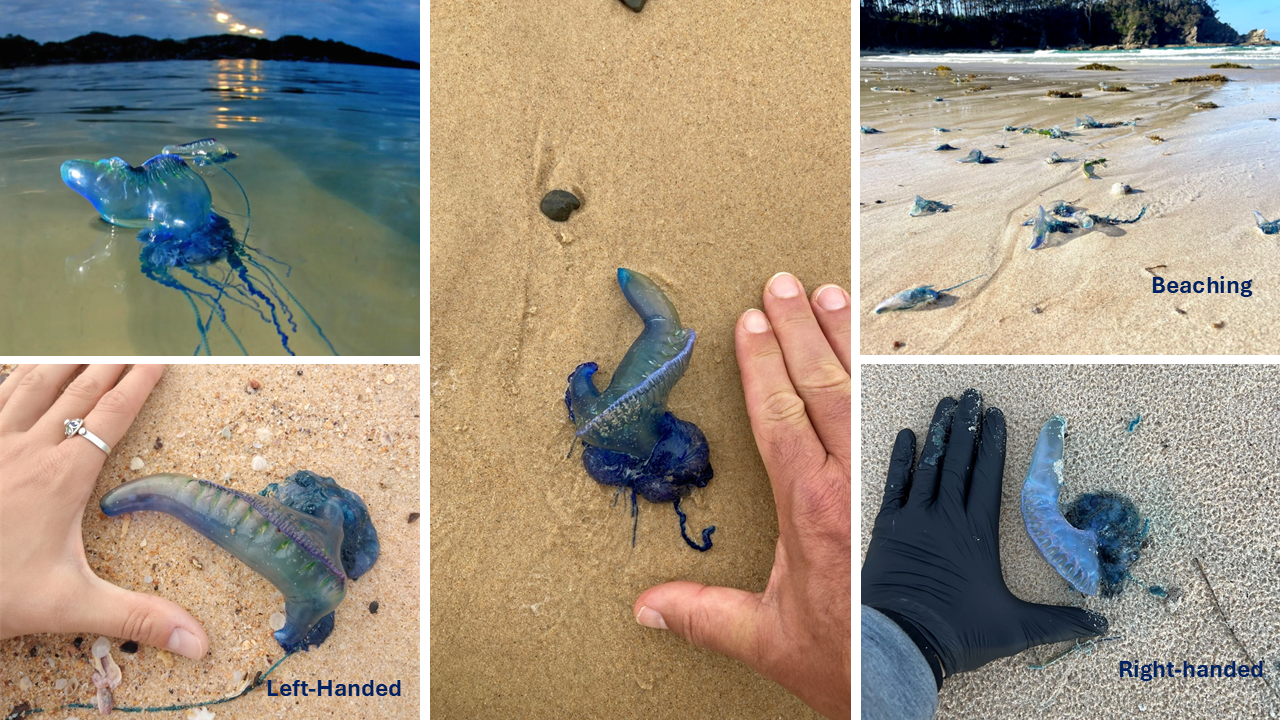}

    \caption{Bluebottle marine stringer (Physalia physalis) across different scenarios and morphological variations obtained from iNaturalist website \cite{bluebottles_australia}. The top-left panel shows bluebottle floating in water, illustrating its natural habitat. The top-right panel displays a beaching event, with multiple stranded bluebottles onshore. The bottom-left panel highlights a left-handed bluebottle, distinguished by the orientation of its sail. The bottom-right panel depicts a right-handed bluebottle, showcasing the opposite sail orientation.}
    \label{fig:Bluebottle_beaching}
    
\end{figure*}

Table \ref{tab:features_sub_features} presents the list of features in the raw and processed dataset, along with associated features and data type.

\begin{table*}[htbp]
    \centering
    \caption{The overview of the initial dataset features, their corresponding preprocessed features with units and the sub features or categories derived from them.}
    \resizebox{1.0\textwidth}{!}{
    \begin{tabular}{|c|c|c|}
    \hline
    \textbf{Initial Features} & \textbf{Preprocessed Features (units)} & \textbf{Sub-features} \\ \hline
Current direction and Wind direction & Wind direction (degrees)  & North, North-East, East, South-East, South, South-West, West, North-West \\  \hline

Current speed and Wind speed &Current direction (degrees) & North, North-East, East, South-East, South, South-West, West, North-West \\ \hline

SST, Date, Length and State &Wind Speed (m/s) & Low, Medium, High, Very High  \\ \hline

Beach, Beach key and Orientation &Current Speed (m/s) & Low, Medium, High, Very High \\ \hline

Surf club, Surf life saving Australia (SLSA) &SST (degrees Celsius) & Low, Medium, High, Very High  \\ \hline

Latitude, Longitude and Embaymentisation  &Month & January, February, March, April, May, June, July, August, September, October, November, December \\ \hline
    \end{tabular}
    }
    \label{tab:features_sub_features}
\end{table*}

 \subsection{ Data Augmentation}
 
\subsubsection{SMOTE}

SMOTE operates in the feature space rather than the data space, generating synthetic samples by oversampling each minority class instance. These synthetic samples are created along the line segments connecting the minority instance to one or more of its $k$-nearest neighbors, which are randomly selected based on the required level of oversampling \citep{chawla2002smote}. 
 
SMOTE is an iterative process that starts by randomly selecting a minority class instance from the training dataset. For the selected instance, the algorithm identifies its $k$-nearest neighbors, and $N$ of these $k$  neighbors are randomly chosen to generate new synthetic samples. The generation process involves interpolation, which is achieved by calculating the difference between the feature vector of the chosen minority instance and each selected neighbor.  

This process generates random points along the ``line segments'' connecting the selected minority instance and its neighbors. The procedure is repeated for various samples in the minority class until the desired number of synthetic samples is produced \citep{fernandez2018smote}. Studies have demonstrated that combining SMOTE with under-sampling techniques results in more robust performance compared to using SMOTE alone \citep{chawla2002smote, sharma2022smotified, junsomboon2017combining}.
 
\subsubsection{Generative Adversarial Networks (GAN)}

GANs  \citep{goodfellow2020generative} consist of two components: a generator network (GG) and a discriminator network (DD), which are trained simultaneously to produce synthetic data samples. The generator GG learns to approximate the data distribution, while the discriminator DD estimates the probability that a given sample originates from the real training data rather than from GG. Essentially, GG aims to generate samples indistinguishable from real data, while DD works adversarially to distinguish between generated samples and true data.

GANs can produce diverse variations of the original dataset, which can be leveraged to enhance model performance \citep{khan2023review}. They have demonstrated exceptional performance in applications such as computer vision (image generation) \citep{han2018gan} and text generation tasks \citep{liao2022text}. 
 
Unlike image and text data, tabular data often contains a mixture of discrete and continuous features, making it more complex and challenging to model and generate realistic synthetic data. Additional difficulties arise when discrete columns are imbalanced, and continuous columns exhibit multiple modes. To address these issues, \cite{xu2019modeling} developed the Conditional Tabular Generative Adversarial Network (CT-GAN), which employs a conditional generator to tackle these challenges effectively.

\cite{sauber2022use} applied CT-GAN to address class imbalance in tabular data and reported significant success in improving the balance of datasets. However, \cite{khan2023review} evaluated the combination of CT-GANs with ensemble learning and concluded that while effective, this combination does not outperform established approaches such as SMOTE, which are computationally more efficient.

In this study, we employ CT-GAN as a data augmentation method due to its suitability for tabular data.

 \subsection{Models}

We employed three different supervised machine learning models: Random Forest, XGBoost, and Multilayer Perceptron. Random Forest and XGBoost, both ensemble learning methods, have proven effective in addressing imbalanced datasets \citep{more2017review, zhang2022research}. Additionally, we included Multilayer Perceptron, a neural network model capable of handling non-linear data, selected based on studies by \cite{pontin2009factors} and \cite{castro2024using}, which demonstrated the effectiveness of artificial neural networks in predicting factors influencing the presence of jellyfish.

Alongside the supervised models, we also utilized an unsupervised learning model, the one-class Support Vector Machine (One-class SVM) \citep{muller2018introduction}. One-class SVM was selected to address the issue of unreliable negative data. To handle unreliable absence data effectively, we incorporated techniques proposed by \cite{haque2022negative} which involves training with one class while generating synthetic negative class for the model evaluation and \cite{zhang2020generative} utilizing CT-GAN to create a new set of negative class (absence) samples, which are then combined with the original positive class (presence).

\subsubsection{Simple neural network}
 
A simple neural network, also known as a Multilayer Perceptron (MLP), is inspired by the biological neurons in the brain and is one of the most widely used machine learning models \citep{zou2009overview, pontin2009factors}. MLP has demonstrated its versatility and flexibility as a function approximator for modeling diverse types of data \citep{lek1999artificial}, showing robustness in handling noisy and incomplete data \citep{pontin2009factors, castro2024using}.

MLP has found applications in various ecological contexts, including terrestrial and aquatic ecosystems \citep{lek1999artificial, quetglas2011use}, remote sensing for evaluating regional eco-environmental quality \citep{shi2007application}, evolutionary ecology \citep{borowiec2022deep, bryant2002use}, global climate change studies \citep{liu2010application}, and marine organisms such as jellyfish \citep{castro2024using, pontin2009factors}.

\begin{figure}[htbp!]
    \centering
    \begin{subfigure}{0.55\textwidth}
        \includegraphics[width=\textwidth]{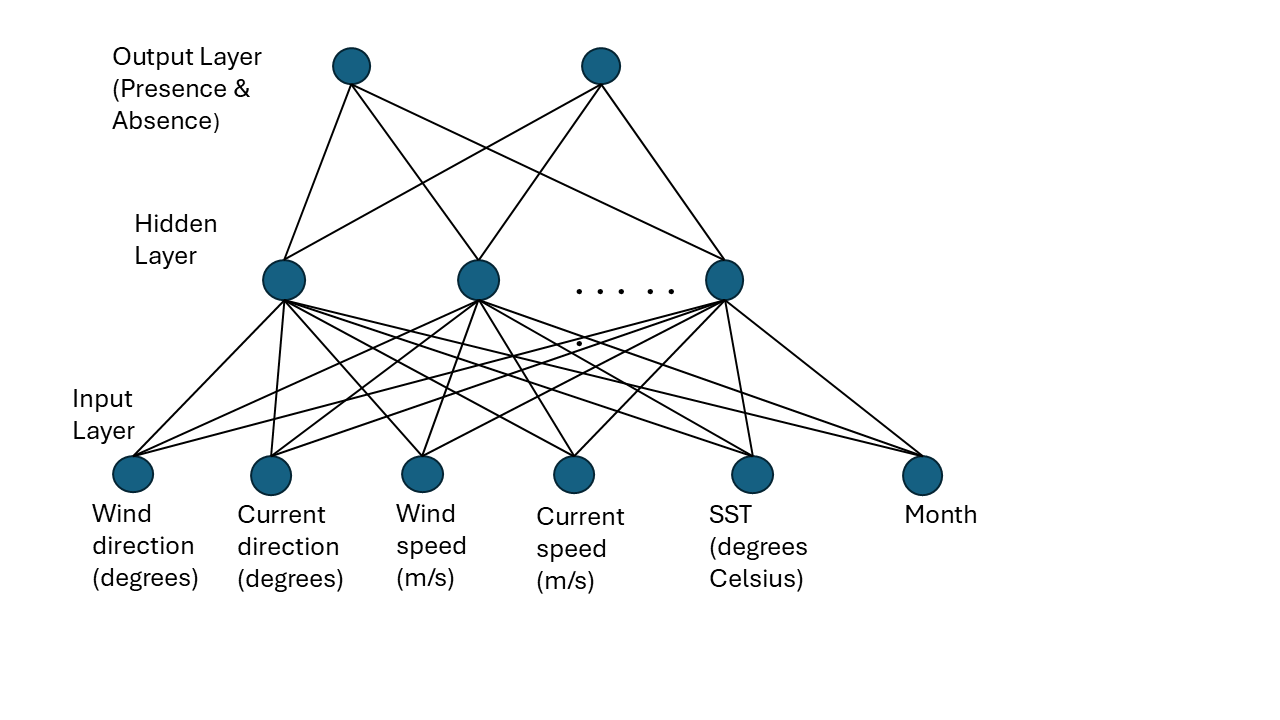} 
        \caption{A simple neural network with all the features represented as the input layers, \\hidden neurons, and output layer (bluebottle presence and absence).}
    \label{fig:Neural_network}
    \end{subfigure}
  
    \begin{subfigure}{0.55\textwidth}
        \includegraphics[width=\textwidth]{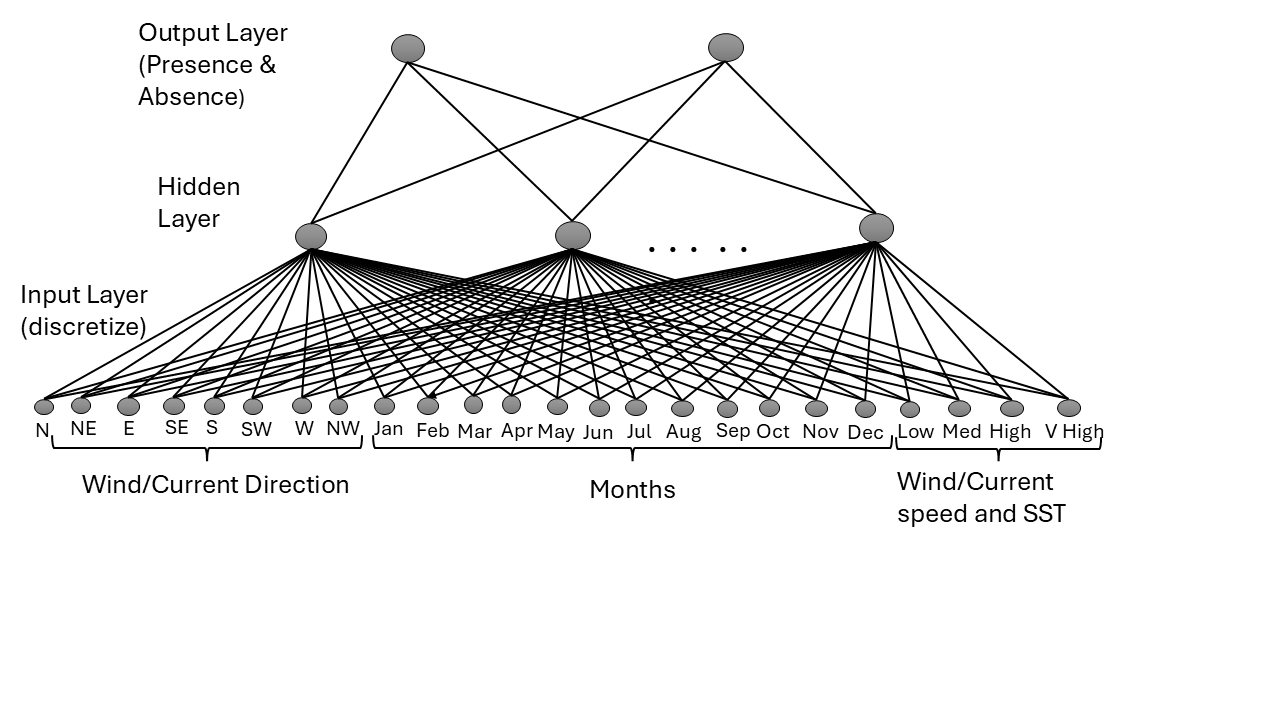} 

    \caption{The Neural Network architecture of subgroup features showing different \\categories of the sub-features}
        \label{fig:mlp_subgrouped}
    \end{subfigure}
    \caption{The Neural Network architecture of the entire features vs the subgroup features, where wind and current direction share similar sub-features, while SST, wind and current speed share similar sub-features. }
    \label{fig:mlp_framework}
\end{figure}

Figure~\ref{fig:mlp_framework} illustrates the implementation of MLP for detecting bluebottles based on data regarding their presence and absence. The features used in the model include both categorical and numerical variables. As depicted in Figure~\ref{fig:Neural_network} and Figure~\ref{fig:mlp_subgrouped}, the information flows from the input layer through multiple hidden layers before reaching the output layer, providing predictions based on the processed data.
   
\subsubsection{Random Forest}

Random Forest is a supervised ensemble learning model designed for both regression and classification tasks \citep{breiman2001random, more2017review}. It operates by constructing an ensemble of decision trees, each built using a random subset of data samples and features. For classification tasks, the model predicts the class based on the majority vote of the trees. The generalization error of a Random Forest is influenced by the strength of the individual trees and the correlation among them, while internal estimates allow for calculating feature importance \citep{breiman2001random}.

Random Forest is robust in handling missing values, noisy data, and high-dimensional datasets, making it effective for dimensionality reduction \citep{more2017review, breiman2001random}. It has been shown to outperform other classifiers in handling class imbalance, particularly when combined with balancing techniques. Misclassification rates can be further addressed using parallel ensemble methods \citep{more2017review}.

\cite{pal2005random} demonstrated that Random Forest requires fewer user-defined hyperparameters compared to classifiers like Support Vector Machines (SVM). It has also gained prominence in addressing class imbalance problems \citep{more2017review, khalilia2011predicting, khan2023review}.

The application of Random Forest in ecological and environmental studies is growing, as these fields often involve noisy data and numerous features \citep{cutler2007random, liu2014random, farahbakhsh2023prospectivity}. This makes Random Forest particularly well-suited for environmental and ecological tasks.

\subsubsection{Boosting using XGBoost}

Boosting is a family of ensemble learning models designed to transform weak learners into strong learners by iteratively improving model performance \citep{ferreira2012boosting, zhou2012ensemble}. In Boosting, models are added sequentially to the ensemble, with each new model correcting the errors of its predecessors. The final prediction is a weighted combination of the outputs from all models \citep{khan2023review}.

Gradient Boosting improves predictions by creating new models that focus on the residuals (errors) of previous models. These models are then combined to form the final output and can be applied to both regression and classification tasks. Extreme Gradient Boosting (XGBoost) builds on this framework by incorporating regularization in individual learners (trees) to prevent overfitting, offering a more advanced and versatile implementation of Gradient Boosting Machines \citep{ogunleye2019xgboost, chen2016xgboost}.

XGBoost achieves scalability through innovative methods such as an efficient tree learning algorithm designed for sparse data and a theoretically sound weighted quantile sketch method for handling instance weights in approximate tree learning \citep{chen2016xgboost}. However, XGBoost has limitations, including sensitivity to parameter selection and the need for careful fine-tuning to achieve optimal performance \citep{khan2023review}. Additionally, it provides insights into feature importance, allowing users to evaluate the contribution of each input feature to the model.

XGBoost has been successfully applied to various environmental and ecological challenges, including predicting particulate matter levels, flood risk, pollutant concentrations, and more \citep{pan2018application, ma2021xgboost, li2022application}.

\subsubsection{Random Undersampling}

Random Sampling is a data augmentation method that selects kk unique items from a population of nn items, ensuring that every possible combination of kk items has an equal probability of selection \citep{meng2013scalable}. Random Sampling Without Replacement selects a subset of elements from the population, giving each item an equal chance of inclusion while disallowing duplicates \citep{olken1993random}.

Random Undersampling is a technique used to balance target distributions by randomly eliminating instances from the majority class. Several undersampling strategies have been proposed, based on differing noise model theories. One theory suggests that instances near the classification margin between two classes are considered noise \citep{mohammed2020machine}, while another perspective classifies instances surrounded by neighbors with different labels as noise \citep{mohammed2020machine}.

Random Undersampling has been effectively applied to address class imbalance in various environmental and ecological modeling problems \citep{shin2021effects, kaur2018comparing}. In this study, we will employ Random Undersampling Without Replacement to reduce the size of the majority class.

\subsection{One-Class SVM}

The Support Vector Machine (SVM) \citep{muller2018introduction} is a highly effective non-parametric technique rooted in statistical learning theory, widely used in machine learning for classification and regression tasks \citep{hejazi2013one}. SVM is typically a two-class model that incorporates both negative and positive samples to achieve high predictive accuracy while minimizing generalization error in detection and classification tasks \citep{manevitz2001one, hejazi2013one}.

The One-Class SVM, an unsupervised learning extension of SVM, was proposed by Muller et al. \citep{muller2018introduction} and is frequently applied for anomaly and outlier detection. It operates by mapping training data from the input space into a higher-dimensional feature space using a kernel function. In this feature space, it identifies a hyperplane that maximizes the margin between the mapped data points and the origin \citep{hejazi2013one}.

There are two primary approaches for determining the decision boundary in One-Class SVM:
\begin{enumerate}
\item \textbf{Hypersphere Approach:} This defines a hypersphere that separates outliers from the positive class. The shape of the decision boundary is controlled by the $v$-parameter, which balances the trade-off between the proportion of data points classified as positive and as outliers \citep{hejazi2013one}.

\item \textbf{Hyperplane Approach:} This approach calculates a hyperplane in the feature space that ensures a specified fraction of training instances fall beyond it, while simultaneously maximizing the margin from the origin \citep{muller2018introduction}. Due to its simplicity, this approach has become the most commonly used \citep{hejazi2013one}.
\end{enumerate}
One-Class SVM has been successfully applied in various fields, including medical and environmental studies, among others \citep{haque2022negative, chen2017application}. In this study, we will employ the One-Class SVM to address the no-negative (absence) data challenge.


\subsection{Dimensionality reduction for visualisation}

Principal Component Analysis (PCA) \citep{pearson1901liii} is a statistical technique for dimensionality reduction that identifies a hyperplane minimizing the sum of squared perpendicular distances from a set of points in a multidimensional space. PCA determines the directions (principal axes) that capture the maximum variance in the data, allowing for dimensionality reduction while retaining the essential structure of the data \citep{pearson1901liii}.

This technique linearly transforms a multivariate dataset into a set of uncorrelated variables (principal components), ordered by the amount of variance each captures \citep{metsalu2015clustvis}. PCA is often used for data visualization by generating scatter plots of the first two principal components, which represent the directions accounting for the largest variance in the data \citep{metsalu2015clustvis}.

Projecting data onto the first two principal components allows for visual evaluation of class separation or dominant hidden patterns. A distinct separation in the scatter plot indicates that the features are effective for distinguishing classes in classification tasks \citep{abdi2010principal, begam2014visualization}. For example,
\cite{chandra2023unsupervised} used PCA to differentiate major COVID-19 variants of concern from genome data, demonstrating its utility in uncovering meaningful patterns.

Studies across various domains highlight the effectiveness of PCA.
\cite{begam2014visualization} applied PCA to chemical datasets, identifying dominant patterns of drug-likeness through visualization, which provided valuable insights into the underlying characteristics of the data. Similarly, 
\cite{azhar2015classification} used PCA for the classification of river water quality, uncovering two distinct clusters that reflected differences in water quality characteristics, thereby aiding in understanding the patterns within the data.

PCA has also been widely used in combination with other methods. It has been paired with clustering techniques for unsupervised learning \citep{chandra2023unsupervised} and integrated with machine learning models for classification and regression tasks, often improving generalization capabilities \citep{howley2005effect}. By reducing dimensionality, PCA generates a more compact dataset that is less noisy, leading to smaller, more efficient models \citep{uddin2021pca}.

\subsection{Framework}

\begin{figure*}[htbp]
    \centering
    \includegraphics[width=0.75\textwidth]{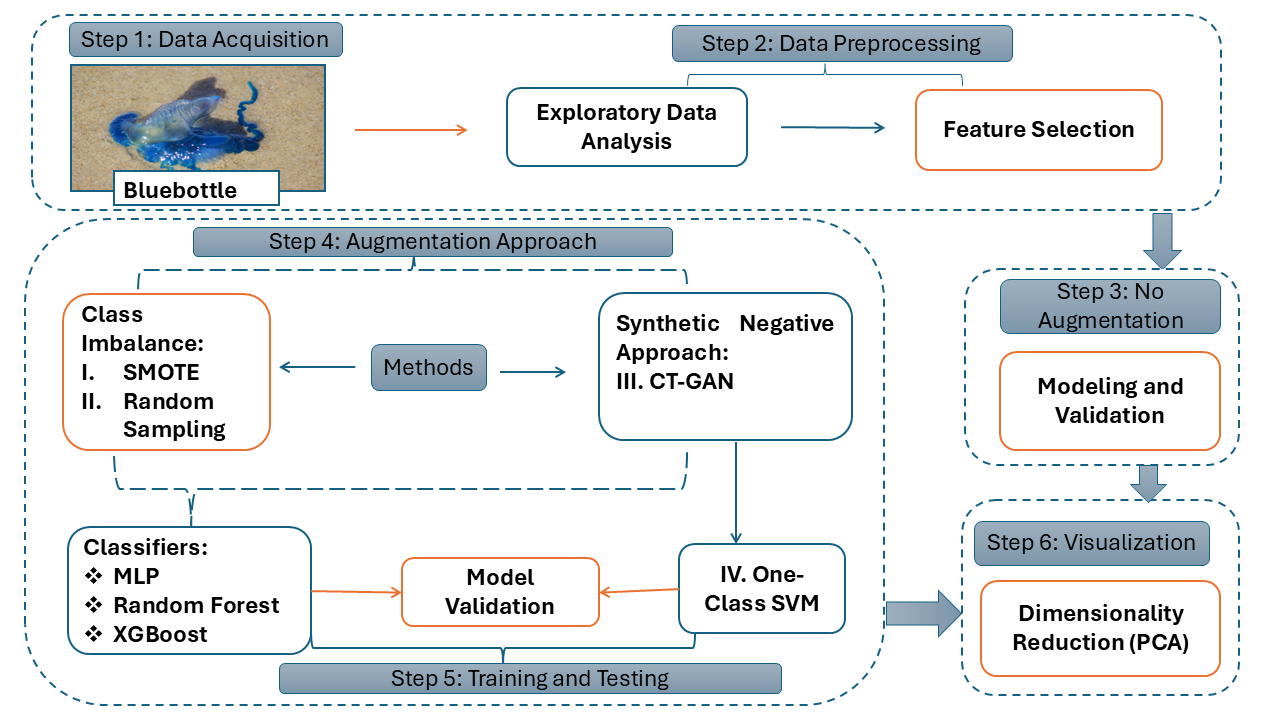}
    \caption{A framework for bluebottle modelling incorporating data preprocessing, augmentation techniques and model validation. This framework consists of six key steps including (1) data acquisition, (2) data preprocessing including exploratory analysis and feature selection, (3) baseline modelling without augmentation, (4) augmentation approach using both class imbalance approach and synthetic negative approach, (5) model training and validation, and (6) dimensionality reduction using PCA for visualisation of the different approaches implemented}
    \label{fig:framework}
\end{figure*}

We now present a machine learning framework addressing bluebottle data as a case of unreliable negative data and class imbalance, focusing on environmental factors influencing beaching events (Figure \ref{fig:framework}).

In Step 1, data is extracted from council lifeguard reports of the selected beaches. This is followed by data cleaning and preprocessing, including removing outliers and duplicates, handling missing values, and addressing data skewness.

In Step 2, exploratory data analysis is conducted using plots such as correlation matrices, density plots, seasonality trends, and line plots to examine class imbalance and data patterns. Feature selection is performed using a correlation matrix to identify the most relevant predictors for modeling.

Location-specific variables (e.g., beach key, embaymentisation, orientation, latitude, longitude, and surf club) are excluded to ensure that the findings are generalizable to the region and not restricted to the dataset's specific locations.

In Step 3, we focus on modeling the data and identifying the most suitable machine learning models for handling class imbalance. The dataset is split into training and testing sets, and models such as Multilayer Perceptron (MLP), Random Forest, and XGBoost are evaluated. Performance metrics, including precision, recall, and F1-score, are reported for both the majority and minority class labels to ensure a comprehensive assessment of each model's effectiveness in addressing class imbalance.

In Step 4, we implement data augmentation techniques to address class imbalance, data overlap, and unreliable negative data (absence of bluebottles) by assuming the absence of a definitive negative class.

We first evaluate two data augmentation methods for handling class imbalance: (i) SMOTE and (ii) Random Sampling. Random Undersampling without replacement is applied to balance the dataset by randomly selecting a subset of the negative class to match the size of the positive class, ensuring equal representation. This process addresses both class imbalance and unreliable negative data, as illustrated in Figure \ref{fig:framework}.

For the Synthetic Negative Approach, we assume no reliable negative data exists and use (iii) CT-GAN to generate the entire negative class. This method creates a synthetic negative dataset of the same size as the positive class, while preserving the original negative class distribution. This approach helps mitigate issues associated with unreliable negative data.

\cite{castro2024using} emphasized that the recorded absence of jellyfish does not always accurately indicate their actual absence, while 
\cite{rider2013classifier} highlighted biases arising from poorly defined negative classes and the challenges faced by conventional classification models. To address these concerns, we treat absence data as unreliable and employ synthetic data generation strategies to overcome these limitations.

In Step 5, we focus on model training and validation using the processed and augmented data (SMOTE, CT-GAN, and Random Undersampling) from Steps~2 and 4. The selected machine learning models—Multilayer Perceptron (MLP), Random Forest, and XGBoost—are trained and evaluated using both the original and synthetic negative data, encompassing both negative and positive classes.

Additionally, we implement the (iv) One-Class SVM as an alternative approach to handle unreliable negative data. This method assumes the absence of a reliable negative class and involves training the model exclusively on the positive class. Validation is performed using both the positive class and a negative class, where the negative samples are derived from CT-GAN-generated synthetic data.

By applying these approaches, we systematically compare the performance of models trained on original versus synthetic data, ensuring robustness in addressing the challenges posed by unreliable negative data.

In Step 6, we apply dimensionality reduction using PCA to visualize the class boundaries of the data. A scatter plot is created using the two principal components for both the original dataset and datasets generated through different augmentation techniques. This visualization helps to better understand the class separation and the prediction results, providing insights into potential strategies for further improving model accuracy.

Additionally, PCA-based visualization is used to evaluate and compare the effectiveness of the various data augmentation strategies. Classification metrics are reported alongside the visualizations to assess the impact of these strategies on model performance. This step ensures a comprehensive understanding of the augmented datasets and their influence on the classification tasks.


\subsubsection{Evaluation Metrics}

Evaluation metrics are essential for designing and assessing machine learning models \citep{gu2009evaluation}, particularly when addressing class imbalance problems. A comprehensive set of metrics is required to ensure the model's performance is evaluated fairly across both majority and minority classes, capturing its effectiveness in imbalanced scenarios.

The most commonly used metrics for binary classification include classification accuracy, precision, recall, F1-score, and AUC. Classification accuracy evaluates the model's effectiveness by measuring the percentage of correctly classified samples. While simple to implement, it is not suitable for highly imbalanced datasets \citep{bekkar2013evaluation}. This is because classification accuracy often fails to reflect the model's true performance, particularly for the minority class, where it can be misleading due to the dominance of the majority class in the dataset. 

To accurately evaluate model performance, we must consider four key metrics: True Positives (TP): The number of positive samples correctly identified as positive; True Negatives (TN): The number of negative samples correctly identified as negative; False Positives (FP): The number of negative samples incorrectly identified as positive; False Negatives (FN): The number of positive samples misclassified as negative.

A confusion matrix provides a structured way to report these counts, offering a clearer understanding of the relationship between the actual and predicted classes \citep{sokolova2009systematic}. It is a fundamental tool for assessing classification performance, especially in imbalanced datasets.
 
Other metrics are more effective for addressing class imbalance problems. Precision reflects the correctness of a model by measuring the proportion of predicted positive instances that are true positives. It is defined as:
$$
\text{Precision} = \frac{\text{True Positives (TP)}}{\text{True Positives (TP)+False Positives (FP)}}.
$$
Recall, on the other hand, evaluates the model's completeness by assessing the proportion of actual positive instances that are correctly identified. It is defined as:
$$
\text{Recall} = \frac{\text{True Positives (TP)}}{\text{True Positives (TP)+False Negatives (FN)}}.
$$

Unlike classification accuracy, Precision and Recall are less sensitive to data distribution changes, making them more robust for imbalanced datasets. They are particularly useful for detecting imbalances between two classes and providing a clearer understanding of a model's performance in such scenarios.


The F1 score is the harmonic mean of Precision and Recall, providing a single metric that balances both correctness (Precision) and completeness (Recall). It is particularly useful in cases where there is class imbalance, as it ensures that both metrics are considered equally. A high F1 score indicates that the model performs well in identifying the positive class while maintaining a balance between Precision and Recall.


The Receiver Operating Characteristic (ROC) curve provides a visual representation of a model's performance across varying threshold values. It plots the true positive rate (TPR) against the false positive rate (FPR), illustrating the trade-off between sensitivity and specificity for the dataset.

The Area Under the Curve (AUC) quantifies the ROC curve's performance, serving as a summary measure of the model's ability to distinguish between classes. A higher AUC indicates a better model, signifying improved accuracy in correctly predicting and separating the classes \citep{bekkar2013evaluation, khan2023review}.


\subsubsection{Technical and implementation details}

We utilize Python libraries such as NumPy, Pandas, and scikit-learn for data processing, analysis, and modeling, with the code available in our GitHub repository \footnote{\url{https://github.com/DARE-ML/bluebottle-xai}}. For all machine learning models used in this study (MLP, Random Forest, XGBoost), we conduct 30 independent experimental runs. Each run involves randomly assigned parameters (e.g., weights and biases in MLP) and employs a randomly shuffled training/test data split, with 60\% of the data allocated for training and 40\% for testing.

Our data was further processed where we subgrounded certain features so that we get a better understanding of the cause of bluebottle beaching. Table \ref{tab:features_sub_features} presents detail of the original feature and the subgroups of the feature.

We selected model hyperparameters from the literature and trial experimental runs for the respective models. In the case of MLP, we utilised 100 hidden neurons and two output neurons, with 6 input neurons for each feature. We used the Adam optimiser with a maximum of 400 epochs and a learning rate initialisation of 0.1 and alpha of 0.0001.  

We configured the XGBoost classifier with a learning rate of 0.1, 100 trees, and a scale positive weight of 3. The model was trained without label encoding, using logarithmic loss (logloss) as the evaluation metric and a maximum tree depth of 2. This configuration was chosen to balance model complexity and performance while addressing class imbalance effectively.

The Random Forest model was used with its default parameters for modeling without applying any data augmentation techniques. This baseline configuration provides a reference for assessing the impact of augmentation and other parameter adjustments on model performance.

In the data augmentation scenario, specific hyperparameter configurations were applied to optimize the models. For SMOTE, we used SMOTENC \citep{mukherjee2021smote}, which is suitable for datasets with a mix of categorical and continuous features. The sampling strategy was set to ``minority'', and the random state was specified as ``run''.

For the One-Class SVM, the gamma parameter was set to ``auto'', and the kernel was configured as ``rbf'' to effectively capture nonlinear relationships. The CT-GAN model was implemented using its default hyperparameters.

The Random Forest model underwent significant hyperparameter adjustments. Bootstrap sampling was disabled (\texttt{bootstrap = False}), and the maximum depth of the trees was set to 7. A minimum of 10 samples was required to split an internal node, and the number of features considered at each split was defined as the logarithm base 2 of the total number of features. The model was trained with 150 trees, with each leaf requiring a minimum of 2 samples and a maximum of 20 leaf nodes. Additionally, the complexity parameter for minimal cost-complexity pruning was set to 0.01. These configurations were tailored to improve performance and address challenges associated with the augmented data.

\section{Results}

\subsection{Data Analysis}

This stage focuses on exploratory data analysis (EDA) and feature selection. The dataset comprises sixteen features represented in the first column of Table~\ref{tab:features_sub_features}. The beach name, council report, beach key-identification, surf club, SLSA, state, latitude, longitude, length, orientation, embaymentisation were removed from the modelling analysis as explained in Step 2 of the framework diagram. After preprocessing the data, we have six input features for the modelling stage, detailed in the second column. The final column presents the transformation of the features into categorized or processed variables named sub-features.  The target variable is the presence/absence of the Bluebottle (\textit{physalia})found at a particular location. 


\begin{table}[htbp]
    \centering
     \caption{A summary of the value ranges for sub-features SST, Current Speed, and Wind Speed, categorised into Low, Medium, High and Very High.}
     \resizebox{0.45\textwidth}{!}{
    \begin{tabular}{|c|c|c|c|}
    \hline
\textbf{Sub-features}& \textbf{SST [$^oC$]} & \textbf{Current Speed [m s$^{-1}$]} & \textbf{Wind Speed [m s$^{-1}$]}  \\ \hline

Low   & 13.638 - 16.773 & 0.007 - 0.118 & 0.547 - 1.109 \\ \hline

Medium & 16.773 - 19.909 & 0.118 - 0.243 & 1.109 - 1.671 \\ \hline

High & 19.909 - 23.045 & 0.243 - 0.368 & 1.671 - 2.232 \\ \hline

Very High & 23.045 - 26.180 & 0.368 - 0.493 & 2.232 - 2.794 \\ \hline
    \end{tabular}
        }
    \label{tab:discritize_features}
\end{table}

Table \ref{tab:discritize_features} summarizes the value ranges for three features: sea surface temperature (SST), current speed, and wind speed, which are categorized into four levels: low, medium, high and very high. This categorization facilitates the discretization of continuous data into meaningful levels for further analysis and modelling. Additionally, the wind direction (where the wind comes from) and current direction (where the current comes from) are continuous variables, with their categories defined in Table \ref{tab:features_sub_features}. These directions are represented in angular ranges as follows: North ($0\,^\circ$, $360\,^\circ$), North-East($45\,^\circ$), East ($90\,^\circ$), South-East ($135\,^\circ$), South ($180\,^\circ$), South-West ($225\,^\circ$), West ($270\,^\circ$) and North-West ($315\,^\circ$). Table \ref{tab:descriptive_stat} presents the total number of bluebottle presence and absence instances according beaches, along with key feature statistics (mean and standard deviation) for each location.

\begin{table*}[htbp]
    \centering
    \caption{The total number of presence and absence per beaches as well as the mean and standard deviation (SD) of the features with respect to each beaches.}
    \resizebox{0.9\textwidth}{!}{
    \begin{tabular}{|l|c|c|c|c|c|c|c|}  \hline
\textbf{Beaches}& \textbf{Total Presence} & \textbf{Total Absence} & \textbf{SST(mean, SD)} & \textbf{Wind direction(mean, SD)} & \textbf{current direction(mean, SD)}& \textbf{Current Speed(mean, SD)} & \textbf{Wind Speed(mean, SD)}  \\ \hline

Clovelly & 39 & 803 & 21.910 (1.805) & 117.226 (90.439)& 106.259 (127.235) & 0.218 (0.154) & 5.225(2.615) \\ \hline

Coogee & 80 & 1446 & 21.017 (2.103) & 161.849 (106.837) & 96.597 (119.755) & 0.215 (0.153) & 5.367 (2.773)\\ \hline

Maroubra & 126 & 1357 & 21.017 (2.106) & 161.507 (106.814) & 95.265 (118.339) & 0.215 (0.154) & 5.766 (2.974) \\ \hline
    \end{tabular}
    }
    \label{tab:descriptive_stat}
\end{table*}

\begin{figure*}[htbp]
    \centering
    \includegraphics[width=0.65\textwidth]{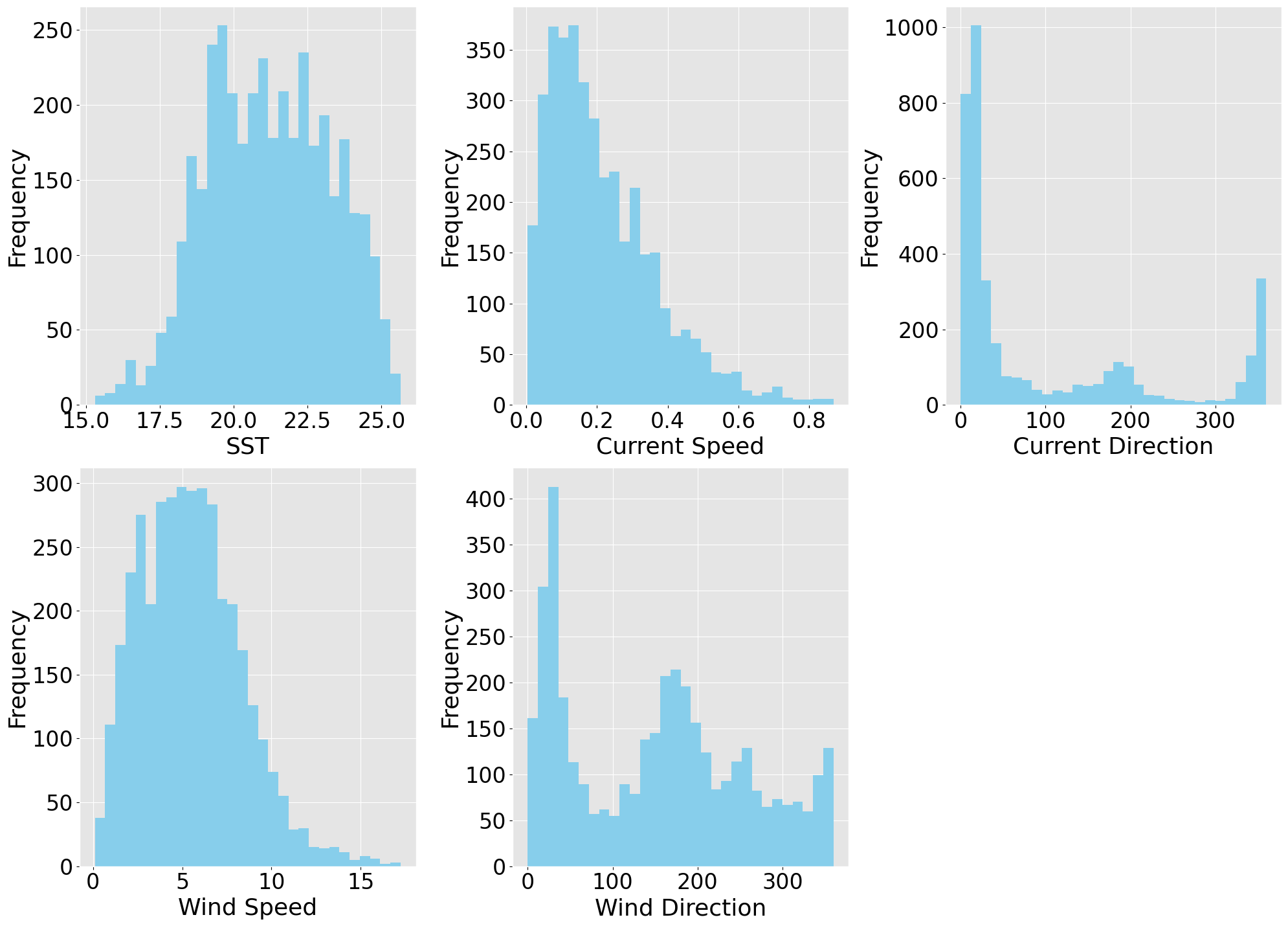}
    \caption{Distribution of the continuous features for bluebottle presence. We notice that the SST exhibits an approximately bell-shaped curve, current speed and wind speed are right-skewed, while current direction and wind direction display bimodal distribution.}
    \label{fig:histogram_plot}
\end{figure*}

Figure \ref{fig:histogram_plot} presents the distribution of the continuous features of the bluebottle presence/absence. We observe that the SST is almost bell-shaped, with most values clustered between $18\,^\circ \text{C}$ and $24\,^\circ \text{C}$. This indicates that the majority of the data falls within a moderate sea surface temperature range, with a slight skew towards lower temperatures. The current speed and wind speed are right-skewed (positively skewed), with most values of Current speed concentrating at the lower end (0 to 0.4m/s) with fewer observations at higher current speeds while most wind speed falls between 2.5 to 10m/s. The current direction and wind direction have a bimodal distribution because their direction is circular (circular variables; \cite{lee2010circular}).  
For the current direction, there is a significant peak at around $0\,^\circ$ and $360\,^\circ$ (north) and a smaller peak at the west. This suggests that the current flows are predominantly aligned along two major directions, with north and west being dominant. A current from the north is typical from the local oceanography (East Australian Current, \cite{bourg2022driving}) The distribution of the wind direction is somewhat uniform across several ranges, with peaks at around $0\,^\circ$ and $360\,^\circ$ (northerly winds, consistent with the local sea breeze), and $180\,^\circ$ (southerly winds). Similar to the current direction, this indicates wind direction is spread over a wide range but with some dominant directions.

\begin{figure}[htbp!]
    \centering
      \begin{subfigure}{0.45\textwidth}
      \includegraphics[width=\linewidth]{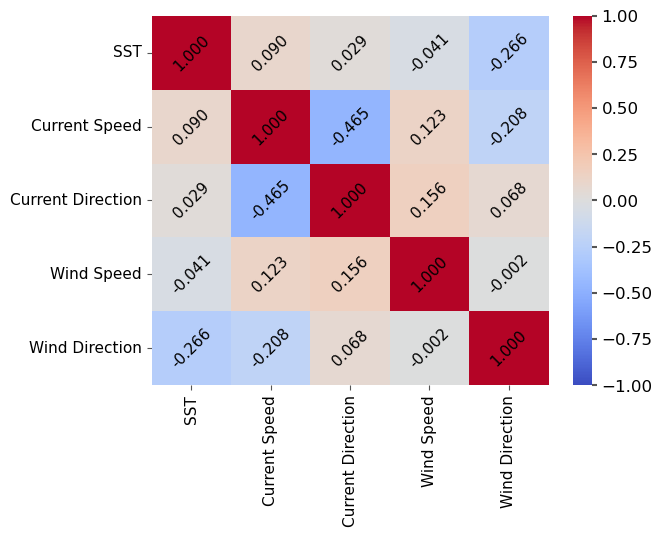} 
        \caption{Correlation matrix for continuous features: blue shades indicate negative correlations, while red shades represent positive correlations.}
        \label{fig:corr_matrix}
     \end{subfigure}
\begin{subfigure}{0.45\textwidth}
        \includegraphics[width=\linewidth]{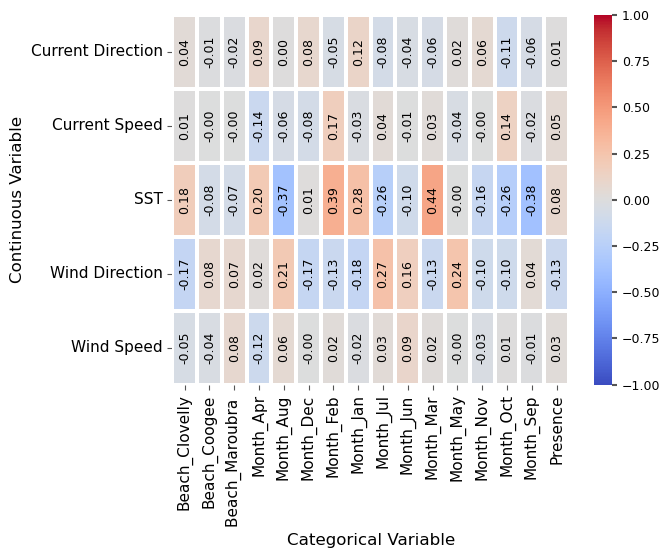}
        \caption{Correlation matrix for continuous features, sub-grouped as separate features. The colour scale represents the strength of correlations, ranging from -1 (strong negative correlation) to +1 (strong positive correlation).}        
\label{fig:contin_cat_matrix}
   \end{subfigure}
   \caption{The correlation plot of the continuous and categorical features. The top plot is for the continuous while bottom is for the nominal features}
\end{figure}

  Figure \ref{fig:corr_matrix} displays the correlation matrix of the continuous variables that indicates that current speed and current direction have a negative correction. Additionally, wind direction shows a negative correlation with SST and with current speed (the ocean temperature is warmer with southward flowing East Australian CUrrent, and cooler in response to wind-induced mixing). Overall, the continuous variables do not exhibit strong correlations with one another, suggesting that all of them are suitable for modelling, as there is no evidence of multicollinearity among them.

Figure \ref{fig:contin_cat_matrix} shows the correlation between the categorical and continuous variables using point biserial correlation \citep{tate1954correlation}. We utilise Pearson's product-moment correlation when one variable is dichotomous, with only two possible values (typically coded as 0 and 1), and other variables are continuous \citep{kornbrot2014point}. We can observe that SST has moderate positive correlations with January to April (warm months) and negative correlations with August to October (cool months). The wind direction is positively correlated with May, July and August. 



\begin{figure}[htbp!]
    \centering
      \begin{subfigure}{0.45\textwidth}
      \includegraphics[width=\linewidth]{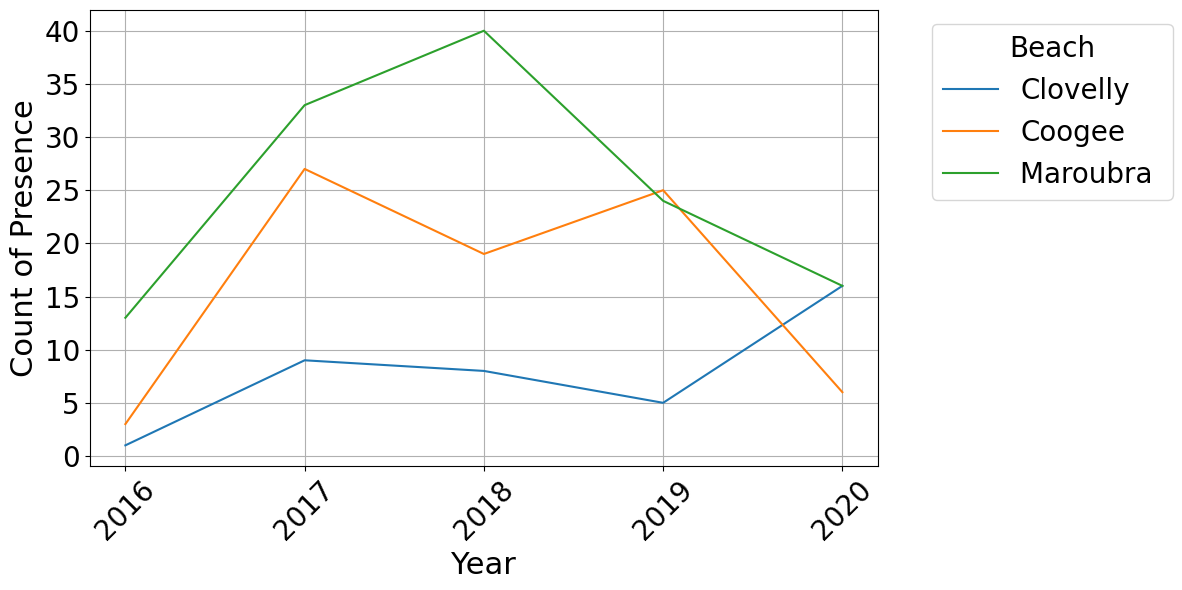} 
        \caption{Distribution of bluebottle presence across three beaches: Clovelly, Coogee, and Maroubra, from 2016 to 2020.}
        \label{fig:year_presence}
     \end{subfigure}
\begin{subfigure}{0.45\textwidth}
        \includegraphics[width=\linewidth]{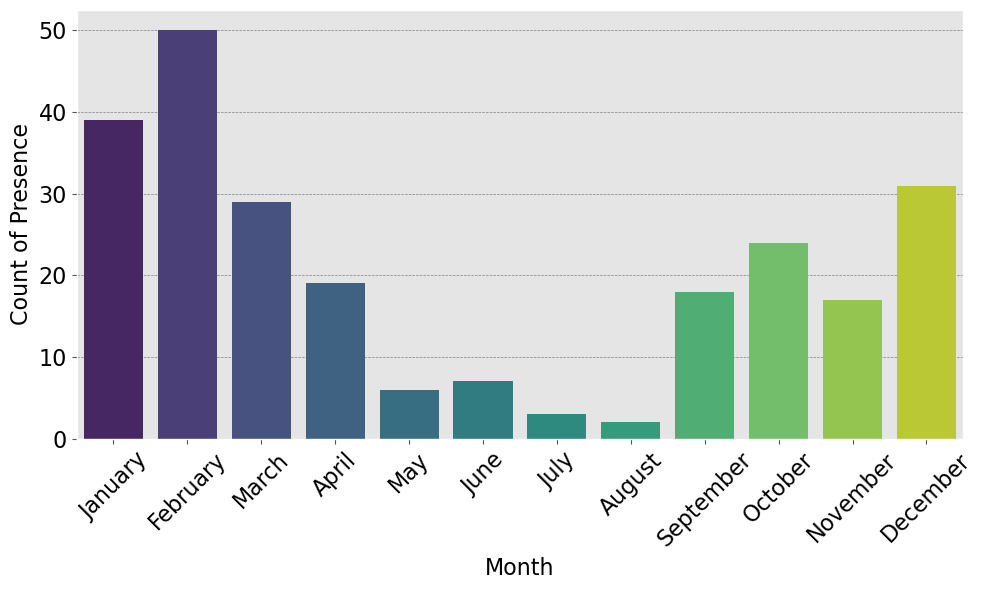}
        \caption{Monthly distribution of bluebottle presence, highlighting peak occurrences}        
        \label{fig:seasonal}
   \end{subfigure}
   \caption{The yearly and Monthly Distribution of Bluebottle Occurrence across different beaches}
\end{figure}

Figure~\ref{fig:year_presence}   illustrates the yearly temporal distribution of bluebottle occurrences across three beaches (Clovelly, Coogee and Maroubra) - from 2016 to 2020. Note that Clovelly is only sampled from September to April. 

Figure~\ref{fig:seasonal} illustrates the monthly distribution of bluebottle presence. The plot indicates that bluebottle beaching is more frequently observed from September to April, with the highest occurrences between December and February, peaking in February. This pattern highlights that bluebottle presence is seasonal, peaking in the austral summer, with notable occurrences in spring and autumn, but very few during winter. 

\subsection{Modelling and Evaluation}


We first modelled the dataset with three different classifiers, where we used the presence and absence data from the original data with a class imbalance   (6 percent positive and 94 percent negative). We utilise MLP, XGBoost, and Random Forest without any data augmentation to the class imbalance, to see how the classifiers performed on the dataset. After that, we implemented different data augmentation techniques, including  SMOTE and Random Undersampling for class imbalance and CT-GAN for no negative class. We carry out 30 individual model training experimental runs for each model and report the mean and standard deviation using selected scores (accuracy, F1, and AUC).

\subsubsection{Results without Data Augmentation}


\begin{table*}[htbp!]
  \centering
  \begin{minipage}[b]{1.0\linewidth}
    \centering
    \caption{Results for bluebottle prediction using classification models without data augmentation for both train and test. We report the mean score from 30 experimental runs, with the standard deviation (in brackets) The F1 score on the test data is highlighted in bold to emphasize the model'ss performance on unseen data}
    \resizebox{1.0\textwidth}{!}{
    \begin{tabular}{|c|c|c|c|c|c|c|}
    \hline
    \multirow{2}{*}{\textbf{Models}} & \multicolumn{3}{c|}{\textbf{Train }} & \multicolumn{3}{c|}{\textbf{Test }} \\
    \cline{2-7}
     & \textbf{Accuracy} (Mean, Std) & \textbf{F1 Score} (Mean, Std) & \textbf{AUC} (Mean, Std) & \textbf{Accuracy} (Mean, Std) & \textbf{F1 Score} (Mean, Std) & \textbf{AUC} (Mean, Std) \\
    \hline
    MLP &0.936(0.003) &0.008(0.023) &0.693(0.062) &0.935(0.005) &\textbf{0.003(0.009)}&0.639(0.034) \\ \hline
    Random Forest &0.986(0.002) &0.879(0.016) &0.998(0.000) & 0.914(0.005)& \textbf{0.109(0.029)} & 0.669(0.026) \\ \hline
    XGBoost  &0.940(0.004) & 0.243(0.058)& 0.876(0.011)&0.931(0.002) & \textbf{0.110(0.032)}& 0.735(0.021) \\ \hline
    \end{tabular}
    }
    \label{tab:metrics_noaugmentation}
  \end{minipage}%
  
  \hspace{2.0cm}
  \begin{minipage}[b]{1.0\linewidth}
    \centering
    \caption{Results from the best experimental run in Table \ref{tab:metrics_noaugmentation} showing Absence and Presence for training and test data. The precision, Recall, F1 score of the test data for both Absence and Presence are all in bold.}
    \resizebox{1.0\textwidth}{!}{
    \begin{tabular}{l|c|c|c|c|c|c|c|c|c|c|c|c}       \toprule
    \multirow{2}{*}{\textbf{Models}} & \multicolumn{3}{c|}{\textbf{{Train Class - Absence}}} & \multicolumn{3}{c|}{\textbf{{Train Class - Presence}}} & \multicolumn{3}{c|}{\textbf{{Test Class - Absence}}} & \multicolumn{3}{c@{}}{\textbf{{Test Class - Presence}}} \\ 
    \cmidrule(lr){2-4} \cmidrule(lr){5-7} \cmidrule(lr){8-10} \cmidrule(lr){11-13}
    & \textbf{Precision} & \textbf{Recall} & \textbf{F1 Score} & \textbf{Precision} & \textbf{Recall} & \textbf{F1 Score} & \textbf{Precision} & \textbf{Recall} & \textbf{F1 Score} & \textbf{Precision} & \textbf{Recall} & \textbf{F1 Score} \\ 
    \midrule
    MLP  & 0.940 & 1.000 &  0.969& 1.000 & 0.016 & 0.031 & 0.931& 0.999& 0.964& \textbf{0.000}& \textbf{0.000} & \textbf{0.000} \\ \hline
    Random Forest & 0.988& 0.997 & 0.992 & 0.945& 0.805& 0.869& 0.934& 0.970& 0.952& \textbf{0.170}&\textbf{0.083} & \textbf{0.111}\\ \hline
    XGBoost &0.945 &0.994  & 0.969& 0.542& 0.102& 0.171& 0.933& 0.998&0.965 & \textbf{0.571}&  \textbf{0.041} & \textbf{0.077}\\ \bottomrule
    \end{tabular}
    }  \label{tab:classification_metrics_noaugmentation}
  \end{minipage}
\end{table*}
Table \ref{tab:metrics_noaugmentation} presents the results  for the effect of class imbalance and class overlap within the dataset, showing the accuracy, F1 score and AUC. We observe that MLP achieved high accuracy on both training (0.936) and test (0.935) sets. The Random Forest and XGBoost models demonstrate higher accuracy and AUC scores for both training and test sets when compared to MLP.  The problem is visible only when we review the F1 score, which is the main metric for class imbalanced problems and applies to this problem.  However, the F1 scores are almost negligible for MLP on both training (0.008) and test (0.003), indicating poor performance in capturing the minority class (bluebottle presence).   This indicates that MLP has not been effectively capturing the minority class, possibly due to class imbalance and class overlap.  We notice that the F1 score for Random Forest is much higher than XGBoost for training data but close when it comes to test data, implying that the Random Forest model overtrained. 



 Table \ref{tab:classification_metrics_noaugmentation} presents the Precision, Recall and F1 scores for training and test sets across each classification model for bluebottle \textit{presence} and \textit{absence} class labels on the training and test dataset, respectively.  
We observe the challenge of class imbalance, as all three models perform well for the majority class (Absence) but struggle significantly with the minority class (Presence). Although  Random Forest and XGBoost outperform MLP in identifying the Presence class, their overall performance remains poor without data augmentation techniques.

\begin{table}[htbp!]
    \centering
    \small
    \caption{Confusion matrix obtained for true negative (TN), false negative (FN), false positive (FP) and true positive (TP) across the models.}
    \small
    \begin{tabular}{|c|c|c|c|}
        \hline
        \multicolumn{2}{|c|}{\textbf{MLP}} & \multicolumn{2}{c|}{\textbf{Actual Values}} \\ \hline
        \multicolumn{2}{|c|}{} & Absence & Presence \\ \hline
        \multirow{2}{*}{\textbf{Predicted Values}} & Negative (0) & TN(1306) & FN(1) \\ \cline{2-4} 
        &Positive (1) & FP(97) & \textbf{TP(0)} \\ \hline
         \multicolumn{2}{|c|}{\textbf{Random Forest}} & \multicolumn{2}{c|}{\textbf{Actual Values}} \\ \hline
        \multicolumn{2}{|c|}{} & Absence & Presence \\ \hline
        \multirow{2}{*}{\textbf{Predicted Values}} & Negative (0) & TN(1268) & FN(39) \\ \cline{2-4} 
         & Positive (1) & FP(89) & \textbf{TP(8)} \\ \hline
         \multicolumn{2}{|c|}{\textbf{XGBoost}} & \multicolumn{2}{c|}{\textbf{Actual Values}} \\ \hline
        \multicolumn{2}{|c|}{} & Absence & Presence \\ \hline
        \multirow{2}{*}{\textbf{Predicted Values}} & Negative (0) & TN(1304) & FN(3) \\ \cline{2-4} 
         & Positive (1) & FP(93) & \textbf{TP(4)} \\ \hline
    \end{tabular}
    \label{tab:confusionmatrix}
\end{table}

Table \ref{tab:confusionmatrix} presents the confusion matrix for a typical model training run (out of 30 experiments), showing the actual versus the predicted values across the models. The results indicate that all classifiers perform well in identifying the majority class (Absence), but struggle significantly with correctly predicting the minority class (Presence) likely due to class imbalance as shown in Table \ref{tab:classification_metrics_noaugmentation}. The results emphasise the strong performance of the models on the majority class and highlight their inability to accurately predict the minority class. This underscores the importance of addressing class imbalance, overlap and unreliable negative data, such as incorporating data augmentation techniques, to improve the model's sensitivity to the majority class.


\subsubsection{Results with Data Augmentation}
\label{subsec:data_augementation}


We next present the results of   data augmentation techniques for enhancing the model performance, including methods for handling class imbalance (SMOTE and Random Undersampling) and the no negative approach for unreliable negative class (CT-GAN) to 
mitigate class imbalance and class overlap.  Furthermore, we use the One-Class SVM to take an alternative approach by assuming there is no negative data.

\begin{table*}[htbp!]
    \centering
    \begin{minipage}[b]{\linewidth}
    \centering
    \caption{Results for bluebottle prediction performance across selected models with data augmentation for 30 experimental runs. Note that we provide the mean performance with the standard deviation in brackets. The AUC for the one-class SVM is reported as NAN because the model trained using only one class. The F1 score on the test data from the best-performing techniques is highlighted in bold.}
    \resizebox{\textwidth}{!}{
    \begin{tabular}{|l|c|c|c|c|c|c|c|}
        \hline
        \textbf{Methods} & \multirow{2}{*}{\textbf{Models}} & \multicolumn{3}{c|}{\textbf{Train}} & \multicolumn{3}{c|}{\textbf{Test}} \\
        \cline{3-8}
         & & \textbf{Accuracy}   & \textbf{F1 Score}   & \textbf{AUC}  & \textbf{Accuracy}  & \textbf{F1 Score}   & \textbf{AUC}  \\
        \hline
        \multirow{3}{*}{SMOTE} & MLP & 0.848 (0.018) & 0.855 (0.020) & 0.914 (0.014) & 0.744 (0.026) & 0.170 (0.024) & 0.654 (0.032) \\ \cline{2-8}
         
        & Random Forest & 0.710 (0.019) & 0.757 (0.013) & 0.808(0.019) & 0.531 (0.040) & 0.182 (0.013) & 0.718 (0.022) \\ \cline{2-8}
       
        & XGBoost & 0.724 (0.013) & 0.779 (0.025) & 0.864 (0.009) & 0.483 (0.025) & 0.169 (0.011) & 0.701 (0.028) \\ \hline
         
        \multirow{3}{*}{Random Undersampling} & MLP & 0.724(0.049) & 0.736 (0.061) & 0.785 (0.050) & 0.637 (0.037) & \textbf{0.649 (0.037)} & 0.689 (0.036) \\ \cline{2-8}
        
        & Random Forest & 0.768 (0.023) & 0.790 (0.022) & 0.883 (0.028) & 0.678(0.028) & \textbf{0.705 (0.021)} & 0.727 (0.026) \\ \cline{2-8}
        
        & XGBoost & 0.679 (0.025) & 0.752 (0.017) & 0.814 (0.015) & 0.623 (0.027) & \textbf{0.710 (0.023)} & 0.720 (0.028) \\ \hline
        \multirow{3}{*}{\shortstack{Synthetic \\Negative\\ Approach \\ (CT-GAN)}} & MLP & 0.794 (0.055) & 0.795 (0.047) & 0.878 (0.048) & 0.718 (0.046) & \textbf{0.726 (0.038)} & 0.800 (0.042) \\ \cline{2-8}
        
        & Random Forest & 0.891 (0.016) & 0.888 (0.016) & 0.956 (0.007) & 0.767 (0.036) & \textbf{0.775 (0.036)} & 0.860 (0.029) \\ \cline{2-8}
        
        & XGBoost & 0.843 (0.022) & 0.862 (0.017) & 0.968 (0.008) & 0.724 (0.035) & \textbf{0.767 (0.028)} & 0.843 (0.030) \\ \hline
        \multirow{1}{*}{-} & One-Class SVM & 0.500(0.008) & 0.667 (0.007) & nan (nan) & 0.618 (0.035) & 0.561 (0.047) & 0.618 (0.035) \\ \hline
    \end{tabular}}
    \label{tab:performance_comparison_augmentation}
    \end{minipage}%
  \hspace{2.0cm} 
  
  \begin{minipage}[b]{1.0\linewidth}
         \centering
         \caption{Results from the best experimental run in Table \ref{tab:performance_comparison_augmentation} showing Absence and Presence for training and test data.}
   \resizebox{\textwidth}{!}{
   \begin{tabular}{|l|l|c|c|c|c|c|c|c|c|c|c|c|c|}
       \toprule
      \textbf{Methods} & \textbf{Models} & \multicolumn{3}{c|} {\textbf{{Class - Absence}}} & \multicolumn{3}{c|}{\textbf{{Class - Presence}}} & \multicolumn{3}{c|}{\textbf{{Class - Absence}}} & \multicolumn{3}{c|}{\textbf{{Class - Presence}}} \\ 
        \cmidrule(lr){3-5} \cmidrule(lr){6-8} \cmidrule(lr){9-11} \cmidrule(lr){12-14}
& & Precision & Recall & F1 Score & Precision & Recall & F1 Score & Precision & Recall & F1 Score & Precision & Recall & F1 Score \\ 
       \midrule
       \multirow{3}{*}{SMOTE} 
       & MLP & 0.922 & 0.748 & 0.826 & 0.788 & 0.936 & 0.856 & 0.952 & 0.735 & 0.830 & 0.122 & 0.495 & 0.200 \\ \cline{2-14}
       & Random Forest & 0.878 & 0.523 & 0.655 & 0.660 & 0.927 & 0.711 & 0.970 & 0.512& 0.670 & 0.106 & 0.784 & 0.187 \\ \cline{2-14}
      & XGBoost & 0.943 & 0.500 & 0.653 & 0.659 & 0.970 & 0.785 & 0.961 & 0.476 & 0.637 & 0.100 & 0.742 & 0.169 \\ 
       \midrule
    \multirow{3}{*}{\shortstack{Random \\Undersampling}} 
       & MLP & 0.742 & 0.736 & 0.739 & 0.761 & 0.766& 0.763& 0.708 & 0.656 & 0.681& \textbf{0.637} & \textbf{0.691} & \textbf{0.663} \\ \cline{2-14}
       & Random Forest & 0.854 & 0.589 & 0.697 & 0.707 & 0.908 & 0.795 & 0.758 & 0.521 & 0.617& \textbf{0.597} & \textbf{0.810} & \textbf{0.687} \\ \cline{2-14}
       & XGBoost & 1.000 & 0.488 & 0.656 & 0.681 & 1.000 & 0.810 & 0.826 & 0.396 & 0.567 & \textbf{0.905}& \textbf{0.697} & \textbf{0.769} \\ \midrule
\multirow{3}{*}{\shortstack{Synthetic \\Negative\\ Approach \\ (CT-GAN)}}         & MLP & 0.828 & 0.511 & 0.632 & 0.623 & 0.883& 0.731 & 0.759 & 0.524 & 0.620& \textbf{0.672} & \textbf{0.854} & \textbf{0.752} \\ \cline{2-14}
       & Random Forest & 0.875 & 0.943 & 0.908 & 0.932 & 0.853 & 0.891 & 0.723 & 0.810 & 0.764 & \textbf{0.814} & \textbf{0.729} & \textbf{0.769} \\ \cline{2-14}
       & XGBoost & 1.000 & 0.716 & 0.835 & 0.763 & 1.000 & 0.866 & 0.753 & 0.655 & 0.701 & \textbf{0.729}& \textbf{0.813} & \textbf{0.769} \\ \midrule
       -- & One-Class SVM& 0.000 & 0.000 & 0.000 & 1.000 & 0.496& 0.663& 0.600 & 0.667 & 0.632& 0.625 & 0.556 & 0.588 \\ \midrule
      
 \end{tabular}
  }
 \label{tab:classification_metrics_augmentation}
    \end{minipage}
\end{table*}

\begin{table}[htbp!]
    \centering
    \small
    \caption{Confusion matrix for the best experimental run using Synthetic Negative Approach 1}
    \small
    \begin{tabular}{|c|c|c|c|}
        \hline
        \multicolumn{2}{|c|}{\textbf{MLP}} & \multicolumn{2}{c|}{\textbf{Actual Values}} \\ \hline
        \multicolumn{2}{|c|}{} & Negative (0) & Positive (1) \\ \hline
        \multirow{2}{*}{\textbf{Predicted Values}} & Negative (0) & TN(44) & FN(40) \\ \cline{2-4} 
         & Positive (1) & FP(14) & \textbf{TP(82)} \\ \hline
         \multicolumn{2}{|c|}{\textbf{Random Forest}} & \multicolumn{2}{c|}{\textbf{Actual Values}} \\ \hline
        \multicolumn{2}{|c|}{} & Negative (0) & Positive (1) \\ \hline
        \multirow{2}{*}{\textbf{Predicted Values}} & Negative (0) & TN(68) & FN(16) \\ \cline{2-4} 
         & Positive (1) & FP(26) & \textbf{TP(70)} \\ \hline
         \multicolumn{2}{|c|}{\textbf{XGBoost}} & \multicolumn{2}{c|}{\textbf{Actual Values}} \\ \hline
        \multicolumn{2}{|c|}{} & Negative (0) & Positive (1) \\ \hline
        \multirow{2}{*}{\textbf{Predicted Values}} & Negative (0) & TN(55) & FN(29) \\ \cline{2-4} 
         & Positive (1) & FP(18) & \textbf{TP(78)} \\ \hline
         
    \end{tabular}
      
    \label{tab:confusionmatrixaugment}
\end{table}

Table \ref{tab:performance_comparison_augmentation} compares the performance of SMOTE and Random Undersampling for addressing the class imbalance in bluebottle prediction, using  MLP, Random Forests, XGBoost and One-Class SVM. We report selected performance metrics including, classification accuracy, F1 score, and AUC   for both training and test sets, based on 30 independent model training runs. We observe that the classification models with SMOTE  improved in training performance across all three metrics (F1, AUC and accuracy), with MLP achieving the highest accuracy, followed by XGBoost and then Random Forest. However, the test performance dropped drastically across all metrics, particularly in the F1 score with MLP (0.170), Random Forest (0.182) and XGBoost (0.169).
Random Undersampling showed similar performance metrics for training data when compared to SMOTE.  On the test set, Random Undersampling with Random Forest demonstrated much higher accuracy (0.678), F1 score (0.705) and AUC (0.727) than  SMOTE. The Synthetic Negative approach, which involves treating the negative class as no negative (unreliable) uses two approaches (CT-GAN and One-Class SVM). In Approach 1 using CT-GAN, all classifiers show the best overall performance, when compared to SMOTE and Random Undersampling for the training set. Similarly, the Random Forest outperformed other classifiers on the test set (accuracy = 0.767, F1 score = 0.775 and AUC = 0.860). In the case of Approach 2 involving One-Class SVM, on the training data and test data, it had a lower performance across all metrics (accuracy = 0.500, F1 score = 0.667). The  AUC = NAN because we trained using One-Class SVM when compared to the other approaches. 
Overall, we observe the best results in the case of the Synthetic Negative Approach 1, with high performance on both training and test sets across all metrics, particularly with Random Forest. Random Undersampling also produced good performance and  the One-Class SVM   performed slightly poorly and SMOTE had the lowest F1 score on the test data.
In the case of SMOTE, we observe that there is a slight improvement when compared to the previous result without any data augmentation (Table \ref{tab:classification_metrics_noaugmentation}).

Table \ref{tab:classification_metrics_augmentation} presents the classification report of the best experimental run from Table \ref{tab:performance_comparison_augmentation} for both training and test data for the Absence and Presence classes of bluebottle across different approaches and models using Precision, Recall and F1 score. Since the F1 score is composed of both Precision and Recall, we limit our discussion to the F1 score from the focus of individual classes. Generally, we can observe that the minority class (Presence) faces the challenge of accurate predictions for SMOTE, which improves with Random Undersampling and Synthetic Negative Approach 1. The One-Class SVM struggles a bit but improved better than SMOTE.


Table \ref{tab:confusionmatrixaugment} presents the confusion matrix for the respective data augmentation methods and classification models. The data augmentation approach significantly improves the classifier's ability to identify bluebottle Presence with a much higher true positive class (MLP = 82, RF = 70, and XGBoost = 78) when compared to no data augmentation (Table \ref{tab:metrics_noaugmentation}. The false positive counts are reduced across the models (MLP = 14, Random Forest = 26, and XGBoost = 18), indicating better discrimination between the classes. The true negatives remain high, ensuring that the majority of negative (Absence) class predictions are reliable. Overall, the models are far more balanced, achieving a substantial increase in true positives while maintaining strong performance on the true negatives. This highlights the effectiveness of Synthetic Negative Approach 1 (CT-GAN) in mitigating the effect of class imbalance and overlap.

\subsection{Synthetic data visualisation}

Next, we compare the distribution of the Synthetic Negative Approach 1 (CT-GAN) generated data with the real data to understand how the synthetic data enhanced in achieving better model predictive performance. 
\begin{figure*}[htbp]
    \centering
    \begin{subfigure}{0.29\textwidth}
        \includegraphics[width=\linewidth]{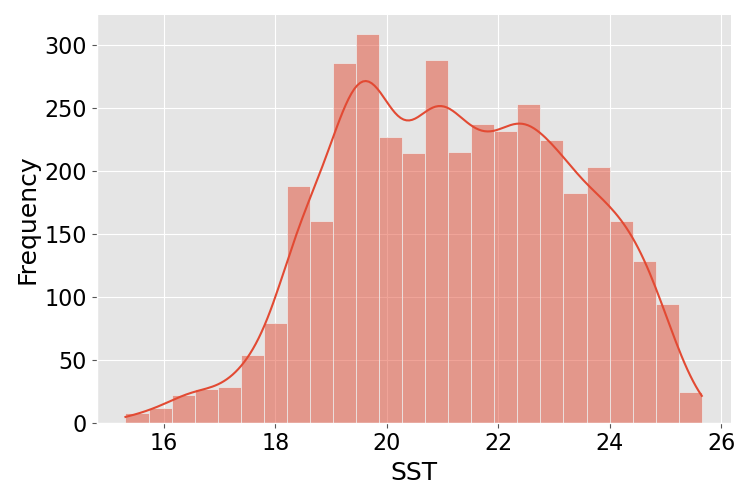} 
        \caption{Real data distribution: Sea Surface Temperature}
        \label{fig:densitySST}
    \end{subfigure}
    \hfill
    \begin{subfigure}{0.29\textwidth}
        \includegraphics[width=\linewidth]{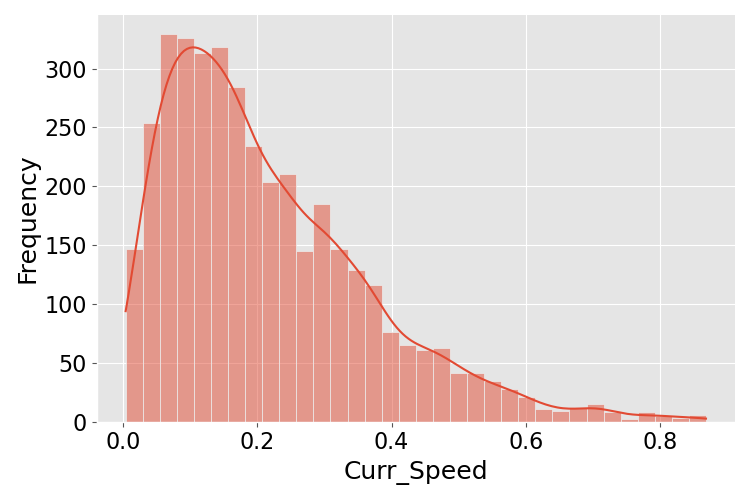} 
        \caption{Real data distribution: Current Speed}
        \label{fig:densityCurrSpeed}
    \end{subfigure}
    \hfill
    \begin{subfigure}{0.29\textwidth}
        \includegraphics[width=\linewidth]{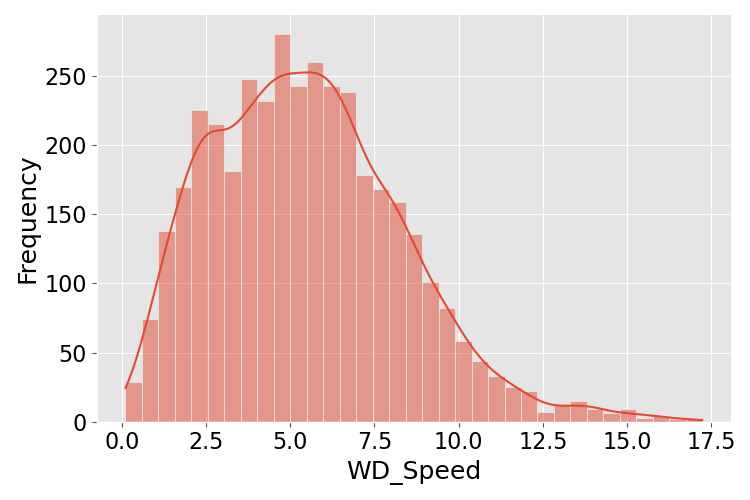} 
        \caption{Real data distribution: Wind Speed}
        \label{fig:densitywdspeed}
    \end{subfigure}
    
    \vspace{0.1cm}
    \begin{subfigure}{0.29\textwidth}
        \includegraphics[width=\linewidth]{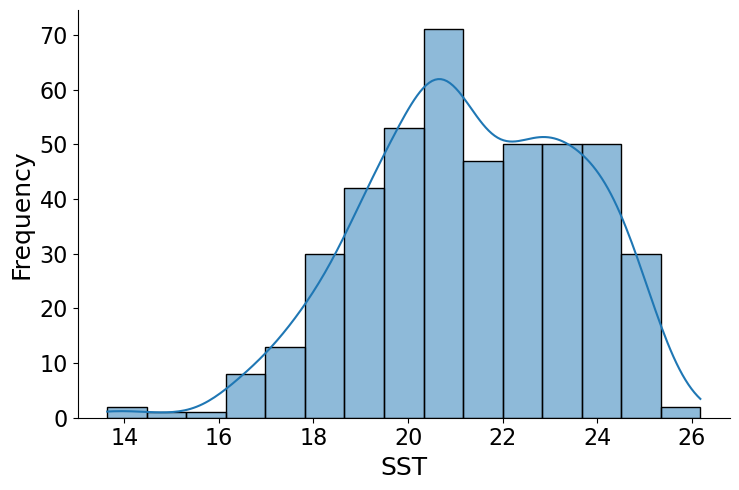} 
        \caption{Synthetic data distribution: SST (CT-GAN)}
        \label{fig:density_sst_ctgab}
    \end{subfigure}
    \hfill
    \begin{subfigure}{0.29\textwidth}
        \includegraphics[width=\linewidth]{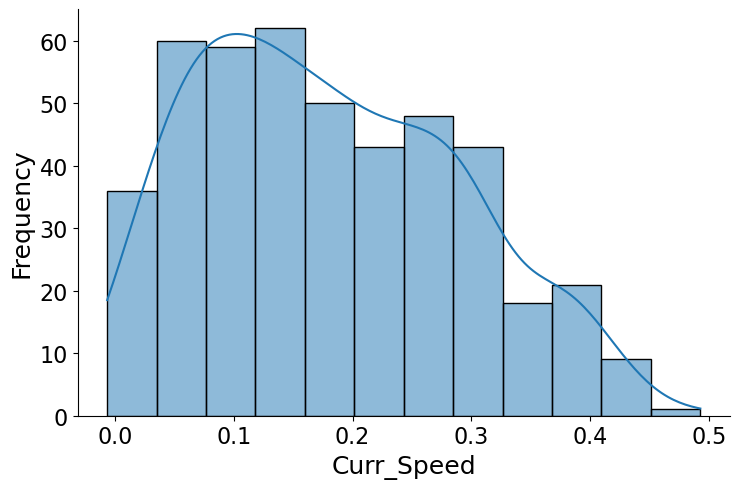} 
        \caption{Synthetic data distribution: Current Speed (CT-GAN)}
        \label{fig:density)Curr_peed_ctgan}
    \end{subfigure}
    \hfill
    \begin{subfigure}{0.29\textwidth}
        \includegraphics[width=\linewidth]{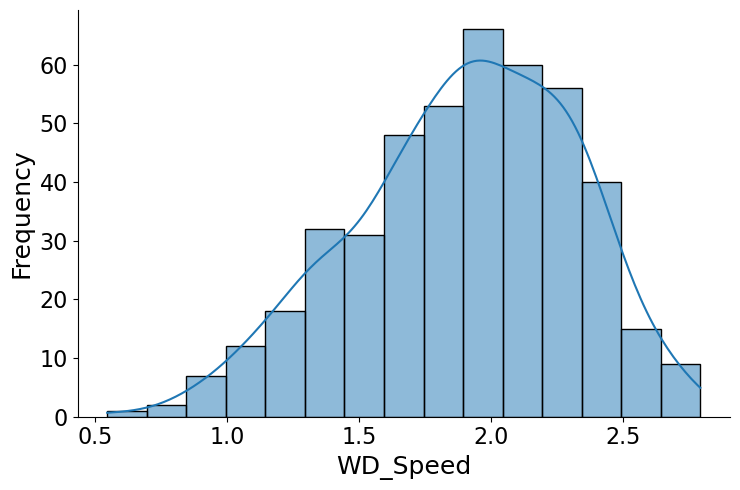} 
        \caption{Synthetic data distribution: Wind Speed (CT-GAN)}
        \label{fig:density_wdspeed_ctgan}
    \end{subfigure}
    
    \caption{Density plots for actual data distributions and CT-GAN generated data for three non-circular variables:  SST, Current Speed, and Wind Speed. The top row (red) is the real data while the bottom row (blue) is the generated data}
    \label{fig:density_plot}
\end{figure*}
Figure \ref{fig:density_plot} presents the distribution of the real data compared to the synthetic data generated by CT-GAN, where the synthetic data demonstrates varying levels of accuracy in capturing the characteristics of the real distributions. In SST, CT-GAN effectively captures the overall range but exhibits a blocky and less smooth appearance. In the case of Current Speed, the synthetic data reflects the right-skewed nature and general range of the real data distribution. However, for Wind Speed,   the synthetic data fails to replicate the pronounced right skewness observed in the real data due to undersampling when matching the size of the positive class. Overall, although CT-GAN approximates the general shapes of the distributions, it struggles with smoothness, tail behaviour and peak characteristics,  which may stem from sample reduction during data generation.

\begin{figure*}[t!] 
    \centering
    \begin{subfigure}{0.40\textwidth}
        \includegraphics[width=\linewidth]{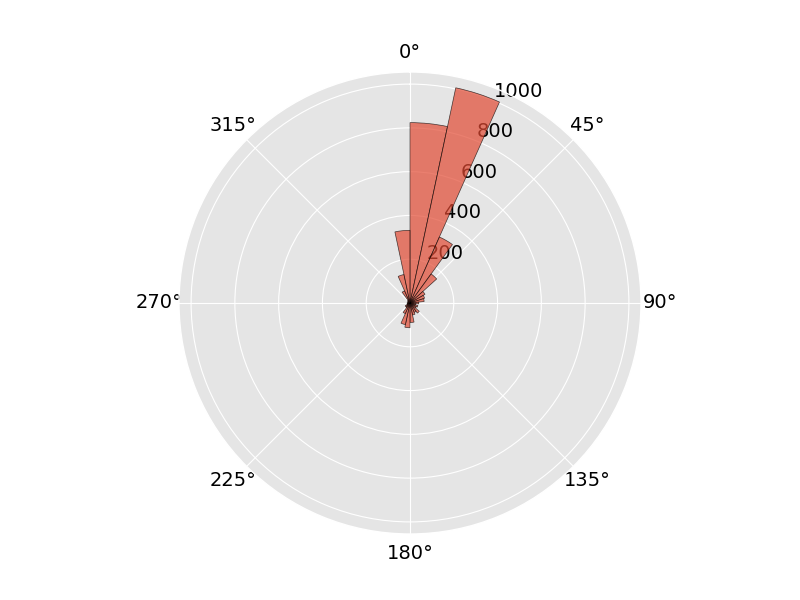}
        \caption{Actual data distribution: Current Direction}
    \label{fig:actual_currdir}
    \end{subfigure}
    \hfill
    \begin{subfigure}{0.40\textwidth}
        \includegraphics[width=\linewidth]{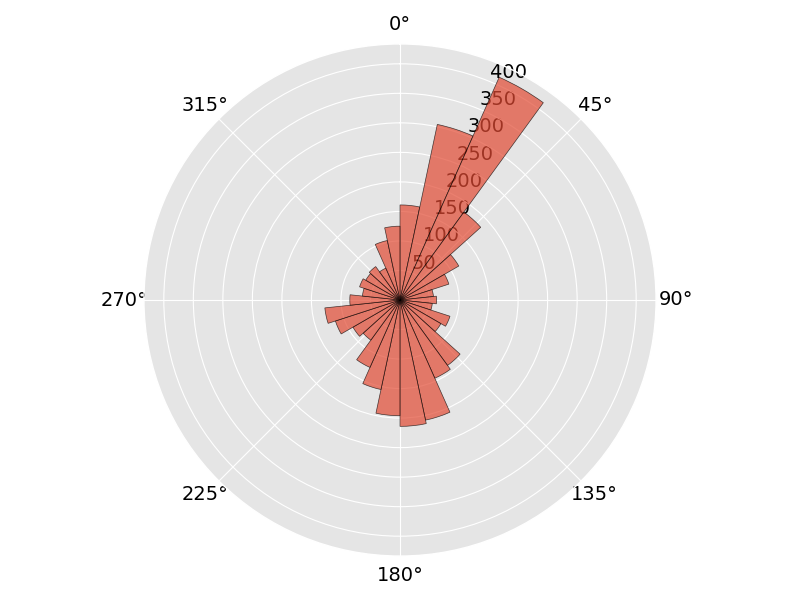}
        \caption{Actual data distribution: Wind Direction}
        \label{fig:actualwddir}
    \end{subfigure}
    \vspace{0.1cm}
    \begin{subfigure}{0.40\textwidth}
        \includegraphics[width=\linewidth]{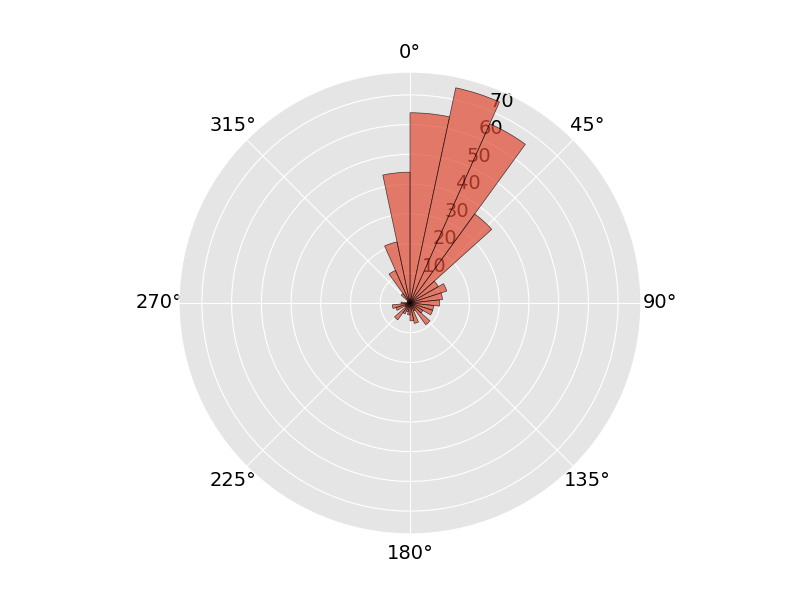}
        \caption{Synthetic data distribution: Current Direction(CT-GAN)}
    \label{fig:ctgan_currdir}
    \end{subfigure}
    \hfill
    \begin{subfigure}{0.40\textwidth}
        \includegraphics[width=\linewidth]{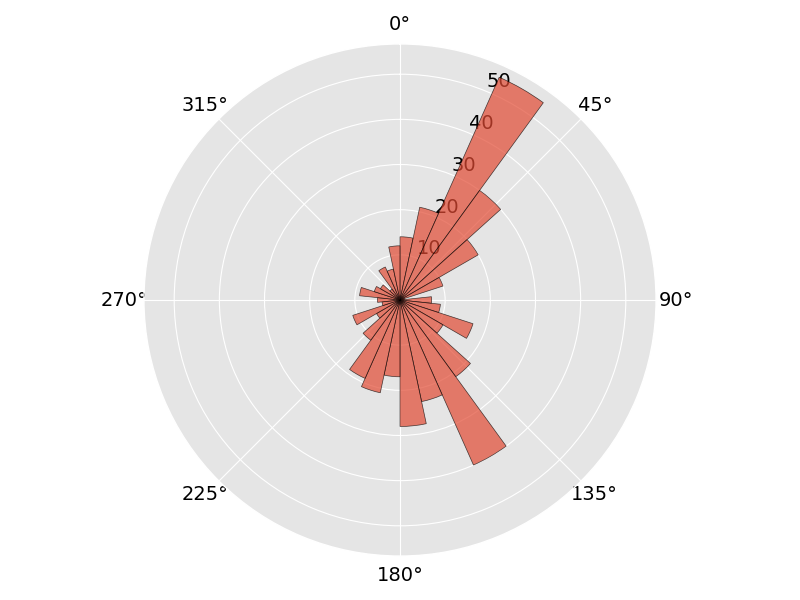}
        \caption{Synthetic data distribution: Wind Direction (CT-GAN)}
        \label{fig:ctgan_wddir}
    \end{subfigure}
    \caption{Circular histograms comparing actual data and synthetic data (CT-GAN) for two circular variables: Current Direction and Wind Direction. The top row represents actual data, while the bottom row shows synthetic generated data.}
    \label{fig:circular_plots}
\end{figure*}

 Fig \ref{fig:circular_plots} compares the actual data and CT-GAN generated data distribution for the two circular variables wind direction and current direction. The top row represents the actual data, while the bottom row shows the CT-GAN-generated data. In the case of the Current Direction, the actual data exhibits a concentrated distribution, with notable peaks around $0\,^\circ$  and $360\,^\circ$ (North), along with smaller peaks around other directions. The CT-GAN generated plot closely replicates this pattern, capturing the concentration at $0\,^\circ$ and $360\,^\circ$ (North), but with some variation that resembles the shape and spread of the original data. 
The Wind Direction in the actual data is more dispersed, with higher concentrations at $0\,^\circ$ and $270\,^\circ$. The CT-GAN generated Wind Direction approximates the general distribution, but the pattern appears somewhat less defined, with a wider spread and less precision compared to the original data, especially around $270\,^\circ$ (Westerly winds, usually associated with absence \cite{bourg2022driving}). 
Overall, the CT-GAN model effectively replicates the directional trends of both current direction and wind direction distributions. However, the synthetic data introduces more variability, particularly for the wind direction, where the distribution becomes more spread out. This suggests that while CT-GAN can capture the general distribution, it may lose precision in the representation of distinct peaks.


\subsection{Visualisation}

\begin{figure*}[htbp!]
    \centering
    \begin{subfigure}{0.45\textwidth}
        \includegraphics[width=\linewidth]{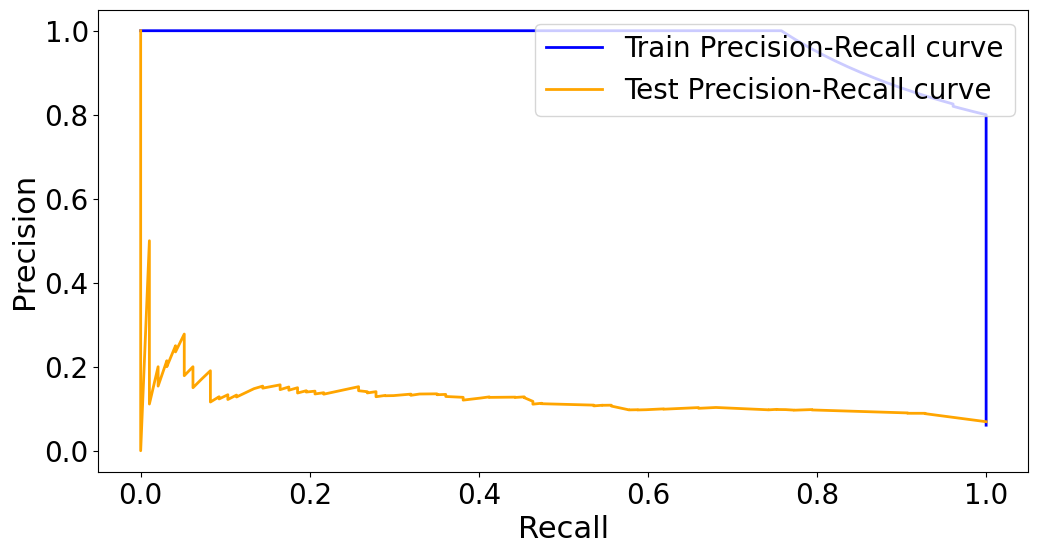} 
        \caption{No data augmentation}
        \label{fig:precision_actual_rf}
    \end{subfigure}
    \hfill
    \begin{subfigure}{0.45\textwidth}
        \includegraphics[width=\linewidth]{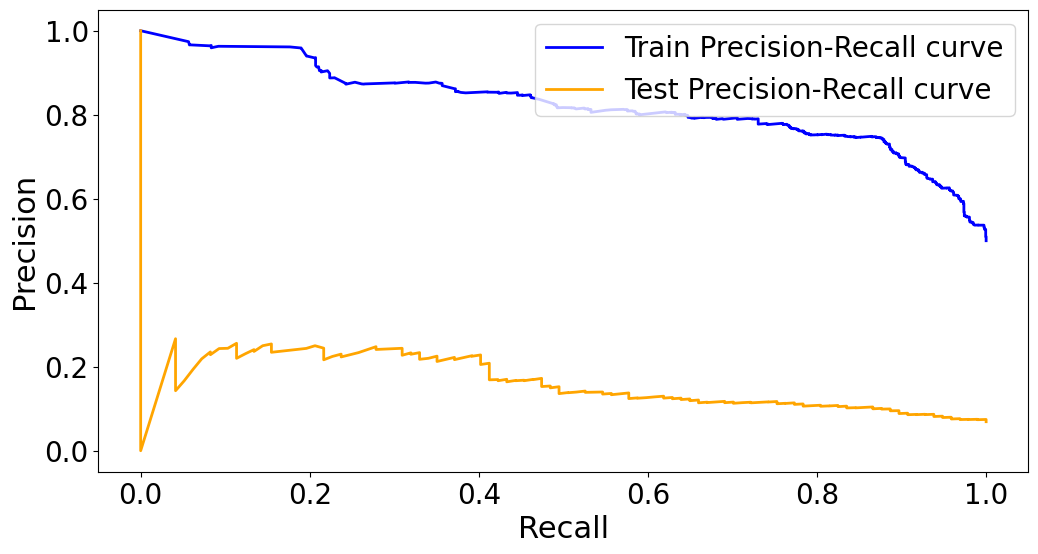} 
        \caption{SMOTE}
\label{fig:precision_smote_rf}
    \end{subfigure}
    
    \vspace{0.1cm}
    \begin{subfigure}{0.45\textwidth}
        \includegraphics[width=\linewidth]{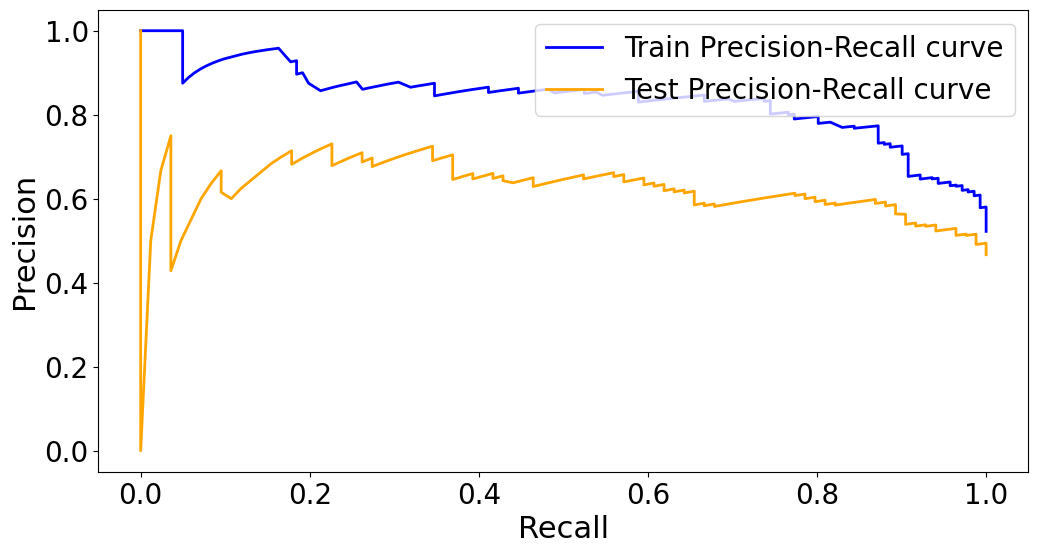} 
        \caption{Random Undersampling}
        \label{fig:precision_random_sample_rf}
    \end{subfigure}
    \hfill
    \begin{subfigure}{0.45\textwidth}
        \includegraphics[width=\linewidth]{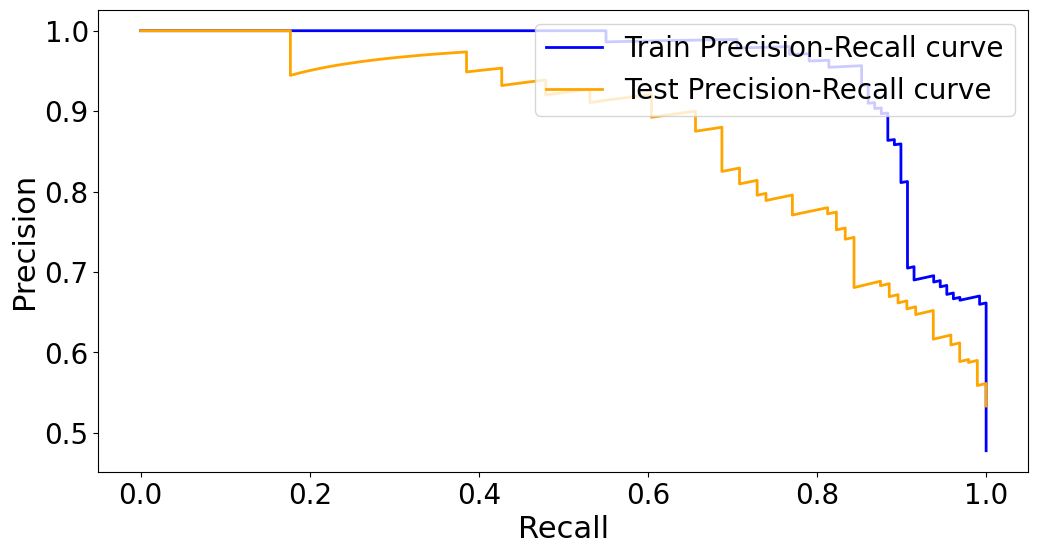} 
        \caption{No negative approach}
        \label{fig:precision_ctgan_rf}
    \end{subfigure}

    \caption{Precision-Recall curves comparing the performance of various data handling techniques (a) original dataset, (b) SMOTE-augmented dataset, (c) Random sampling dataset, and (d) CT-GAN  (no negative approach).}
    \label{fig:train_test_Curve}
\end{figure*}

We next utilise the Random Forest model, which performed best across the different data augmentation methods. Figure \ref{fig:train_test_Curve} depicts the precision-recall curves for training and testing data across various data handling techniques (no augmentation, SMOTE, Random Undersampling, and no negative approach via CT-GAN). We can observe that the no augmentation strategy shows high performance on training data but poor generalisation on the test data, indicated by a steep drop in precision for the test set. The training precision-recall curve of the SMOTE data remains strong, but test performance improves slightly compared to the original data, though overfitting is still evident. Random Undersampling test data shows better alignment with the training curve, suggesting reduced overfitting and improved generalisation compared to SMOTE. The no negative approach (CT-GAN) demonstrates the best generalisation among all techniques, with the test and train curves closer together, including balanced model performance on both datasets.

\begin{figure*}[htbp!]
    \centering
    \begin{subfigure}{0.45\textwidth}
        \includegraphics[width=\linewidth]{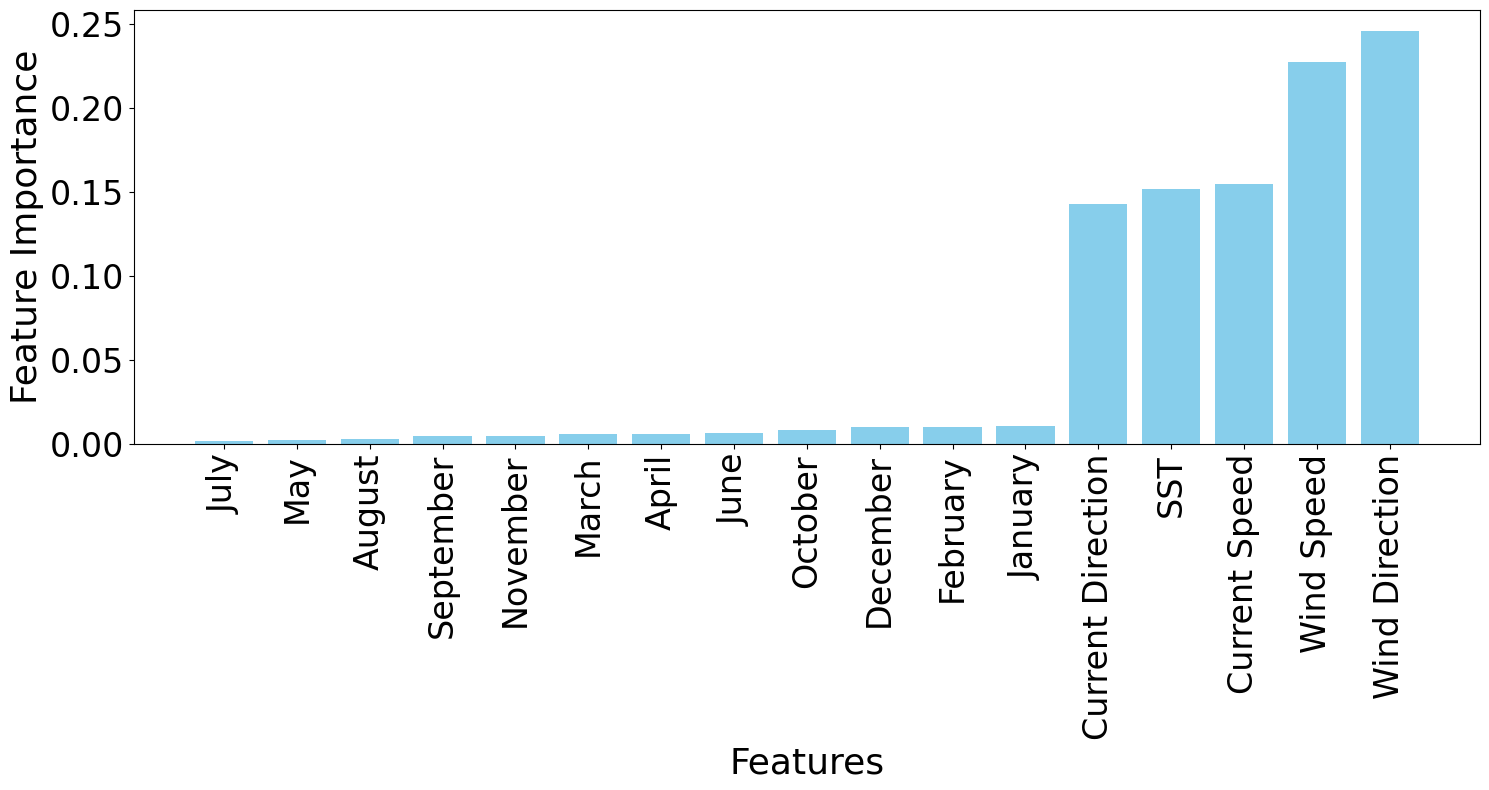} 
        \caption{Individual contribution from actual data.}
\label{fig:feature_importance_actual}
    \end{subfigure}
    \hfill
    \begin{subfigure}{0.45\textwidth}
        \includegraphics[width=\linewidth]{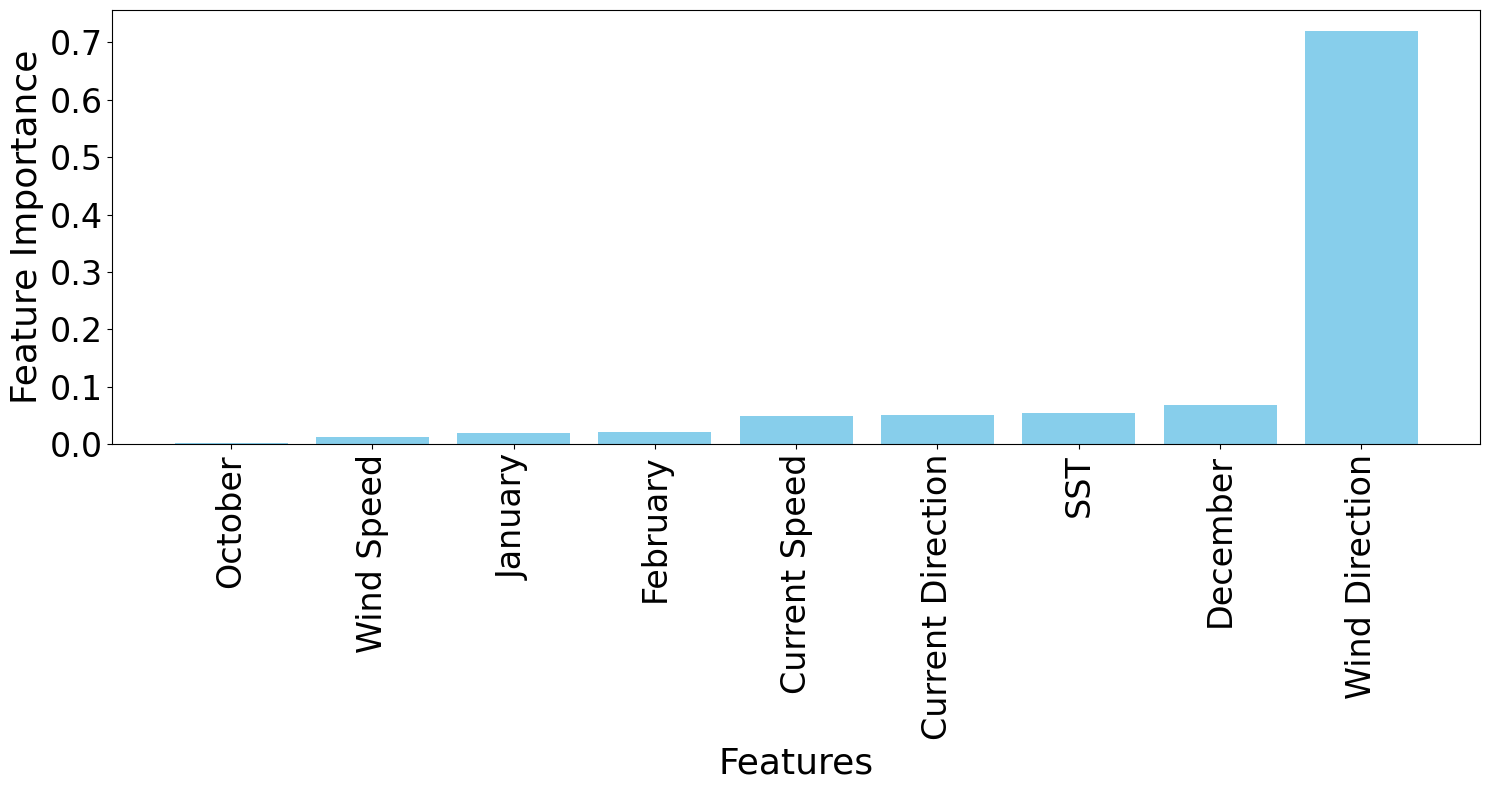} 
        \caption{Individual contribution from SMOTE data.}
\label{fig:features_importance_smote}
    \end{subfigure}

    \vspace{0.1cm}
    \begin{subfigure}{0.4\textwidth}
        \includegraphics[width=\linewidth]{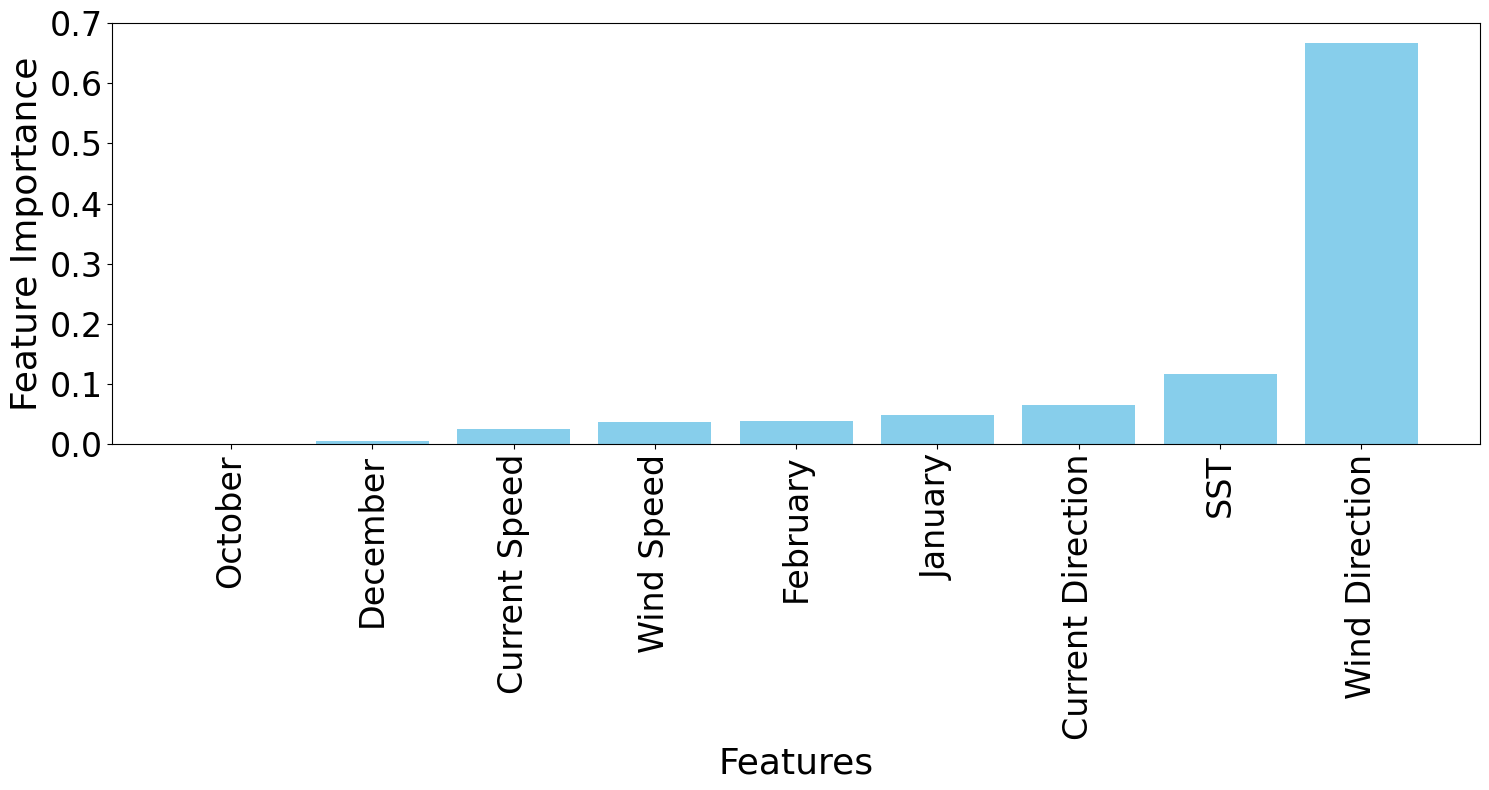} 
        \caption{Individual contribution from Random Sampling data.}
        \label{fig:features_importance_sampling}
    \end{subfigure}
    \hfill
    \begin{subfigure}{0.4\textwidth}
        \includegraphics[width=\linewidth]{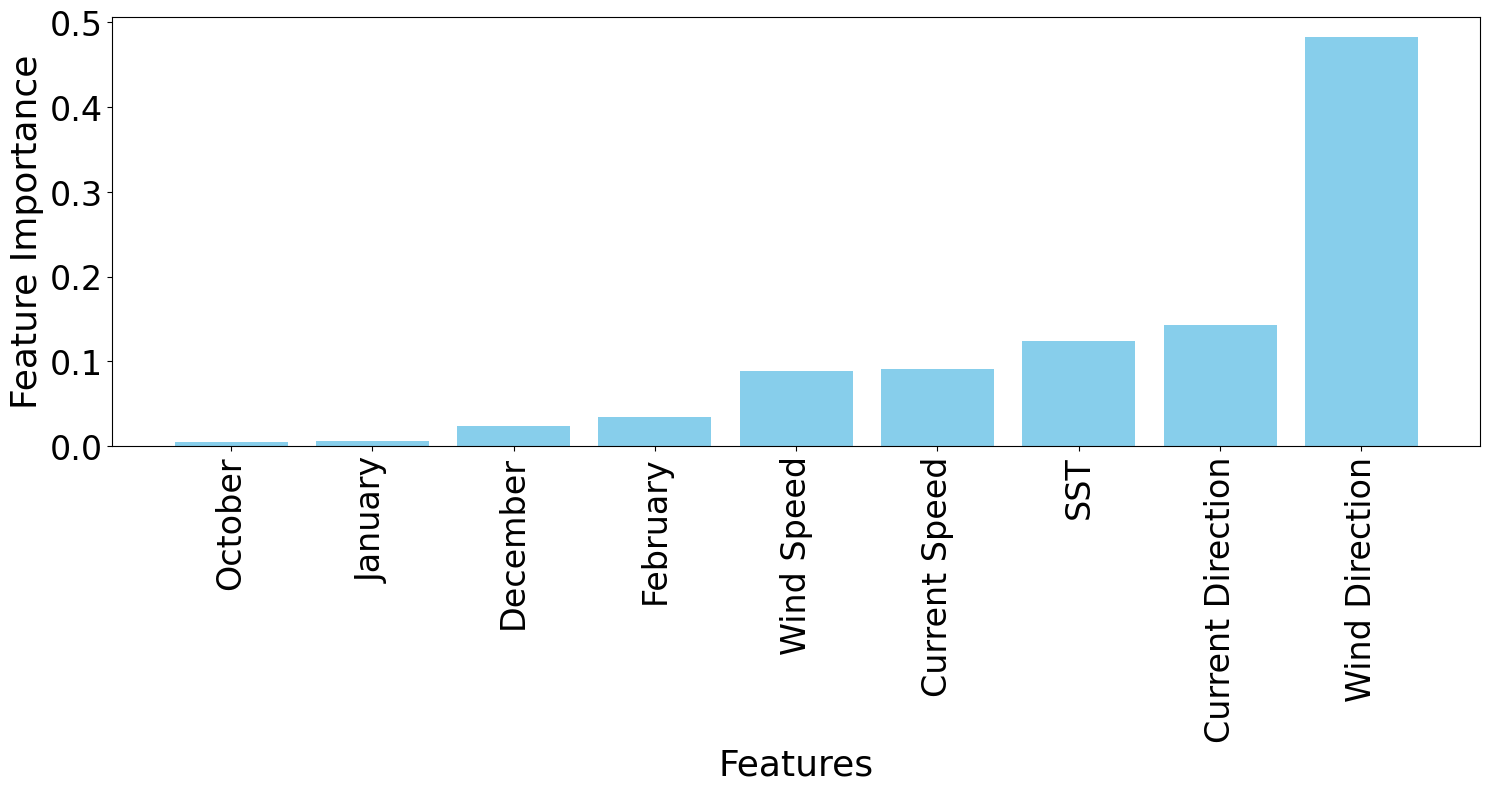} 
        \caption{Individual contribution from CT-GAN data.}
\label{fig:feature_importance_ctgan}
    \end{subfigure}
    \hfill
    \begin{subfigure}{0.4\textwidth}
        \includegraphics[width=\linewidth]{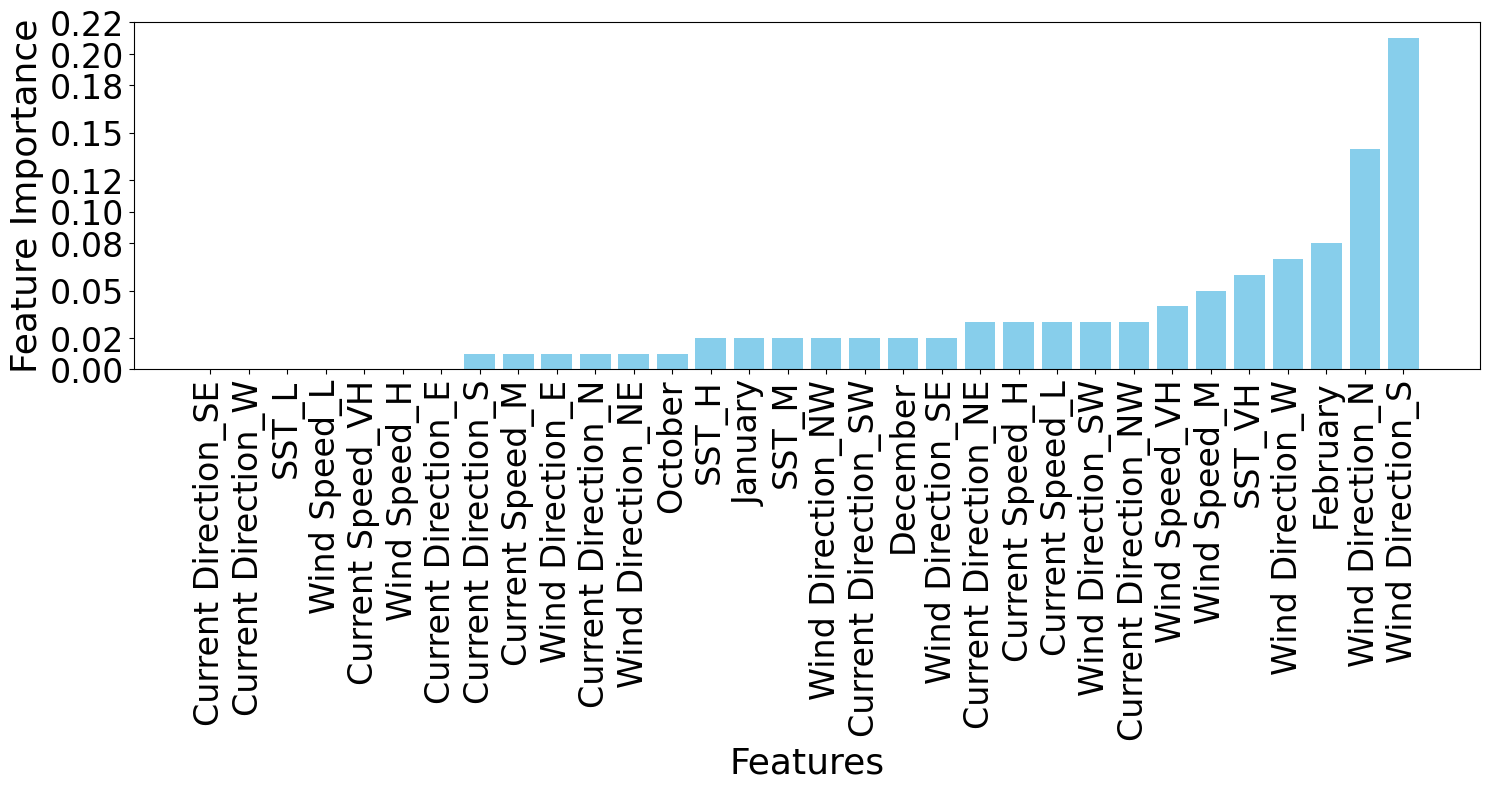} 
        \caption{Individual contribution from thdiscretised features.}
        \label{fig:feature_importance_discretizer}
    \end{subfigure}

    \caption{Feature Importance Analysis Across Different Data Balancing Techniques (Actual Data, SMOTE, Random sampling, and CT-GAN).}
\label{fig:feature_importance_plot}
\end{figure*}

Figure~\ref{fig:feature_importance_plot} shows feature importance analysis across the selected techniques (no data augmentation, SMOTE, random Undersampling, and CT-GAN). The plot compares the feature importance for predicting the target variable across various techniques. The wind direction attribute is dominant across all techniques,  consistently showing the highest importance, particularly in the case of SMOTE, random undersampling and CT-GAN. In Figure \ref{fig:features_importance_sampling} and \ref{fig:feature_importance_ctgan}, we observe that SST and current direction appear as the second and third most influential features. Finally, we discriticised the features of CT-GAN  to determine which direction of the wind contributes the most. Figure \ref{fig:feature_importance_discretizer} shows that wind direction from the North and South drives the bluebottle migration especially in the month of February.

\begin{figure*}[htbp!]
    \centering
    \begin{subfigure}{0.45\textwidth}
        \includegraphics[width=\linewidth]{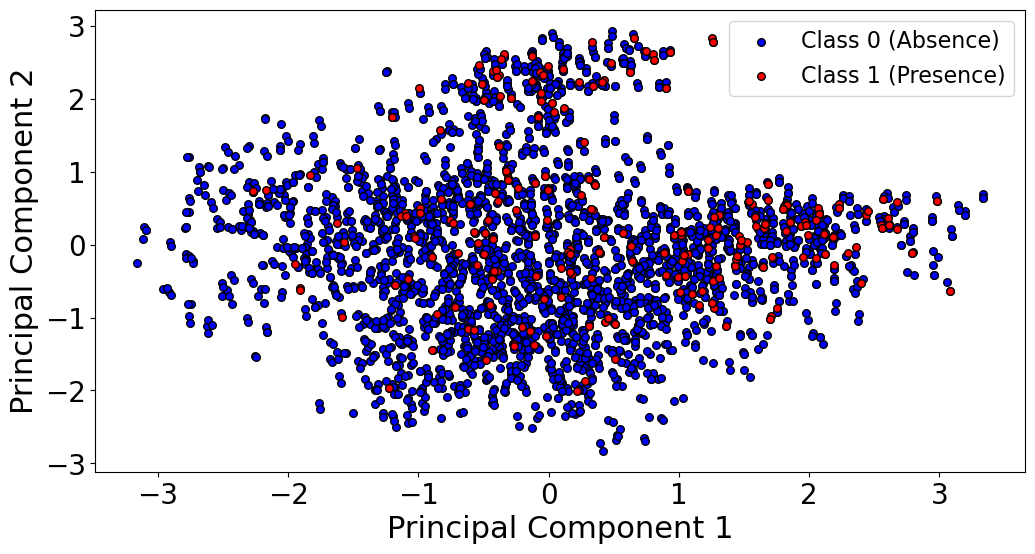} 
        \caption{No data augmentation}
        \label{fig:pca_actual}
    \end{subfigure}
    \hfill
    \begin{subfigure}{0.45\textwidth}
        \includegraphics[width=\linewidth]{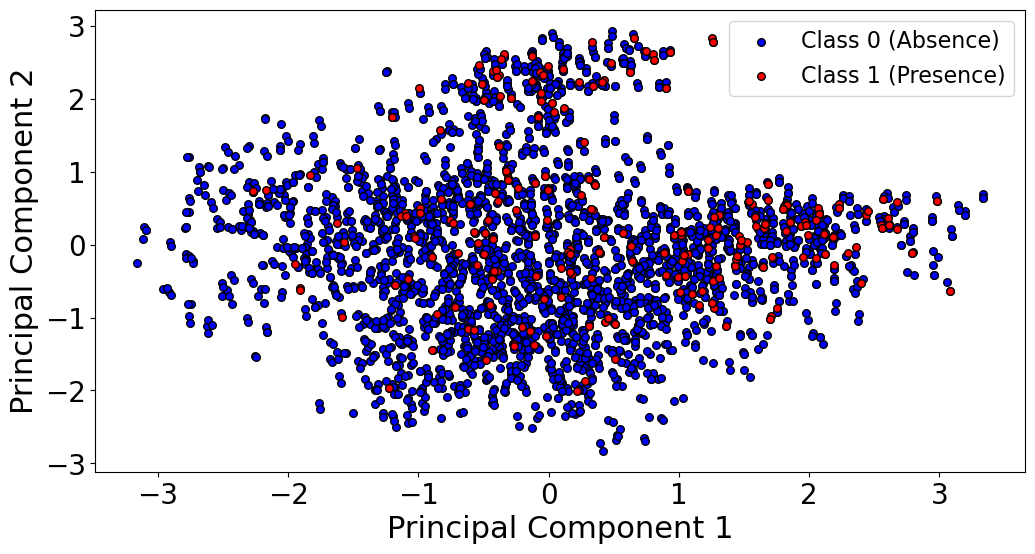} 
        \caption{  SMOTE argumentation }
        \label{fig:PCA_SMOTE}
    \end{subfigure}
    
    \vspace{0.1cm}
    \begin{subfigure}{0.45\textwidth}
        \includegraphics[width=\linewidth]{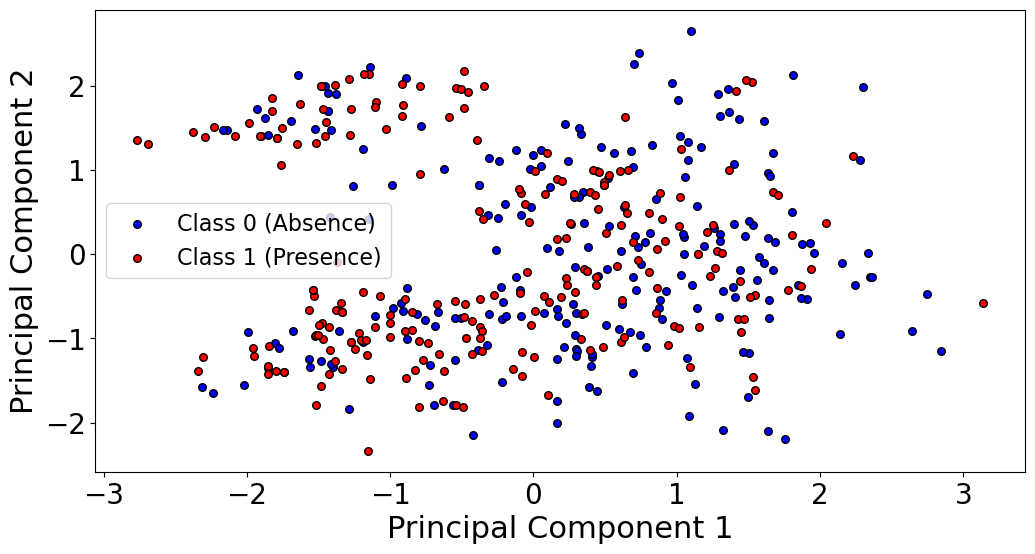} 
        \caption{Random Undersampling}
        \label{fig:pca_random_Sampling}
    \end{subfigure}
    \hfill
    \begin{subfigure}{0.45\textwidth}
        \includegraphics[width=\linewidth]{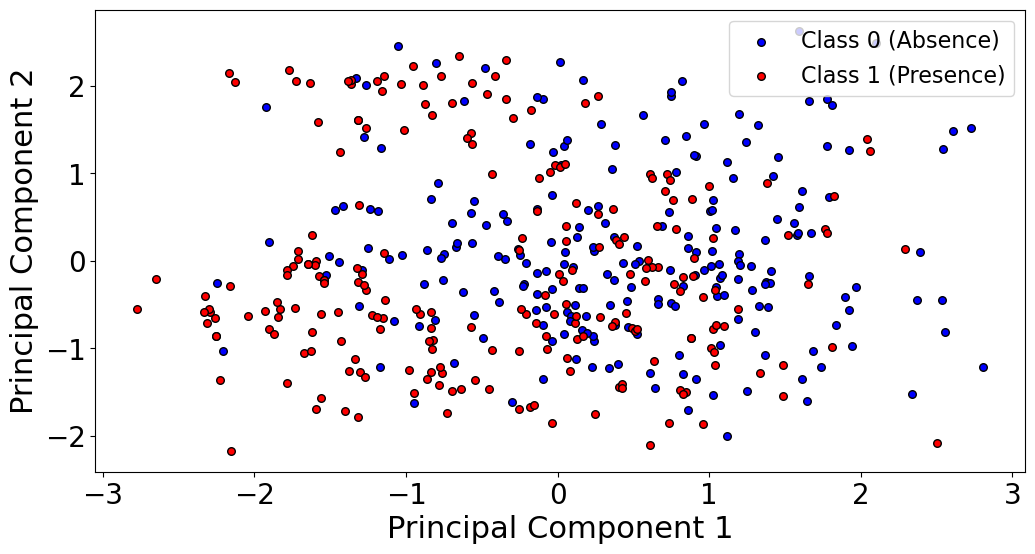} 
        \caption{Synthetic negative approach}
        \label{fig:PCA_CTGAN}
    \end{subfigure}

    \caption{The dataset distribution projected onto the first two principal components using a scatter-plot, with classes distinguished by colour: red representing the positive class and blue representing negative class for different techniques.}
    \label{fig:pca_plot}

\end{figure*}


We further implemented principal component analysis (PCA) to visualise the distribution and separation of the two classes as PCA simplifies complex, high dimensional data into lower-dimensional space. PCA scatter-plot of two major principal components has the ability about overlap of classes for the different data augmentation techniques.

  Figure \ref{fig:pca_plot} presents a PCA scatter-plot depicting the distribution of the dataset projected onto the first two principal components for the different data augmentation methods (No augmentation, SMOTE, random Sampling, CT-GAN) differentiated by the bluebottle Absence (blue) and Presence (red). Figure \ref{fig:pca_actual} exhibit significant class imbalance and overlap, with scattered points and no clustering among classes. This lack of clear separation between the classes suggests that they may not be linearly separable in the feature space, highlighting the complexity of the classification task. Figure \ref{fig:PCA_SMOTE} introduces synthetic points for the minority class (presence) displaying equal number of classes. While the data appears more balanced, the two classes still exhibits significant overlap, and no clear clusters or linear separation are visible. The points in Figure \ref{fig:pca_random_Sampling} are more spread out compared to SMOTE and actual data. There are more visible gaps and clusters between the two classes, which suggests better separation compared to the previous plots. Figure \ref{fig:PCA_CTGAN} shows a more noticeable clustering of the two classes. The points are more spread out and the minority class (presence) forms tighter, more distinct clusters. The visual separation between the two classes is more apparent here, suggesting better synthetic data generation and distribution. Overall, the CT-GAN data figure \ref{fig:PCA_CTGAN} demonstrated the most linear separation and distinct clusters between the two classes. The Random sampling figure \ref{fig:pca_random_Sampling} also shows some improvements but is not as pronounced as CT-GAN.

\section{Discussion}


In this study, we employed selected machine learning and data augmentation techniques   to investigate
what factors lead to the bluebottle marine stinger beaching given class imbalance in the data. We implemented various data augmentation techniques such as SMOTE, Random Undersampling and synthetic negative approach (CT-GAN) to address class imbalance, class overlap and unreliable negative class labels. The results revealed that SMOTE was able to generate synthetic data to address imbalance, but could not address overlap and unreliable negative class associated with the data. The Random Undersampling and synthetic negative approach provided much better  performances with F1 Score (Table \ref{tab:performance_comparison_augmentation}) and precision-recall curves (Figure \ref{fig:train_test_Curve}),  when compared to SMOTE and no augmentation approach. The PCA visualisation (Figure \ref{fig:pca_plot}) clearly shows the effect of unreliable negative class, class imbalance and overlap in the prediction of bluebottle presence and how the synthetic negative approach  improves class separation and pattern detection compared to SMOTE and Random Undersampling.


The bluebottle observational dataset contains features exhibiting circular behaviour such as wind direction and current direction. This characteristic makes the data a combination of linear and circular components. However, most research on jellyfish and  bluebottle beaching tends to overlook the circular nature of certain features (e.g. \cite{martins2024unravelling}) or consider the components along and across the coastline separtely (e.g. \citet{bourg2022driving}). The word 'Circular data' refers to points or directions represented on a unit circle \citep{crujeiras2017introduction}. This type of data is commonly encountered in various natural and physical sciences, particularly in marine sciences, where understanding ocean currents and wind patterns is crucial for marine operations, such as navigation, search and rescue missions, and the dispersion of pollutants in the ocean \citep{oliveira2014circsizer}. 
\cite{oliveira2014circsizer} stated that estimating the density of circular data, as well as performing regression when the exploratory variable is circular, presents a significant challenge and interest across various applied fields. 
\cite{zulkipli2021synthetic} generated synthetic data for univariate circular data following the von Mises distribution \citep{wood1994simulation}, whereas  we employed CT-GAN to generate synthetic circular data that captures the distribution of the real dataset and utilized machine learning models capable of capturing the nonlinear (circular nature) relationship within the dataset.




Despite the findings of this study, it is crucial to acknowledge its limitations. Firstly, the dataset lacks information on the population dynamic, biological and life cycle of bluebottles offshore before drifting to the coast. This omission contributes to class overlap and limits the amount of information available for modelling. Consequently, the environmental factors included in the dataset are insufficient for building highly accurate predictive models, as the data shows that bluebottle presence and absence can occur under similar environmental conditions. Indeed, while we know that northeasterly and southerly winds bring bluebottles to shore in the region \citep{bourg2022driving, bourg2024ocean}, there are many instances with the same conditions, but not bluebottle beaching, for instance if the population is further in the ocean or less abundant due to their life-cycle. As noted by 
\cite{castro2024using}, while the absence of bluebottles provides valuable insights, it comes with inherent limitations, as the absence recorded in the data may not always represent a true absence caused by to the explanatory variables available. 
Another limitation is the seasonality influence of the dataset. The presence and absence of the bluebottle beaching is strongly influenced by seasonality as the probability of occurrence or non-occurrence fluctuates depending on the time of the year as stated by \cite{bourg2022driving} and also shown in this study. The seasonal nature of the phenomenon presents an opportunity to improve future models such as considering time lag or using deep learning time series model like long short term memory (LSTM) as was done by \cite{kapoor2023cyclone} in the cyclone prediction. 



Given the seasonal nature of bluebottle occurrence, 
\cite{pontin2009using} enhanced jellyfish prediction by utilising time-lagged input data. The accuracy of bluebottle prediction could be further improved through spatio-temporal modelling, particularly if migration data were available, as demonstrated by 
\cite{martins2022modelling}, instead of relying solely on beaching data (data on when and where bluebottles are washed ashore). Future studies should investigate the integration of spatial and temporal patterns of bluebottle movement from their oceanic origins to the coastline as this would offer a more comprehensive understanding of their dynamics, including migration routes, timing, and environmental influences.
Moreover, other research could explore the uncertainty inherent in the data by integrating Bayesian deep learning \citep{lansner1989one} with synthetic data generation techniques to address unreliable negative labels. This approach combines Bayesian deep learning with synthetic data methods to effectively manage unreliable labels and quantify uncertainties. Such approaches would facilitate uncertainty quantification and projection in the predictions during the synthetic data generation process, which was incorporated by 
\cite{kapoor2023cyclone} in the prediction of cyclone track and wind-intensity. Finally, imputation of both linear and circular data using variational Bayesian deep learning \citep{kulkarni2024bayes} could be explored as a means to improve bluebottle occurrence prediction.   Finally, to improve the accuracy of predictions regarding bluebottle presence, incorporating temporal and spatial patterns could help address the lack of understanding of critical factors such as organism population dynamics and life cycle. These results align with known physical processes that contribute to bluebottle presence along the east coast of Australia. Still, they also introduce, for the first time, a successful machine learning methodology to accurately predict both presence and absence on a daily timescale, significantly reducing false positives inherent in this type of environmental data. This represents a crucial step toward developing a marine stingers prediction system for beach-goers, where false alarms must be avoided to maintain public trust and ensure an effective response.



\section{Conclusions}

In this paper, we explored selected data augmentation techniques for addressing class imbalance, class overlap and the challenge of unreliable negative class data for the prediction of bluebottle marine beaching. We evaluated the impact of no data augmentation versus three data augmentation techniques, including a synthetic negative approach that completely discarded the available negative data. The circular nature of some of the features such as wind and current direction was taken into consideration during synthetic data generation and modelling as we employed machine learning models capable of capturing the nonlinear (circular nature) relationship. Additionally, we identified the environmental factors that drive bluebottle beaching and used PCA for data visualisation, demonstrating the effects of each data augmentation technique.

Our results demonstrated that the commonly used data augmentation technique (SMOTE) was ineffective in addressing class overlap and the unreliability of the negative class, even after balancing the dataset. In contrast, the synthetic negative approach and Random Undersampling approach successfully managed both class imbalance and overlap, as well as the unreliable negative data. Among the three machine learning models, Random Forest outperformed the others, particularly when using the synthetic negative approach. We identified wind direction, especially from the North and South, as a key factor influencing bluebottle migration. Additionally, during February, when sea surface temperatures are particularly high, there is an increased likelihood of bluebottles washing up on the beach.



\section{Acknowledgemt}

We would like to thank Daniel E. Hewitt from UNSW Sydney. 
\section*{Data and Code Availability}
We have made the code and data for our proposed methodology openly available through a GitHub repository, accompanied by detailed descriptive information.
\\ - Name: bluebottle-xai
\\ - Developer: Amuche Ibenegbu
\\ - Contact email: amucheibenegbu@gmail.com
\\ - Compatible Operating System: Windows/Mac/Linux 
\\ - Developed and tested: Windows 11 Enterprise, Version 23H2
\\ - Size of repository: 13.1 MB
\\ - Year Published: 2024
\\ - Source: GitHub
\\ - repository link: \url{https://github.com/DARE-ML/bluebottle-xai}



 \bibliographystyle{elsarticle-harv}
 \bibliography{cas-refs}

\begin{thebibliography}{123}
\expandafter\ifx\csname natexlab\endcsname\relax\def\natexlab#1{#1}\fi
\providecommand{\url}[1]{\texttt{#1}}
\providecommand{\href}[2]{#2}
\providecommand{\path}[1]{#1}
\providecommand{\DOIprefix}{doi:}
\providecommand{\ArXivprefix}{arXiv:}
\providecommand{\URLprefix}{URL: }
\providecommand{\Pubmedprefix}{pmid:}
\providecommand{\doi}[1]{\href{http://dx.doi.org/#1}{\path{#1}}}
\providecommand{\Pubmed}[1]{\href{pmid:#1}{\path{#1}}}
\providecommand{\bibinfo}[2]{#2}
\ifx\xfnm\relax \def\xfnm[#1]{\unskip,\space#1}\fi
\bibitem[{Abdi and Williams(2010)}]{abdi2010principal}
\bibinfo{author}{Abdi, H.}, \bibinfo{author}{Williams, L.J.}, \bibinfo{year}{2010}.
\newblock \bibinfo{title}{Principal component analysis}.
\newblock \bibinfo{journal}{Wiley interdisciplinary reviews: computational statistics} \bibinfo{volume}{2}, \bibinfo{pages}{433--459}.
\bibitem[{Albajes-Eizagirre et~al.(2011)Albajes-Eizagirre, Romero, Soria-Frisch and Vanhellemont}]{albajes2011jellyfish}
\bibinfo{author}{Albajes-Eizagirre, A.}, \bibinfo{author}{Romero, L.}, \bibinfo{author}{Soria-Frisch, A.}, \bibinfo{author}{Vanhellemont, Q.}, \bibinfo{year}{2011}.
\newblock \bibinfo{title}{Jellyfish prediction of occurrence from remote sensing data and a non-linear pattern recognition approach}, in: \bibinfo{booktitle}{Remote Sensing for Agriculture, Ecosystems, and Hydrology XIII}, \bibinfo{organization}{SPIE}. pp. \bibinfo{pages}{382--391}.
\bibitem[{Anand et~al.(1993)Anand, Mehrotra, Mohan and Ranka}]{anand1993improved}
\bibinfo{author}{Anand, R.}, \bibinfo{author}{Mehrotra, K.G.}, \bibinfo{author}{Mohan, C.K.}, \bibinfo{author}{Ranka, S.}, \bibinfo{year}{1993}.
\newblock \bibinfo{title}{An improved algorithm for neural network classification of imbalanced training sets}.
\newblock \bibinfo{journal}{IEEE transactions on neural networks} \bibinfo{volume}{4}, \bibinfo{pages}{962--969}.
\bibitem[{Azhar et~al.(2015)Azhar, Aris, Yusoff, Ramli and Juahir}]{azhar2015classification}
\bibinfo{author}{Azhar, S.C.}, \bibinfo{author}{Aris, A.Z.}, \bibinfo{author}{Yusoff, M.K.}, \bibinfo{author}{Ramli, M.F.}, \bibinfo{author}{Juahir, H.}, \bibinfo{year}{2015}.
\newblock \bibinfo{title}{Classification of river water quality using multivariate analysis}.
\newblock \bibinfo{journal}{Procedia Environmental Sciences} \bibinfo{volume}{30}, \bibinfo{pages}{79--84}.
\bibitem[{Bach et~al.(2019)Bach, Werner and Palt}]{bach2019proposal}
\bibinfo{author}{Bach, M.}, \bibinfo{author}{Werner, A.}, \bibinfo{author}{Palt, M.}, \bibinfo{year}{2019}.
\newblock \bibinfo{title}{The proposal of undersampling method for learning from imbalanced datasets}.
\newblock \bibinfo{journal}{Procedia Computer Science} \bibinfo{volume}{159}, \bibinfo{pages}{125--134}.
\bibitem[{Bayer et~al.(2022)Bayer, Kaufhold and Reuter}]{bayer2022survey}
\bibinfo{author}{Bayer, M.}, \bibinfo{author}{Kaufhold, M.A.}, \bibinfo{author}{Reuter, C.}, \bibinfo{year}{2022}.
\newblock \bibinfo{title}{A survey on data augmentation for text classification}.
\newblock \bibinfo{journal}{ACM Computing Surveys} \bibinfo{volume}{55}, \bibinfo{pages}{1--39}.
\bibitem[{Begam and Kumar(2014)}]{begam2014visualization}
\bibinfo{author}{Begam, B.F.}, \bibinfo{author}{Kumar, J.S.}, \bibinfo{year}{2014}.
\newblock \bibinfo{title}{Visualization of chemical space using principal component analysis}.
\newblock \bibinfo{journal}{World Applied Sciences Journal} \bibinfo{volume}{29}, \bibinfo{pages}{53--59}.
\bibitem[{Bekkar et~al.(2013)Bekkar, Djemaa and Alitouche}]{bekkar2013evaluation}
\bibinfo{author}{Bekkar, M.}, \bibinfo{author}{Djemaa, H.K.}, \bibinfo{author}{Alitouche, T.A.}, \bibinfo{year}{2013}.
\newblock \bibinfo{title}{Evaluation measures for models assessment over imbalanced data sets}.
\newblock \bibinfo{journal}{J Inf Eng Appl} \bibinfo{volume}{3}.
\bibitem[{Bellido et~al.(2020)Bellido, Baez, Souviron-Priego, Ferri-Yanez, Salas, L{\'o}pez and Real}]{bellido2020atmospheric}
\bibinfo{author}{Bellido, J.J.}, \bibinfo{author}{Baez, J.C.}, \bibinfo{author}{Souviron-Priego, L.}, \bibinfo{author}{Ferri-Yanez, F.}, \bibinfo{author}{Salas, C.}, \bibinfo{author}{L{\'o}pez, J.A.}, \bibinfo{author}{Real, R.}, \bibinfo{year}{2020}.
\newblock \bibinfo{title}{Atmospheric indices allow anticipating the incidence of jellyfish coastal swarms}.
\newblock \bibinfo{journal}{Mediterranean Marine Science} \bibinfo{volume}{21}, \bibinfo{pages}{289--297}.
\bibitem[{Borowiec et~al.(2022)Borowiec, Dikow, Frandsen, McKeeken, Valentini and White}]{borowiec2022deep}
\bibinfo{author}{Borowiec, M.L.}, \bibinfo{author}{Dikow, R.B.}, \bibinfo{author}{Frandsen, P.B.}, \bibinfo{author}{McKeeken, A.}, \bibinfo{author}{Valentini, G.}, \bibinfo{author}{White, A.E.}, \bibinfo{year}{2022}.
\newblock \bibinfo{title}{Deep learning as a tool for ecology and evolution}.
\newblock \bibinfo{journal}{Methods in Ecology and Evolution} \bibinfo{volume}{13}, \bibinfo{pages}{1640--1660}.
\bibitem[{Bourg et~al.(2022)Bourg, Schaeffer, Cetina-Heredia, Lawes and Lee}]{bourg2022driving}
\bibinfo{author}{Bourg, N.}, \bibinfo{author}{Schaeffer, A.}, \bibinfo{author}{Cetina-Heredia, P.}, \bibinfo{author}{Lawes, J.C.}, \bibinfo{author}{Lee, D.}, \bibinfo{year}{2022}.
\newblock \bibinfo{title}{Driving the blue fleet: Temporal variability and drivers behind bluebottle (physalia physalis) beachings off sydney, australia}.
\newblock \bibinfo{journal}{Plos one} \bibinfo{volume}{17}, \bibinfo{pages}{e0265593}.
\bibitem[{Bourg et~al.(2024)Bourg, Schaeffer, Molcard, Luneau, Hewitt and Chemin}]{bourg2024ocean}
\bibinfo{author}{Bourg, N.}, \bibinfo{author}{Schaeffer, A.}, \bibinfo{author}{Molcard, A.}, \bibinfo{author}{Luneau, C.}, \bibinfo{author}{Hewitt, D.E.}, \bibinfo{author}{Chemin, R.}, \bibinfo{year}{2024}.
\newblock \bibinfo{title}{Ocean wanderers: A lab-based investigation into the effect of wind and morphology on the drift of physalia spp.}
\newblock \bibinfo{journal}{Marine Pollution Bulletin} \bibinfo{volume}{207}, \bibinfo{pages}{116856}.
\bibitem[{Breiman(2001)}]{breiman2001random}
\bibinfo{author}{Breiman, L.}, \bibinfo{year}{2001}.
\newblock \bibinfo{title}{Random forests}.
\newblock \bibinfo{journal}{Machine learning} \bibinfo{volume}{45}, \bibinfo{pages}{5--32}.
\bibitem[{Bryant and Shreeve(2002)}]{bryant2002use}
\bibinfo{author}{Bryant, S.R.}, \bibinfo{author}{Shreeve, T.G.}, \bibinfo{year}{2002}.
\newblock \bibinfo{title}{The use of artificial neural networks in ecological analysis: estimating microhabitat temperature}.
\newblock \bibinfo{journal}{Ecological Entomology} \bibinfo{volume}{27}, \bibinfo{pages}{424--432}.
\bibitem[{Canepa et~al.(2014)Canepa, Fuentes, Sabat{\'e}s, Piraino, Boero and Gili}]{canepa2014pelagia}
\bibinfo{author}{Canepa, A.}, \bibinfo{author}{Fuentes, V.}, \bibinfo{author}{Sabat{\'e}s, A.}, \bibinfo{author}{Piraino, S.}, \bibinfo{author}{Boero, F.}, \bibinfo{author}{Gili, J.M.}, \bibinfo{year}{2014}.
\newblock \bibinfo{title}{Pelagia noctiluca in the mediterranean sea}.
\newblock \bibinfo{journal}{Jellyfish blooms} , \bibinfo{pages}{237--266}.
\bibitem[{Castro-Guti{\'e}rrez et~al.(2024)Castro-Guti{\'e}rrez, Guti{\'e}rrez-Estrada and B{\'a}ez}]{castro2024using}
\bibinfo{author}{Castro-Guti{\'e}rrez, J.}, \bibinfo{author}{Guti{\'e}rrez-Estrada, J.}, \bibinfo{author}{B{\'a}ez, J.}, \bibinfo{year}{2024}.
\newblock \bibinfo{title}{Using artificial neural networks and citizen science data to assess jellyfish presence along coastal areas}.
\newblock \bibinfo{journal}{Journal of Applied Ecology} .
\bibitem[{Cegolon et~al.(2013)Cegolon, Heymann, Lange and Mastrangelo}]{cegolon2013jellyfish}
\bibinfo{author}{Cegolon, L.}, \bibinfo{author}{Heymann, W.C.}, \bibinfo{author}{Lange, J.H.}, \bibinfo{author}{Mastrangelo, G.}, \bibinfo{year}{2013}.
\newblock \bibinfo{title}{Jellyfish stings and their management: a review}.
\newblock \bibinfo{journal}{Marine drugs} \bibinfo{volume}{11}, \bibinfo{pages}{523--550}.
\bibitem[{Chamberlain et~al.(2021)Chamberlain, Oke, Fiedler, Beggs, Brassington and Divakaran}]{chamberlain2021next}
\bibinfo{author}{Chamberlain, M.A.}, \bibinfo{author}{Oke, P.R.}, \bibinfo{author}{Fiedler, R.A.}, \bibinfo{author}{Beggs, H.M.}, \bibinfo{author}{Brassington, G.B.}, \bibinfo{author}{Divakaran, P.}, \bibinfo{year}{2021}.
\newblock \bibinfo{title}{Next generation of bluelink ocean reanalysis with multiscale data assimilation: Bran2020}.
\newblock \bibinfo{journal}{Earth System Science Data} \bibinfo{volume}{13}, \bibinfo{pages}{5663--5688}.
\bibitem[{Chandra et~al.(2023)Chandra, Bansal, Kang, Blau, Agarwal, Singh, Wilson and Vasan}]{chandra2023unsupervised}
\bibinfo{author}{Chandra, R.}, \bibinfo{author}{Bansal, C.}, \bibinfo{author}{Kang, M.}, \bibinfo{author}{Blau, T.}, \bibinfo{author}{Agarwal, V.}, \bibinfo{author}{Singh, P.}, \bibinfo{author}{Wilson, L.O.}, \bibinfo{author}{Vasan, S.}, \bibinfo{year}{2023}.
\newblock \bibinfo{title}{Unsupervised machine learning framework for discriminating major variants of concern during covid-19}.
\newblock \bibinfo{journal}{Plos one} \bibinfo{volume}{18}, \bibinfo{pages}{e0285719}.
\bibitem[{Chawla et~al.(2002)Chawla, Bowyer, Hall and Kegelmeyer}]{chawla2002smote}
\bibinfo{author}{Chawla, N.V.}, \bibinfo{author}{Bowyer, K.W.}, \bibinfo{author}{Hall, L.O.}, \bibinfo{author}{Kegelmeyer, W.P.}, \bibinfo{year}{2002}.
\newblock \bibinfo{title}{Smote: synthetic minority over-sampling technique}.
\newblock \bibinfo{journal}{Journal of artificial intelligence research} \bibinfo{volume}{16}, \bibinfo{pages}{321--357}.
\bibitem[{Cheah et~al.(2023)Cheah, Yang and Lee}]{cheah2023enhancing}
\bibinfo{author}{Cheah, P.C.Y.}, \bibinfo{author}{Yang, Y.}, \bibinfo{author}{Lee, B.G.}, \bibinfo{year}{2023}.
\newblock \bibinfo{title}{Enhancing financial fraud detection through addressing class imbalance using hybrid smote-gan techniques}.
\newblock \bibinfo{journal}{International Journal of Financial Studies} \bibinfo{volume}{11}, \bibinfo{pages}{110}.
\bibitem[{Chen and Guestrin(2016)}]{chen2016xgboost}
\bibinfo{author}{Chen, T.}, \bibinfo{author}{Guestrin, C.}, \bibinfo{year}{2016}.
\newblock \bibinfo{title}{Xgboost: A scalable tree boosting system}, in: \bibinfo{booktitle}{Proceedings of the 22nd acm sigkdd international conference on knowledge discovery and data mining}, pp. \bibinfo{pages}{785--794}.
\bibitem[{Chen and Wu(2017)}]{chen2017application}
\bibinfo{author}{Chen, Y.}, \bibinfo{author}{Wu, W.}, \bibinfo{year}{2017}.
\newblock \bibinfo{title}{Application of one-class support vector machine to quickly identify multivariate anomalies from geochemical exploration data}.
\newblock \bibinfo{journal}{Geochemistry: Exploration, Environment, Analysis} \bibinfo{volume}{17}, \bibinfo{pages}{231--238}.
\bibitem[{Church et~al.(2024)Church, Abedon, Ahuja, Anthony, Ramirez, Rojas, Albinsson, {\'A}lvarez~Trasobares, Bergemann, Bogdanovic et~al.}]{church2024global}
\bibinfo{author}{Church, S.H.}, \bibinfo{author}{Abedon, R.B.}, \bibinfo{author}{Ahuja, N.}, \bibinfo{author}{Anthony, C.J.}, \bibinfo{author}{Ramirez, D.A.}, \bibinfo{author}{Rojas, L.M.}, \bibinfo{author}{Albinsson, M.E.}, \bibinfo{author}{{\'A}lvarez~Trasobares, I.}, \bibinfo{author}{Bergemann, R.E.}, \bibinfo{author}{Bogdanovic, O.}, et~al., \bibinfo{year}{2024}.
\newblock \bibinfo{title}{Global genomics of the man-o’-war (physalia) reveals biodiversity at the ocean surface}.
\newblock \bibinfo{journal}{bioRxiv} , \bibinfo{pages}{2024--07}.
\bibitem[{Clements et~al.(2024)Clements, Thompson, Hannoun and Dickenson}]{clements2024classification}
\bibinfo{author}{Clements, E.}, \bibinfo{author}{Thompson, K.A.}, \bibinfo{author}{Hannoun, D.}, \bibinfo{author}{Dickenson, E.R.}, \bibinfo{year}{2024}.
\newblock \bibinfo{title}{Classification machine learning to detect de facto reuse and cyanobacteria at a drinking water intake}.
\newblock \bibinfo{journal}{Science of the Total Environment} \bibinfo{volume}{948}, \bibinfo{pages}{174690}.
\bibitem[{Coombs et~al.(2019)Coombs, Deaville, Sabin, Allan, O'Connell, Berrow, Smith, Brownlow, Doeschate, Penrose et~al.}]{coombs2019can}
\bibinfo{author}{Coombs, E.J.}, \bibinfo{author}{Deaville, R.}, \bibinfo{author}{Sabin, R.C.}, \bibinfo{author}{Allan, L.}, \bibinfo{author}{O'Connell, M.}, \bibinfo{author}{Berrow, S.}, \bibinfo{author}{Smith, B.}, \bibinfo{author}{Brownlow, A.}, \bibinfo{author}{Doeschate, M.T.}, \bibinfo{author}{Penrose, R.}, et~al., \bibinfo{year}{2019}.
\newblock \bibinfo{title}{What can cetacean stranding records tell us? a study of uk and {Irish} cetacean diversity over the past 100 years}.
\newblock \bibinfo{journal}{Marine Mammal Science} \bibinfo{volume}{35}, \bibinfo{pages}{1527--1555}.
\bibitem[{Crujeiras(2017)}]{crujeiras2017introduction}
\bibinfo{author}{Crujeiras, R.M.}, \bibinfo{year}{2017}.
\newblock \bibinfo{title}{An introduction to statistical methods for circular data}.
\newblock \bibinfo{journal}{Boletin de Estadistica e Investigacion Operativa} \bibinfo{volume}{33}, \bibinfo{pages}{85--107}.
\bibitem[{Cutler et~al.(2007)Cutler, Edwards~Jr, Beard, Cutler, Hess, Gibson and Lawler}]{cutler2007random}
\bibinfo{author}{Cutler, D.R.}, \bibinfo{author}{Edwards~Jr, T.C.}, \bibinfo{author}{Beard, K.H.}, \bibinfo{author}{Cutler, A.}, \bibinfo{author}{Hess, K.T.}, \bibinfo{author}{Gibson, J.}, \bibinfo{author}{Lawler, J.J.}, \bibinfo{year}{2007}.
\newblock \bibinfo{title}{Random forests for classification in ecology}.
\newblock \bibinfo{journal}{Ecology} \bibinfo{volume}{88}, \bibinfo{pages}{2783--2792}.
\bibitem[{Duarte et~al.(2020)Duarte, Andriolo and Gon{\c{c}}alves}]{duarte2020addressing}
\bibinfo{author}{Duarte, D.}, \bibinfo{author}{Andriolo, U.}, \bibinfo{author}{Gon{\c{c}}alves, G.}, \bibinfo{year}{2020}.
\newblock \bibinfo{title}{Addressing the class imbalance problem in the automatic image classification of coastal litter from orthophotos derived from uas imagery}.
\newblock \bibinfo{journal}{ISPRS Annals of the Photogrammetry, Remote Sensing and Spatial Information Sciences} \bibinfo{volume}{3}, \bibinfo{pages}{439--445}.
\bibitem[{Farahbakhsh et~al.(2023)Farahbakhsh, Maughan and M{\"u}ller}]{farahbakhsh2023prospectivity}
\bibinfo{author}{Farahbakhsh, E.}, \bibinfo{author}{Maughan, J.}, \bibinfo{author}{M{\"u}ller, R.D.}, \bibinfo{year}{2023}.
\newblock \bibinfo{title}{Prospectivity modelling of critical mineral deposits using a generative adversarial network with oversampling and positive-unlabelled bagging}.
\newblock \bibinfo{journal}{Ore Geology Reviews} , \bibinfo{pages}{105665}.
\bibitem[{Fern{\'a}ndez et~al.(2018)Fern{\'a}ndez, Garcia, Herrera and Chawla}]{fernandez2018smote}
\bibinfo{author}{Fern{\'a}ndez, A.}, \bibinfo{author}{Garcia, S.}, \bibinfo{author}{Herrera, F.}, \bibinfo{author}{Chawla, N.V.}, \bibinfo{year}{2018}.
\newblock \bibinfo{title}{Smote for learning from imbalanced data: progress and challenges, marking the 15-year anniversary}.
\newblock \bibinfo{journal}{Journal of artificial intelligence research} \bibinfo{volume}{61}, \bibinfo{pages}{863--905}.
\bibitem[{Ferreira and Figueiredo(2012)}]{ferreira2012boosting}
\bibinfo{author}{Ferreira, A.J.}, \bibinfo{author}{Figueiredo, M.A.}, \bibinfo{year}{2012}.
\newblock \bibinfo{title}{Boosting algorithms: A review of methods, theory, and applications}.
\newblock \bibinfo{journal}{Ensemble machine learning: Methods and applications} , \bibinfo{pages}{35--85}.
\bibitem[{Ferrer and Pastor(2017)}]{ferrer2017portuguese}
\bibinfo{author}{Ferrer, L.}, \bibinfo{author}{Pastor, A.}, \bibinfo{year}{2017}.
\newblock \bibinfo{title}{The portuguese man-of-war: Gone with the wind}.
\newblock \bibinfo{journal}{Regional Studies in Marine Science} \bibinfo{volume}{14}, \bibinfo{pages}{53--62}.
\bibitem[{Fung et~al.(2005)Fung, Yu, Lu and Yu}]{fung2005text}
\bibinfo{author}{Fung, G.P.C.}, \bibinfo{author}{Yu, J.X.}, \bibinfo{author}{Lu, H.}, \bibinfo{author}{Yu, P.S.}, \bibinfo{year}{2005}.
\newblock \bibinfo{title}{Text classification without negative examples revisit}.
\newblock \bibinfo{journal}{IEEE transactions on Knowledge and Data Engineering} \bibinfo{volume}{18}, \bibinfo{pages}{6--20}.
\bibitem[{Galar et~al.(2011)Galar, Fernandez, Barrenechea, Bustince and Herrera}]{galar2011review}
\bibinfo{author}{Galar, M.}, \bibinfo{author}{Fernandez, A.}, \bibinfo{author}{Barrenechea, E.}, \bibinfo{author}{Bustince, H.}, \bibinfo{author}{Herrera, F.}, \bibinfo{year}{2011}.
\newblock \bibinfo{title}{A review on ensembles for the class imbalance problem: bagging-, boosting-, and hybrid-based approaches}.
\newblock \bibinfo{journal}{IEEE Transactions on Systems, Man, and Cybernetics, Part C (Applications and Reviews)} \bibinfo{volume}{42}, \bibinfo{pages}{463--484}.
\bibitem[{Ghosh et~al.(2024)Ghosh, Bellinger, Corizzo, Branco, Krawczyk and Japkowicz}]{ghosh2024class}
\bibinfo{author}{Ghosh, K.}, \bibinfo{author}{Bellinger, C.}, \bibinfo{author}{Corizzo, R.}, \bibinfo{author}{Branco, P.}, \bibinfo{author}{Krawczyk, B.}, \bibinfo{author}{Japkowicz, N.}, \bibinfo{year}{2024}.
\newblock \bibinfo{title}{The class imbalance problem in deep learning}.
\newblock \bibinfo{journal}{Machine Learning} \bibinfo{volume}{113}, \bibinfo{pages}{4845--4901}.
\bibitem[{Goodfellow et~al.(2014)Goodfellow, Pouget-Abadie, Mirza, Xu, Warde-Farley, Ozair, Courville and Bengio}]{goodfellow2014generative}
\bibinfo{author}{Goodfellow, I.}, \bibinfo{author}{Pouget-Abadie, J.}, \bibinfo{author}{Mirza, M.}, \bibinfo{author}{Xu, B.}, \bibinfo{author}{Warde-Farley, D.}, \bibinfo{author}{Ozair, S.}, \bibinfo{author}{Courville, A.}, \bibinfo{author}{Bengio, Y.}, \bibinfo{year}{2014}.
\newblock \bibinfo{title}{Generative adversarial nets}.
\newblock \bibinfo{journal}{Advances in neural information processing systems} \bibinfo{volume}{27}.
\bibitem[{Goodfellow et~al.(2020)Goodfellow, Pouget-Abadie, Mirza, Xu, Warde-Farley, Ozair, Courville and Bengio}]{goodfellow2020generative}
\bibinfo{author}{Goodfellow, I.}, \bibinfo{author}{Pouget-Abadie, J.}, \bibinfo{author}{Mirza, M.}, \bibinfo{author}{Xu, B.}, \bibinfo{author}{Warde-Farley, D.}, \bibinfo{author}{Ozair, S.}, \bibinfo{author}{Courville, A.}, \bibinfo{author}{Bengio, Y.}, \bibinfo{year}{2020}.
\newblock \bibinfo{title}{Generative adversarial networks}.
\newblock \bibinfo{journal}{Communications of the ACM} \bibinfo{volume}{63}, \bibinfo{pages}{139--144}.
\bibitem[{Gu et~al.(2009)Gu, Zhu and Cai}]{gu2009evaluation}
\bibinfo{author}{Gu, Q.}, \bibinfo{author}{Zhu, L.}, \bibinfo{author}{Cai, Z.}, \bibinfo{year}{2009}.
\newblock \bibinfo{title}{Evaluation measures of the classification performance of imbalanced data sets}, in: \bibinfo{booktitle}{Computational Intelligence and Intelligent Systems: 4th International Symposium, ISICA 2009, Huangshi, China, October 23-25, 2009. Proceedings 4}, \bibinfo{organization}{Springer}. pp. \bibinfo{pages}{461--471}.
\bibitem[{Haixiang et~al.(2017)Haixiang, Yijing, Shang, Mingyun, Yuanyue and Bing}]{haixiang2017learning}
\bibinfo{author}{Haixiang, G.}, \bibinfo{author}{Yijing, L.}, \bibinfo{author}{Shang, J.}, \bibinfo{author}{Mingyun, G.}, \bibinfo{author}{Yuanyue, H.}, \bibinfo{author}{Bing, G.}, \bibinfo{year}{2017}.
\newblock \bibinfo{title}{Learning from class-imbalanced data: Review of methods and applications}.
\newblock \bibinfo{journal}{Expert systems with applications} \bibinfo{volume}{73}, \bibinfo{pages}{220--239}.
\bibitem[{Han et~al.(2018)Han, Hayashi, Rundo, Araki, Shimoda, Muramatsu, Furukawa, Mauri and Nakayama}]{han2018gan}
\bibinfo{author}{Han, C.}, \bibinfo{author}{Hayashi, H.}, \bibinfo{author}{Rundo, L.}, \bibinfo{author}{Araki, R.}, \bibinfo{author}{Shimoda, W.}, \bibinfo{author}{Muramatsu, S.}, \bibinfo{author}{Furukawa, Y.}, \bibinfo{author}{Mauri, G.}, \bibinfo{author}{Nakayama, H.}, \bibinfo{year}{2018}.
\newblock \bibinfo{title}{Gan-based synthetic brain mr image generation}, in: \bibinfo{booktitle}{2018 IEEE 15th international symposium on biomedical imaging (ISBI 2018)}, \bibinfo{organization}{IEEE}. pp. \bibinfo{pages}{734--738}.
\bibitem[{Haque and Tozal(2022)}]{haque2022negative}
\bibinfo{author}{Haque, M.E.}, \bibinfo{author}{Tozal, M.E.}, \bibinfo{year}{2022}.
\newblock \bibinfo{title}{Negative insurance claim generation using distance pooling on positive diagnosis-procedure bipartite graphs}.
\newblock \bibinfo{journal}{ACM Journal of Data and Information Quality (JDIQ)} \bibinfo{volume}{14}, \bibinfo{pages}{1--26}.
\bibitem[{Hejazi and Singh(2013)}]{hejazi2013one}
\bibinfo{author}{Hejazi, M.}, \bibinfo{author}{Singh, Y.P.}, \bibinfo{year}{2013}.
\newblock \bibinfo{title}{One-class support vector machines approach to anomaly detection}.
\newblock \bibinfo{journal}{Applied Artificial Intelligence} \bibinfo{volume}{27}, \bibinfo{pages}{351--366}.
\bibitem[{Helm(2021)}]{helm2021mysterious}
\bibinfo{author}{Helm, R.R.}, \bibinfo{year}{2021}.
\newblock \bibinfo{title}{The mysterious ecosystem at the ocean’s surface}.
\newblock \bibinfo{journal}{PLoS Biology} \bibinfo{volume}{19}, \bibinfo{pages}{e3001046}.
\bibitem[{Hochreiter(1997)}]{hochreiter1997long}
\bibinfo{author}{Hochreiter, S.}, \bibinfo{year}{1997}.
\newblock \bibinfo{title}{Long short-term memory}.
\newblock \bibinfo{journal}{Neural Computation MIT-Press} .
\bibitem[{Hong et~al.(2020)Hong, Kang and Hong}]{hong2020oversampling}
\bibinfo{author}{Hong, J.}, \bibinfo{author}{Kang, H.}, \bibinfo{author}{Hong, T.}, \bibinfo{year}{2020}.
\newblock \bibinfo{title}{Oversampling-based prediction of environmental complaints related to construction projects with imbalanced empirical-data learning}.
\newblock \bibinfo{journal}{Renewable and Sustainable Energy Reviews} \bibinfo{volume}{134}, \bibinfo{pages}{110402}.
\bibitem[{Howley et~al.(2005)Howley, Madden, O’Connell and Ryder}]{howley2005effect}
\bibinfo{author}{Howley, T.}, \bibinfo{author}{Madden, M.G.}, \bibinfo{author}{O’Connell, M.L.}, \bibinfo{author}{Ryder, A.G.}, \bibinfo{year}{2005}.
\newblock \bibinfo{title}{The effect of principal component analysis on machine learning accuracy with high dimensional spectral data}, in: \bibinfo{booktitle}{International Conference on Innovative Techniques and Applications of Artificial Intelligence}, \bibinfo{organization}{Springer}. pp. \bibinfo{pages}{209--222}.
\bibitem[{Hussein et~al.(2019)Hussein, Li, Yohannese and Bashir}]{hussein2019smote}
\bibinfo{author}{Hussein, A.S.}, \bibinfo{author}{Li, T.}, \bibinfo{author}{Yohannese, C.W.}, \bibinfo{author}{Bashir, K.}, \bibinfo{year}{2019}.
\newblock \bibinfo{title}{A-smote: A new preprocessing approach for highly imbalanced datasets by improving smote}.
\newblock \bibinfo{journal}{International Journal of Computational Intelligence Systems} \bibinfo{volume}{12}, \bibinfo{pages}{1412--1422}.
\bibitem[{iNaturalist(n.d)}]{bluebottles_australia}
\bibinfo{author}{iNaturalist}, \bibinfo{year}{n.d}.
\newblock \bibinfo{title}{Bluebottles inaustralia (physalia physalis)}.
\newblock \URLprefix \url{https://www.inaturalist.org/projects/bluebottles-in-australia-physalia-physalis}. \bibinfo{note}{accessed: 2025-01-15}.
\bibitem[{Joloudari et~al.(2023)Joloudari, Marefat, Nematollahi, Oyelere and Hussain}]{joloudari2023effective}
\bibinfo{author}{Joloudari, J.H.}, \bibinfo{author}{Marefat, A.}, \bibinfo{author}{Nematollahi, M.A.}, \bibinfo{author}{Oyelere, S.S.}, \bibinfo{author}{Hussain, S.}, \bibinfo{year}{2023}.
\newblock \bibinfo{title}{Effective class-imbalance learning based on smote and convolutional neural networks}.
\newblock \bibinfo{journal}{Applied Sciences} \bibinfo{volume}{13}, \bibinfo{pages}{4006}.
\bibitem[{Junsomboon and Phienthrakul(2017)}]{junsomboon2017combining}
\bibinfo{author}{Junsomboon, N.}, \bibinfo{author}{Phienthrakul, T.}, \bibinfo{year}{2017}.
\newblock \bibinfo{title}{Combining over-sampling and under-sampling techniques for imbalance dataset}, in: \bibinfo{booktitle}{Proceedings of the 9th international conference on machine learning and computing}, pp. \bibinfo{pages}{243--247}.
\bibitem[{Kapoor et~al.(2023)Kapoor, Negi, Marshall and Chandra}]{kapoor2023cyclone}
\bibinfo{author}{Kapoor, A.}, \bibinfo{author}{Negi, A.}, \bibinfo{author}{Marshall, L.}, \bibinfo{author}{Chandra, R.}, \bibinfo{year}{2023}.
\newblock \bibinfo{title}{Cyclone trajectory and intensity prediction with uncertainty quantification using variational recurrent neural networks}.
\newblock \bibinfo{journal}{Environmental Modelling \& Software} \bibinfo{volume}{162}, \bibinfo{pages}{105654}.
\bibitem[{Kaur and Gosain(2018)}]{kaur2018comparing}
\bibinfo{author}{Kaur, P.}, \bibinfo{author}{Gosain, A.}, \bibinfo{year}{2018}.
\newblock \bibinfo{title}{Comparing the behavior of oversampling and undersampling approach of class imbalance learning by combining class imbalance problem with noise}, in: \bibinfo{booktitle}{ICT Based Innovations: Proceedings of CSI 2015}, \bibinfo{organization}{Springer}. pp. \bibinfo{pages}{23--30}.
\bibitem[{Khalilia et~al.(2011)Khalilia, Chakraborty and Popescu}]{khalilia2011predicting}
\bibinfo{author}{Khalilia, M.}, \bibinfo{author}{Chakraborty, S.}, \bibinfo{author}{Popescu, M.}, \bibinfo{year}{2011}.
\newblock \bibinfo{title}{Predicting disease risks from highly imbalanced data using random forest}.
\newblock \bibinfo{journal}{BMC medical informatics and decision making} \bibinfo{volume}{11}, \bibinfo{pages}{1--13}.
\bibitem[{Khan et~al.(2023)Khan, Chaudhari and Chandra}]{khan2023review}
\bibinfo{author}{Khan, A.A.}, \bibinfo{author}{Chaudhari, O.}, \bibinfo{author}{Chandra, R.}, \bibinfo{year}{2023}.
\newblock \bibinfo{title}{A review of ensemble learning and data augmentation models for class imbalanced problems: combination, implementation and evaluation}.
\newblock \bibinfo{journal}{Expert Systems with Applications} , \bibinfo{pages}{122778}.
\bibitem[{Kim et~al.(2020a)Kim, Jeong, Choi and Seo}]{kim2020gan}
\bibinfo{author}{Kim, J.}, \bibinfo{author}{Jeong, K.}, \bibinfo{author}{Choi, H.}, \bibinfo{author}{Seo, K.}, \bibinfo{year}{2020}a.
\newblock \bibinfo{title}{Gan-based anomaly detection in imbalance problems}, in: \bibinfo{booktitle}{Computer Vision--ECCV 2020 Workshops: Glasgow, UK, August 23--28, 2020, Proceedings, Part VI 16}, \bibinfo{organization}{Springer}. pp. \bibinfo{pages}{128--145}.
\bibitem[{Kim et~al.(2022)Kim, Jonoski and Solomatine}]{kim2022classification}
\bibinfo{author}{Kim, J.}, \bibinfo{author}{Jonoski, A.}, \bibinfo{author}{Solomatine, D.P.}, \bibinfo{year}{2022}.
\newblock \bibinfo{title}{A classification-based machine learning approach to the prediction of cyanobacterial blooms in chilgok weir, south korea}.
\newblock \bibinfo{journal}{Water} \bibinfo{volume}{14}, \bibinfo{pages}{542}.
\bibitem[{Kim et~al.(2020b)Kim, Kim, Mehrotra and Sharma}]{kim2020predicting}
\bibinfo{author}{Kim, S.}, \bibinfo{author}{Kim, S.}, \bibinfo{author}{Mehrotra, R.}, \bibinfo{author}{Sharma, A.}, \bibinfo{year}{2020}b.
\newblock \bibinfo{title}{Predicting cyanobacteria occurrence using climatological and environmental controls}.
\newblock \bibinfo{journal}{Water Research} \bibinfo{volume}{175}, \bibinfo{pages}{115639}.
\bibitem[{Kornbrot(2014)}]{kornbrot2014point}
\bibinfo{author}{Kornbrot, D.}, \bibinfo{year}{2014}.
\newblock \bibinfo{title}{Point biserial correlation}.
\newblock \bibinfo{journal}{Wiley StatsRef: Statistics Reference Online} .
\bibitem[{Krawczyk(2016)}]{krawczyk2016learning}
\bibinfo{author}{Krawczyk, B.}, \bibinfo{year}{2016}.
\newblock \bibinfo{title}{Learning from imbalanced data: open challenges and future directions}.
\newblock \bibinfo{journal}{Progress in artificial intelligence} \bibinfo{volume}{5}, \bibinfo{pages}{221--232}.
\bibitem[{Kubat et~al.(1997)Kubat, Holte and Matwin}]{kubat1997learning}
\bibinfo{author}{Kubat, M.}, \bibinfo{author}{Holte, R.}, \bibinfo{author}{Matwin, S.}, \bibinfo{year}{1997}.
\newblock \bibinfo{title}{Learning when negative examples abound}, in: \bibinfo{booktitle}{Machine Learning: ECML-97: 9th European Conference on Machine Learning Prague, Czech Republic, April 23--25, 1997 Proceedings 9}, \bibinfo{organization}{Springer}. pp. \bibinfo{pages}{146--153}.
\bibitem[{Kulkarni and Chandra(2024)}]{kulkarni2024bayes}
\bibinfo{author}{Kulkarni, O.}, \bibinfo{author}{Chandra, R.}, \bibinfo{year}{2024}.
\newblock \bibinfo{title}{Bayes-catsi: A variational bayesian deep learning framework for medical time series data imputation}.
\newblock \bibinfo{journal}{arXiv preprint arXiv:2410.01847} .
\bibitem[{Lansner and Ekeberg(1989)}]{lansner1989one}
\bibinfo{author}{Lansner, A.}, \bibinfo{author}{Ekeberg, {\"O}.}, \bibinfo{year}{1989}.
\newblock \bibinfo{title}{A one-layer feedback artificial neural network with a bayesian learning rule}.
\newblock \bibinfo{journal}{International journal of neural systems} \bibinfo{volume}{1}, \bibinfo{pages}{77--87}.
\bibitem[{Lee(2010)}]{lee2010circular}
\bibinfo{author}{Lee, A.}, \bibinfo{year}{2010}.
\newblock \bibinfo{title}{Circular data}.
\newblock \bibinfo{journal}{Wiley Interdisciplinary Reviews: Computational Statistics} \bibinfo{volume}{2}, \bibinfo{pages}{477--486}.
\bibitem[{Lee and Park(2021)}]{lee2021gan}
\bibinfo{author}{Lee, J.}, \bibinfo{author}{Park, K.}, \bibinfo{year}{2021}.
\newblock \bibinfo{title}{Gan-based imbalanced data intrusion detection system}.
\newblock \bibinfo{journal}{Personal and Ubiquitous Computing} \bibinfo{volume}{25}, \bibinfo{pages}{121--128}.
\bibitem[{Lek and Gu{\'e}gan(1999)}]{lek1999artificial}
\bibinfo{author}{Lek, S.}, \bibinfo{author}{Gu{\'e}gan, J.F.}, \bibinfo{year}{1999}.
\newblock \bibinfo{title}{Artificial neural networks as a tool in ecological modelling, an introduction}.
\newblock \bibinfo{journal}{Ecological modelling} \bibinfo{volume}{120}, \bibinfo{pages}{65--73}.
\bibitem[{Li et~al.(2022)Li, An, Li, Wang, Yu, Zhou and Geng}]{li2022application}
\bibinfo{author}{Li, J.}, \bibinfo{author}{An, X.}, \bibinfo{author}{Li, Q.}, \bibinfo{author}{Wang, C.}, \bibinfo{author}{Yu, H.}, \bibinfo{author}{Zhou, X.}, \bibinfo{author}{Geng, Y.a.}, \bibinfo{year}{2022}.
\newblock \bibinfo{title}{Application of xgboost algorithm in the optimization of pollutant concentration}.
\newblock \bibinfo{journal}{Atmospheric Research} \bibinfo{volume}{276}, \bibinfo{pages}{106238}.
\bibitem[{Li et~al.(2011)Li, Li and Yu}]{li2011application}
\bibinfo{author}{Li, J.}, \bibinfo{author}{Li, H.}, \bibinfo{author}{Yu, J.L.}, \bibinfo{year}{2011}.
\newblock \bibinfo{title}{Application of random-smote on imbalanced data mining}, in: \bibinfo{booktitle}{2011 Fourth International Conference on Business Intelligence and Financial Engineering}, \bibinfo{organization}{IEEE}. pp. \bibinfo{pages}{130--133}.
\bibitem[{Liao et~al.(2022)Liao, Hu, Yang and Rosenhahn}]{liao2022text}
\bibinfo{author}{Liao, W.}, \bibinfo{author}{Hu, K.}, \bibinfo{author}{Yang, M.Y.}, \bibinfo{author}{Rosenhahn, B.}, \bibinfo{year}{2022}.
\newblock \bibinfo{title}{Text to image generation with semantic-spatial aware gan}, in: \bibinfo{booktitle}{Proceedings of the IEEE/CVF conference on computer vision and pattern recognition}, pp. \bibinfo{pages}{18187--18196}.
\bibitem[{Liu(2014)}]{liu2014random}
\bibinfo{author}{Liu, Y.}, \bibinfo{year}{2014}.
\newblock \bibinfo{title}{Random forest algorithm in big data environment}.
\newblock \bibinfo{journal}{Computer modelling \& new technologies} \bibinfo{volume}{18}, \bibinfo{pages}{147--151}.
\bibitem[{Liu et~al.(2010)Liu, Peng, Xiang, Tian, Deng and Zhao}]{liu2010application}
\bibinfo{author}{Liu, Z.}, \bibinfo{author}{Peng, C.}, \bibinfo{author}{Xiang, W.}, \bibinfo{author}{Tian, D.}, \bibinfo{author}{Deng, X.}, \bibinfo{author}{Zhao, M.}, \bibinfo{year}{2010}.
\newblock \bibinfo{title}{Application of artificial neural networks in global climate change and ecological research: An overview}.
\newblock \bibinfo{journal}{Chinese science bulletin} \bibinfo{volume}{55}, \bibinfo{pages}{3853--3863}.
\bibitem[{Ma et~al.(2021)Ma, Zhao, He, Li, Dong, Wang and Wang}]{ma2021xgboost}
\bibinfo{author}{Ma, M.}, \bibinfo{author}{Zhao, G.}, \bibinfo{author}{He, B.}, \bibinfo{author}{Li, Q.}, \bibinfo{author}{Dong, H.}, \bibinfo{author}{Wang, S.}, \bibinfo{author}{Wang, Z.}, \bibinfo{year}{2021}.
\newblock \bibinfo{title}{Xgboost-based method for flash flood risk assessment}.
\newblock \bibinfo{journal}{Journal of Hydrology} \bibinfo{volume}{598}, \bibinfo{pages}{126382}.
\bibitem[{Manevitz and Yousef(2001)}]{manevitz2001one}
\bibinfo{author}{Manevitz, L.M.}, \bibinfo{author}{Yousef, M.}, \bibinfo{year}{2001}.
\newblock \bibinfo{title}{One-class svms for document classification}.
\newblock \bibinfo{journal}{Journal of machine Learning research} \bibinfo{volume}{2}, \bibinfo{pages}{139--154}.
\bibitem[{Martins(2022)}]{martins2022modelling}
\bibinfo{author}{Martins, L.C.}, \bibinfo{year}{2022}.
\newblock \bibinfo{title}{Modelling the occurrence of Physalia physalis in the North Atlantic Ocean at different spatial and temporal scales}.
\newblock Ph.D. thesis.
\bibitem[{Martins et~al.(2024)Martins, Gomes-Pereira, Dion{\'\i}sio and Assis}]{martins2024unravelling}
\bibinfo{author}{Martins, L.C.}, \bibinfo{author}{Gomes-Pereira, J.N.}, \bibinfo{author}{Dion{\'\i}sio, G.}, \bibinfo{author}{Assis, J.}, \bibinfo{year}{2024}.
\newblock \bibinfo{title}{Unravelling environmental drivers and patterns of portuguese man o'war (physalia physalis) blooms in two ocean regions: North atlantic and the southeast pacific}.
\newblock \bibinfo{journal}{Marine Pollution Bulletin} \bibinfo{volume}{209}, \bibinfo{pages}{117278}.
\bibitem[{Masko and Hensman(2015)}]{masko2015impact}
\bibinfo{author}{Masko, D.}, \bibinfo{author}{Hensman, P.}, \bibinfo{year}{2015}.
\newblock \bibinfo{title}{The impact of imbalanced training data for convolutional neural networks}.
\bibitem[{Mellios et~al.(2020)Mellios, Moe and Laspidou}]{mellios2020machine}
\bibinfo{author}{Mellios, N.}, \bibinfo{author}{Moe, S.J.}, \bibinfo{author}{Laspidou, C.}, \bibinfo{year}{2020}.
\newblock \bibinfo{title}{Machine learning approaches for predicting health risk of cyanobacterial blooms in northern european lakes}.
\newblock \bibinfo{journal}{Water} \bibinfo{volume}{12}, \bibinfo{pages}{1191}.
\bibitem[{Meng(2013)}]{meng2013scalable}
\bibinfo{author}{Meng, X.}, \bibinfo{year}{2013}.
\newblock \bibinfo{title}{Scalable simple random sampling and stratified sampling}, in: \bibinfo{booktitle}{International conference on machine learning}, \bibinfo{organization}{PMLR}. pp. \bibinfo{pages}{531--539}.
\bibitem[{Metsalu and Vilo(2015)}]{metsalu2015clustvis}
\bibinfo{author}{Metsalu, T.}, \bibinfo{author}{Vilo, J.}, \bibinfo{year}{2015}.
\newblock \bibinfo{title}{Clustvis: a web tool for visualizing clustering of multivariate data using principal component analysis and heatmap}.
\newblock \bibinfo{journal}{Nucleic acids research} \bibinfo{volume}{43}, \bibinfo{pages}{W566--W570}.
\bibitem[{Mills(1995)}]{mills1995medusae}
\bibinfo{author}{Mills, C.E.}, \bibinfo{year}{1995}.
\newblock \bibinfo{title}{Medusae, siphonophores, and ctenophores as planktivorous predators in changing global ecosystems}.
\newblock \bibinfo{journal}{ICES Journal of Marine Science} \bibinfo{volume}{52}, \bibinfo{pages}{575--581}.
\bibitem[{Mohammed et~al.(2020)Mohammed, Rawashdeh and Abdullah}]{mohammed2020machine}
\bibinfo{author}{Mohammed, R.}, \bibinfo{author}{Rawashdeh, J.}, \bibinfo{author}{Abdullah, M.}, \bibinfo{year}{2020}.
\newblock \bibinfo{title}{Machine learning with oversampling and undersampling techniques: overview study and experimental results}, in: \bibinfo{booktitle}{2020 11th international conference on information and communication systems (ICICS)}, \bibinfo{organization}{IEEE}. pp. \bibinfo{pages}{243--248}.
\bibitem[{More and Rana(2017)}]{more2017review}
\bibinfo{author}{More, A.}, \bibinfo{author}{Rana, D.P.}, \bibinfo{year}{2017}.
\newblock \bibinfo{title}{Review of random forest classification techniques to resolve data imbalance}, in: \bibinfo{booktitle}{2017 1st International conference on intelligent systems and information management (ICISIM)}, \bibinfo{organization}{IEEE}. pp. \bibinfo{pages}{72--78}.
\bibitem[{Mukherjee and Khushi(2021)}]{mukherjee2021smote}
\bibinfo{author}{Mukherjee, M.}, \bibinfo{author}{Khushi, M.}, \bibinfo{year}{2021}.
\newblock \bibinfo{title}{Smote-enc: A novel smote-based method to generate synthetic data for nominal and continuous features}.
\newblock \bibinfo{journal}{Applied system innovation} \bibinfo{volume}{4}, \bibinfo{pages}{18}.
\bibitem[{M{\"u}ller et~al.(2018)M{\"u}ller, Mika, Tsuda and Sch{\"o}lkopf}]{muller2018introduction}
\bibinfo{author}{M{\"u}ller, K.R.}, \bibinfo{author}{Mika, S.}, \bibinfo{author}{Tsuda, K.}, \bibinfo{author}{Sch{\"o}lkopf, K.}, \bibinfo{year}{2018}.
\newblock \bibinfo{title}{An introduction to kernel-based learning algorithms}, in: \bibinfo{booktitle}{Handbook of neural network signal processing}. \bibinfo{publisher}{CRC Press}, pp. \bibinfo{pages}{4--1}.
\bibitem[{Mumuni and Mumuni(2022)}]{mumuni2022data}
\bibinfo{author}{Mumuni, A.}, \bibinfo{author}{Mumuni, F.}, \bibinfo{year}{2022}.
\newblock \bibinfo{title}{Data augmentation: A comprehensive survey of modern approaches}.
\newblock \bibinfo{journal}{Array} \bibinfo{volume}{16}, \bibinfo{pages}{100258}.
\bibitem[{Munro et~al.(2019)Munro, Vue, Behringer and Dunn}]{munro2019morphology}
\bibinfo{author}{Munro, C.}, \bibinfo{author}{Vue, Z.}, \bibinfo{author}{Behringer, R.R.}, \bibinfo{author}{Dunn, C.W.}, \bibinfo{year}{2019}.
\newblock \bibinfo{title}{Morphology and development of the portuguese man of war, physalia physalis}.
\newblock \bibinfo{journal}{Scientific reports} \bibinfo{volume}{9}, \bibinfo{pages}{15522}.
\bibitem[{Murphey et~al.(2004)Murphey, Guo and Feldkamp}]{murphey2004neural}
\bibinfo{author}{Murphey, Y.L.}, \bibinfo{author}{Guo, H.}, \bibinfo{author}{Feldkamp, L.A.}, \bibinfo{year}{2004}.
\newblock \bibinfo{title}{Neural learning from unbalanced data}.
\newblock \bibinfo{journal}{Applied Intelligence} \bibinfo{volume}{21}, \bibinfo{pages}{117--128}.
\bibitem[{Niaz et~al.(2022)Niaz, Shahariar and Patwary}]{niaz2022class}
\bibinfo{author}{Niaz, N.U.}, \bibinfo{author}{Shahariar, K.N.}, \bibinfo{author}{Patwary, M.J.}, \bibinfo{year}{2022}.
\newblock \bibinfo{title}{Class imbalance problems in machine learning: A review of methods and future challenges}, in: \bibinfo{booktitle}{Proceedings of the 2nd International Conference on Computing Advancements}, pp. \bibinfo{pages}{485--490}.
\bibitem[{Ogunleye and Wang(2019)}]{ogunleye2019xgboost}
\bibinfo{author}{Ogunleye, A.}, \bibinfo{author}{Wang, Q.G.}, \bibinfo{year}{2019}.
\newblock \bibinfo{title}{Xgboost model for chronic kidney disease diagnosis}.
\newblock \bibinfo{journal}{IEEE/ACM transactions on computational biology and bioinformatics} \bibinfo{volume}{17}, \bibinfo{pages}{2131--2140}.
\bibitem[{Oliveira et~al.(2014)Oliveira, Crujeiras and Rodr{\'\i}guez-Casal}]{oliveira2014circsizer}
\bibinfo{author}{Oliveira, M.}, \bibinfo{author}{Crujeiras, R.M.}, \bibinfo{author}{Rodr{\'\i}guez-Casal, A.}, \bibinfo{year}{2014}.
\newblock \bibinfo{title}{Circsizer: an exploratory tool for circular data}.
\newblock \bibinfo{journal}{Environmental and ecological statistics} \bibinfo{volume}{21}, \bibinfo{pages}{143--159}.
\bibitem[{de~Oliveira and Berton(2023)}]{de2023systematic}
\bibinfo{author}{de~Oliveira, W.D.G.}, \bibinfo{author}{Berton, L.}, \bibinfo{year}{2023}.
\newblock \bibinfo{title}{A systematic review for class-imbalance in semi-supervised learning}.
\newblock \bibinfo{journal}{Artificial Intelligence Review} \bibinfo{volume}{56}, \bibinfo{pages}{2349--2382}.
\bibitem[{Olken(1993)}]{olken1993random}
\bibinfo{author}{Olken, F.}, \bibinfo{year}{1993}.
\newblock \bibinfo{title}{Random sampling from databases}.
\newblock Ph.D. thesis. Citeseer.
\bibitem[{Pal(2005)}]{pal2005random}
\bibinfo{author}{Pal, M.}, \bibinfo{year}{2005}.
\newblock \bibinfo{title}{Random forest classifier for remote sensing classification}.
\newblock \bibinfo{journal}{International journal of remote sensing} \bibinfo{volume}{26}, \bibinfo{pages}{217--222}.
\bibitem[{Pan(2018)}]{pan2018application}
\bibinfo{author}{Pan, B.}, \bibinfo{year}{2018}.
\newblock \bibinfo{title}{Application of xgboost algorithm in hourly pm2. 5 concentration prediction}, in: \bibinfo{booktitle}{IOP conference series: earth and environmental science}, \bibinfo{organization}{IOP publishing}. p. \bibinfo{pages}{012127}.
\bibitem[{Park et~al.(2021)Park, Lee, Shin, Chon, Kim, Cho, Kim and Baek}]{park2021machine}
\bibinfo{author}{Park, Y.}, \bibinfo{author}{Lee, H.K.}, \bibinfo{author}{Shin, J.K.}, \bibinfo{author}{Chon, K.}, \bibinfo{author}{Kim, S.}, \bibinfo{author}{Cho, K.H.}, \bibinfo{author}{Kim, J.H.}, \bibinfo{author}{Baek, S.S.}, \bibinfo{year}{2021}.
\newblock \bibinfo{title}{A machine learning approach for early warning of cyanobacterial bloom outbreaks in a freshwater reservoir}.
\newblock \bibinfo{journal}{Journal of Environmental Management} \bibinfo{volume}{288}, \bibinfo{pages}{112415}.
\bibitem[{Pearson(1901)}]{pearson1901liii}
\bibinfo{author}{Pearson, K.}, \bibinfo{year}{1901}.
\newblock \bibinfo{title}{Liii. on lines and planes of closest fit to systems of points in space}.
\newblock \bibinfo{journal}{The London, Edinburgh, and Dublin philosophical magazine and journal of science} \bibinfo{volume}{2}, \bibinfo{pages}{559--572}.
\bibitem[{Pontin and Cruickshank(2012)}]{pontin2012molecular}
\bibinfo{author}{Pontin, D.}, \bibinfo{author}{Cruickshank, R.}, \bibinfo{year}{2012}.
\newblock \bibinfo{title}{Molecular phylogenetics of the genus physalia (cnidaria: Siphonophora) in new zealand coastal waters reveals cryptic diversity}.
\newblock \bibinfo{journal}{Hydrobiologia} \bibinfo{volume}{686}, \bibinfo{pages}{91--105}.
\bibitem[{Pontin(2009)}]{pontin2009factors}
\bibinfo{author}{Pontin, D.R.}, \bibinfo{year}{2009}.
\newblock \bibinfo{title}{Factors influencing the occurrence of stinging jellyfish (Physalia spp.) at New Zealand beaches}.
\newblock Ph.D. thesis. Lincoln University.
\bibitem[{Pontin et~al.(2009)Pontin, Worner and Watts}]{pontin2009using}
\bibinfo{author}{Pontin, D.R.}, \bibinfo{author}{Worner, S.P.}, \bibinfo{author}{Watts, M.J.}, \bibinfo{year}{2009}.
\newblock \bibinfo{title}{Using time lagged input data to improve prediction of stinging jellyfish occurrence at new zealand beaches by multi-layer perceptrons}, in: \bibinfo{booktitle}{Advances in Neuro-Information Processing: 15th International Conference, ICONIP 2008, Auckland, New Zealand, November 25-28, 2008, Revised Selected Papers, Part I 15}, \bibinfo{organization}{Springer}. pp. \bibinfo{pages}{909--916}.
\bibitem[{Purcell(2005)}]{purcell2005climate}
\bibinfo{author}{Purcell, J.E.}, \bibinfo{year}{2005}.
\newblock \bibinfo{title}{Climate effects on formation of jellyfish and ctenophore blooms: a review}.
\newblock \bibinfo{journal}{Journal of the Marine Biological Association of the United kingdom} \bibinfo{volume}{85}, \bibinfo{pages}{461--476}.
\bibitem[{Quetglas et~al.(2011)Quetglas, Ordines and Guijarro}]{quetglas2011use}
\bibinfo{author}{Quetglas, A.}, \bibinfo{author}{Ordines, F.}, \bibinfo{author}{Guijarro, B.}, \bibinfo{year}{2011}.
\newblock \bibinfo{title}{The use of artificial neural networks (ANNs) in aquatic ecology}.
\newblock \bibinfo{publisher}{Intech Open Access Publisher}.
\bibitem[{Rattan et~al.(2021)Rattan, Mittal, Singh and Malik}]{rattan2021analyzing}
\bibinfo{author}{Rattan, V.}, \bibinfo{author}{Mittal, R.}, \bibinfo{author}{Singh, J.}, \bibinfo{author}{Malik, V.}, \bibinfo{year}{2021}.
\newblock \bibinfo{title}{Analyzing the application of smote on machine learning classifiers}, in: \bibinfo{booktitle}{2021 International Conference on Emerging Smart Computing and Informatics (ESCI)}, \bibinfo{organization}{IEEE}. pp. \bibinfo{pages}{692--695}.
\bibitem[{Rezvani and Wang(2023)}]{rezvani2023broad}
\bibinfo{author}{Rezvani, S.}, \bibinfo{author}{Wang, X.}, \bibinfo{year}{2023}.
\newblock \bibinfo{title}{A broad review on class imbalance learning techniques}.
\newblock \bibinfo{journal}{Applied Soft Computing} \bibinfo{volume}{143}, \bibinfo{pages}{110415}.
\bibitem[{Rider et~al.(2013)Rider, Johnson, Davis, Hoens and Chawla}]{rider2013classifier}
\bibinfo{author}{Rider, A.K.}, \bibinfo{author}{Johnson, R.A.}, \bibinfo{author}{Davis, D.A.}, \bibinfo{author}{Hoens, T.R.}, \bibinfo{author}{Chawla, N.V.}, \bibinfo{year}{2013}.
\newblock \bibinfo{title}{Classifier evaluation with missing negative class labels}, in: \bibinfo{booktitle}{Advances in Intelligent Data Analysis XII: 12th International Symposium, IDA 2013, London, UK, October 17-19, 2013. Proceedings 12}, \bibinfo{organization}{Springer}. pp. \bibinfo{pages}{380--391}.
\bibitem[{Sauber-Cole and Khoshgoftaar(2022)}]{sauber2022use}
\bibinfo{author}{Sauber-Cole, R.}, \bibinfo{author}{Khoshgoftaar, T.M.}, \bibinfo{year}{2022}.
\newblock \bibinfo{title}{The use of generative adversarial networks to alleviate class imbalance in tabular data: a survey}.
\newblock \bibinfo{journal}{Journal of Big Data} \bibinfo{volume}{9}, \bibinfo{pages}{98}.
\bibitem[{Sharma et~al.(2022)Sharma, Singh and Chandra}]{sharma2022smotified}
\bibinfo{author}{Sharma, A.}, \bibinfo{author}{Singh, P.K.}, \bibinfo{author}{Chandra, R.}, \bibinfo{year}{2022}.
\newblock \bibinfo{title}{Smotified-gan for class imbalanced pattern classification problems}.
\newblock \bibinfo{journal}{Ieee Access} \bibinfo{volume}{10}, \bibinfo{pages}{30655--30665}.
\bibitem[{Shi and Li(2007)}]{shi2007application}
\bibinfo{author}{Shi, Z.}, \bibinfo{author}{Li, H.}, \bibinfo{year}{2007}.
\newblock \bibinfo{title}{Application of artificial neural network approach and remotely sensed imagery for regional eco-environmental quality evaluation}.
\newblock \bibinfo{journal}{Environmental monitoring and assessment} \bibinfo{volume}{128}, \bibinfo{pages}{217--229}.
\bibitem[{Shin et~al.(2021)Shin, Yoon, Kim, Kim, Go and Cha}]{shin2021effects}
\bibinfo{author}{Shin, J.}, \bibinfo{author}{Yoon, S.}, \bibinfo{author}{Kim, Y.}, \bibinfo{author}{Kim, T.}, \bibinfo{author}{Go, B.}, \bibinfo{author}{Cha, Y.}, \bibinfo{year}{2021}.
\newblock \bibinfo{title}{Effects of class imbalance on resampling and ensemble learning for improved prediction of cyanobacteria blooms}.
\newblock \bibinfo{journal}{Ecological informatics} \bibinfo{volume}{61}, \bibinfo{pages}{101202}.
\bibitem[{Skryjomski and Krawczyk(2017)}]{skryjomski2017influence}
\bibinfo{author}{Skryjomski, P.}, \bibinfo{author}{Krawczyk, B.}, \bibinfo{year}{2017}.
\newblock \bibinfo{title}{Influence of minority class instance types on smote imbalanced data oversampling}, in: \bibinfo{booktitle}{first international workshop on learning with imbalanced domains: theory and applications}, \bibinfo{organization}{Pmlr}. pp. \bibinfo{pages}{7--21}.
\bibitem[{Sokolova and Lapalme(2009)}]{sokolova2009systematic}
\bibinfo{author}{Sokolova, M.}, \bibinfo{author}{Lapalme, G.}, \bibinfo{year}{2009}.
\newblock \bibinfo{title}{A systematic analysis of performance measures for classification tasks}.
\newblock \bibinfo{journal}{Information processing \& management} \bibinfo{volume}{45}, \bibinfo{pages}{427--437}.
\bibitem[{Stein et~al.(1989)Stein, Marraccini, Rothschild and Burnett}]{stein1989fatal}
\bibinfo{author}{Stein, M.R.}, \bibinfo{author}{Marraccini, J.V.}, \bibinfo{author}{Rothschild, N.E.}, \bibinfo{author}{Burnett, J.W.}, \bibinfo{year}{1989}.
\newblock \bibinfo{title}{Fatal portuguese man-o'-war (physalia physalis) envenomation}.
\newblock \bibinfo{journal}{Annals of emergency medicine} \bibinfo{volume}{18}, \bibinfo{pages}{312--315}.
\bibitem[{Su et~al.(2019)Su, Eizenberg, Steinle, Jakob, Fox-Hughes, White, Rennie, Franklin, Dharssi and Zhu}]{su2019barra}
\bibinfo{author}{Su, C.H.}, \bibinfo{author}{Eizenberg, N.}, \bibinfo{author}{Steinle, P.}, \bibinfo{author}{Jakob, D.}, \bibinfo{author}{Fox-Hughes, P.}, \bibinfo{author}{White, C.J.}, \bibinfo{author}{Rennie, S.}, \bibinfo{author}{Franklin, C.}, \bibinfo{author}{Dharssi, I.}, \bibinfo{author}{Zhu, H.}, \bibinfo{year}{2019}.
\newblock \bibinfo{title}{Barra v1. 0: the bureau of meteorology atmospheric high-resolution regional reanalysis for australia}.
\newblock \bibinfo{journal}{Geoscientific Model Development} \bibinfo{volume}{12}, \bibinfo{pages}{2049--2068}.
\bibitem[{Tate(1954)}]{tate1954correlation}
\bibinfo{author}{Tate, R.F.}, \bibinfo{year}{1954}.
\newblock \bibinfo{title}{Correlation between a discrete and a continuous variable. point-biserial correlation}.
\newblock \bibinfo{journal}{The Annals of mathematical statistics} \bibinfo{volume}{25}, \bibinfo{pages}{603--607}.
\bibitem[{Uddin et~al.(2021)Uddin, Mamun and Hossain}]{uddin2021pca}
\bibinfo{author}{Uddin, M.P.}, \bibinfo{author}{Mamun, M.A.}, \bibinfo{author}{Hossain, M.A.}, \bibinfo{year}{2021}.
\newblock \bibinfo{title}{Pca-based feature reduction for hyperspectral remote sensing image classification}.
\newblock \bibinfo{journal}{IETE Technical Review} \bibinfo{volume}{38}, \bibinfo{pages}{377--396}.
\bibitem[{Van~Dyk and Meng(2001)}]{van2001art}
\bibinfo{author}{Van~Dyk, D.A.}, \bibinfo{author}{Meng, X.L.}, \bibinfo{year}{2001}.
\newblock \bibinfo{title}{The art of data augmentation}.
\newblock \bibinfo{journal}{Journal of Computational and Graphical Statistics} \bibinfo{volume}{10}, \bibinfo{pages}{1--50}.
\bibitem[{Wood(1994)}]{wood1994simulation}
\bibinfo{author}{Wood, A.T.}, \bibinfo{year}{1994}.
\newblock \bibinfo{title}{Simulation of the von mises fisher distribution}.
\newblock \bibinfo{journal}{Communications in statistics-simulation and computation} \bibinfo{volume}{23}, \bibinfo{pages}{157--164}.
\bibitem[{Xu et~al.(2019)Xu, Skoularidou, Cuesta-Infante and Veeramachaneni}]{xu2019modeling}
\bibinfo{author}{Xu, L.}, \bibinfo{author}{Skoularidou, M.}, \bibinfo{author}{Cuesta-Infante, A.}, \bibinfo{author}{Veeramachaneni, K.}, \bibinfo{year}{2019}.
\newblock \bibinfo{title}{Modeling tabular data using conditional gan}.
\newblock \bibinfo{journal}{Advances in neural information processing systems} \bibinfo{volume}{32}.
\bibitem[{Yu et~al.(2004)Yu, Han and Chang}]{yu2004pebl}
\bibinfo{author}{Yu, H.}, \bibinfo{author}{Han, J.}, \bibinfo{author}{Chang, K.C.}, \bibinfo{year}{2004}.
\newblock \bibinfo{title}{Pebl: Web page classification without negative examples}.
\newblock \bibinfo{journal}{IEEE Transactions on Knowledge and Data Engineering} \bibinfo{volume}{16}, \bibinfo{pages}{70--81}.
\bibitem[{Zhang et~al.(2022)Zhang, Jia and Shang}]{zhang2022research}
\bibinfo{author}{Zhang, P.}, \bibinfo{author}{Jia, Y.}, \bibinfo{author}{Shang, Y.}, \bibinfo{year}{2022}.
\newblock \bibinfo{title}{Research and application of xgboost in imbalanced data}.
\newblock \bibinfo{journal}{International Journal of Distributed Sensor Networks} \bibinfo{volume}{18}, \bibinfo{pages}{15501329221106935}.
\bibitem[{Zhang et~al.(2020)Zhang, Yang, Chen, Liu, Meng, Wang and Li}]{zhang2020generative}
\bibinfo{author}{Zhang, Z.}, \bibinfo{author}{Yang, L.}, \bibinfo{author}{Chen, L.}, \bibinfo{author}{Liu, Q.}, \bibinfo{author}{Meng, Y.}, \bibinfo{author}{Wang, P.}, \bibinfo{author}{Li, M.}, \bibinfo{year}{2020}.
\newblock \bibinfo{title}{A generative adversarial network--based method for generating negative financial samples}.
\newblock \bibinfo{journal}{International Journal of Distributed Sensor Networks} \bibinfo{volume}{16}, \bibinfo{pages}{1550147720907053}.
\bibitem[{Zhou(2012)}]{zhou2012ensemble}
\bibinfo{author}{Zhou, Z.H.}, \bibinfo{year}{2012}.
\newblock \bibinfo{title}{Ensemble methods: foundations and algorithms}.
\newblock \bibinfo{publisher}{CRC press}.
\bibitem[{Zou et~al.(2009)Zou, Han and So}]{zou2009overview}
\bibinfo{author}{Zou, J.}, \bibinfo{author}{Han, Y.}, \bibinfo{author}{So, S.S.}, \bibinfo{year}{2009}.
\newblock \bibinfo{title}{Overview of artificial neural networks}.
\newblock \bibinfo{journal}{Artificial neural networks: methods and applications} , \bibinfo{pages}{14--22}.
\bibitem[{Zulkipli et~al.(2021)Zulkipli, Satari and Yusoff}]{zulkipli2021synthetic}
\bibinfo{author}{Zulkipli, N.}, \bibinfo{author}{Satari, S.}, \bibinfo{author}{Yusoff, W.W.}, \bibinfo{year}{2021}.
\newblock \bibinfo{title}{A synthetic data generation procedure for univariate circular data with various outliers scenarios using python programming language}, in: \bibinfo{booktitle}{Journal of Physics: Conference Series}, \bibinfo{organization}{IOP Publishing}. p. \bibinfo{pages}{012111}.

\end{thebibliography}

\appendix

\section{No Data augumentation}

\begin{figure*}[htbp]
   \centering
   \includegraphics[width=1.0\textwidth]{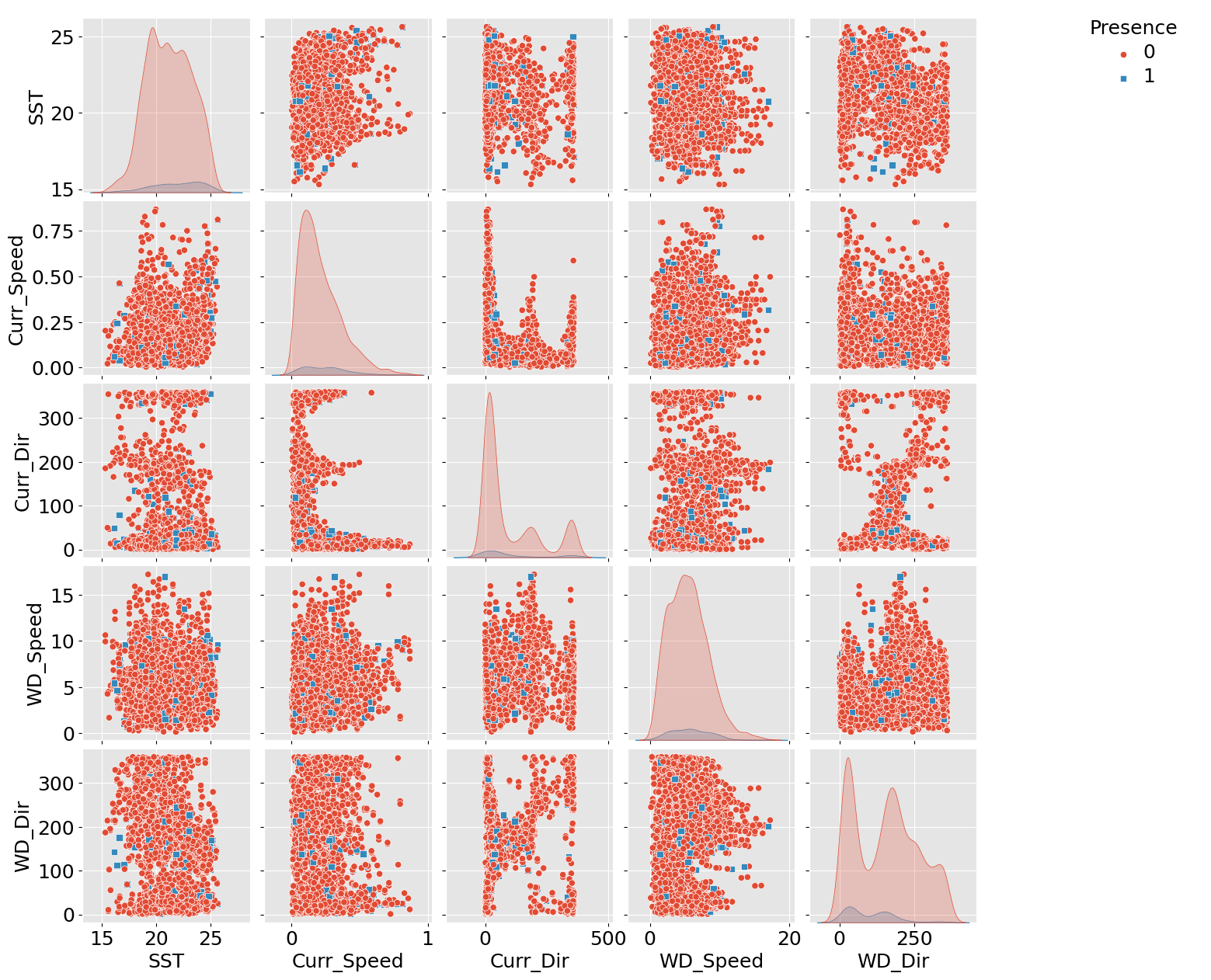}
   \caption{Pair plot of the continuous variables of the actual dataset plotted against the output variable presence (blue) or absence (red) of bluebottle. The plot shows extreme class imbalance with overlap (no clear line of separation between the two classes) with no pair capable of showing a decision boundary between the classes. }
   \label{fig:scatterplot}
\end{figure*} 

\section{Data augumentation}

\begin{figure*}[htbp]
    \centering
    \small
    \includegraphics[width=1.0\textwidth]{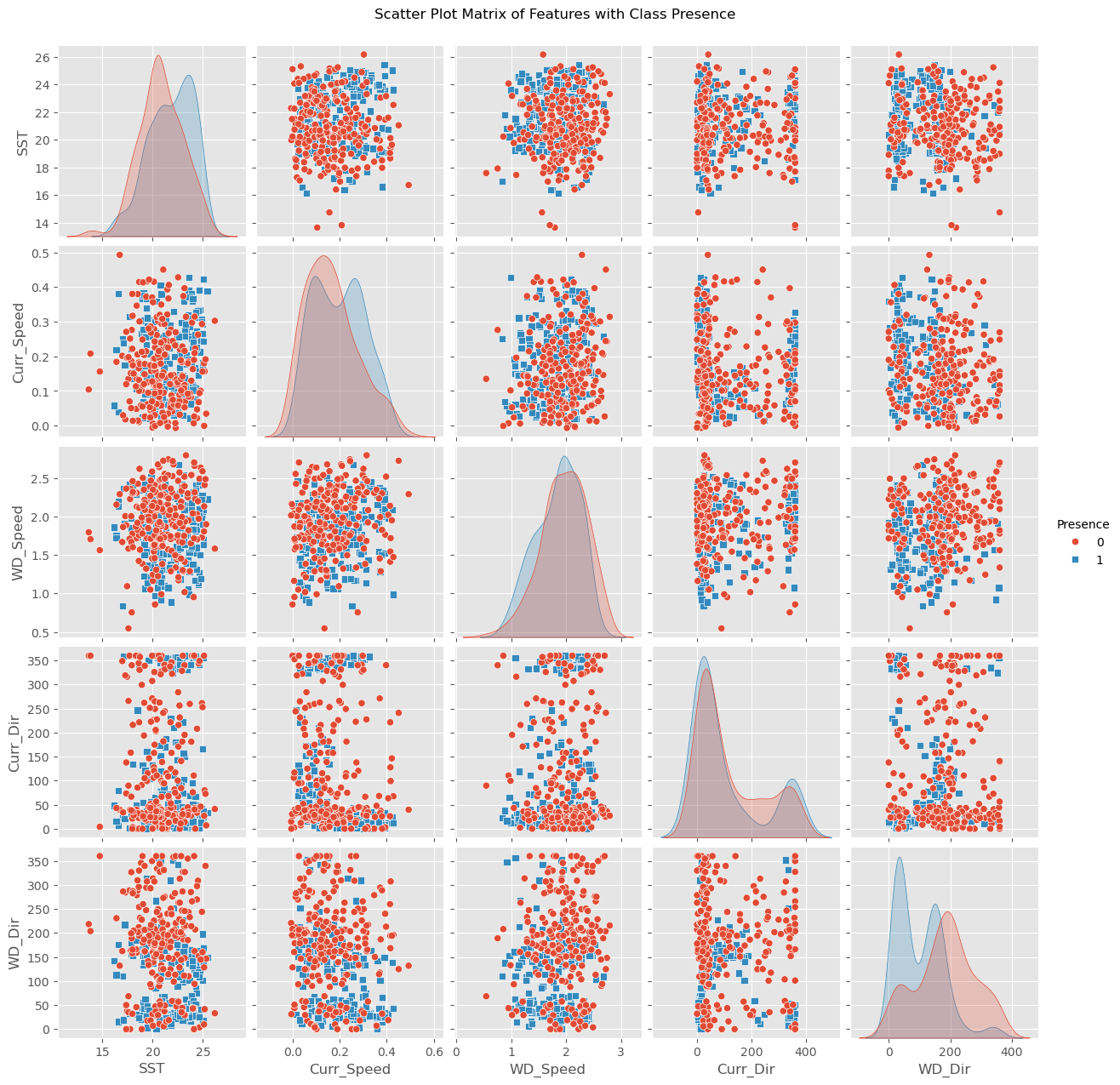}
    \caption{Pair plot of the continuous variables of  the actual positive class and the synthetic negative class generated from CTGAN plotted against the output variable presence (blue) or absence (red) of bluebottle. The plot shows equal samples of classes with possible separation between the two classes) and some pair capable of showing a some clusters and decision boundary between the classes.}
    \label{fig:pairplotctgan}
\end{figure*}
\end{document}